\documentclass{article} 
\usepackage{acl}
\usepackage{times}
\usepackage{microtype}
\usepackage{hyperref}
\usepackage{url}
\usepackage{booktabs}
\usepackage{graphicx}
\usepackage{colortbl}
\usepackage{xcolor}
\usepackage{array}
\usepackage{calc}
\usepackage{pgfmath}
\usepackage{multirow}
\usepackage[most]{tcolorbox}
\usepackage{amsmath}
\usepackage{subcaption}
\usepackage{graphicx}
\usepackage{enumitem}

\newcommand{\name}{CultureScore}

\usepackage{lineno}

\newcolumntype{C}{>{\centering\arraybackslash}p{2.4cm}}

\title{\name: Evaluating Cultural Faithfulness\\in Video Generation Models}

\author{
  \textbf{Anku Rani}\textsuperscript{1} \quad 
  \textbf{Wei Dai}\textsuperscript{1} \quad 
  \textbf{Shravan Nayak}\textsuperscript{2} \quad \\
  \textbf{Pattie Maes}\textsuperscript{1} \quad 
  \textbf{Mahdi M. Kalayeh}\textsuperscript{3} \quad
  \textbf{Paul Pu Liang}\textsuperscript{1} \\
  \textsuperscript{1}Massachusetts Institute of Technology \quad
  \textsuperscript{2}Mila -- Quebec AI Institute \quad
  \textsuperscript{3}Netflix \\
  Correspondence: \texttt{ankurani@mit.edu} \\
  Project page: \url{https://culturescore.vercel.app}
}

\begin{document}


\maketitle

\begin{abstract}
As video generation models like Veo 3.1 and LTX-2 advance, their ability to accurately represent diverse global cultures remains a critical yet understudied frontier. Current metrics, such as VideoScore, only measure visual quality but offer no mechanism for assessing cultural faithfulness. Consequently, a model that replaces a Namaste with a handshake receives the same score as one that generates the gesture correctly.
We propose CultureScore, a compositional evaluation framework that decomposes cultural faithfulness into three granular dimensions: Identity (who is represented), Context (culturally localized background), and Behavior (normative gestures and interactions). We operationalize this framework through an evaluation suite spanning 10 countries, yielding 6,174 generated videos across three state-of-the-art models. Our evaluation reveals that no current model achieves culturally faithful video generation: the best-performing model reaches only 56.8\% overall CultureScore, with Behavior the most challenging dimension; no model exceeds 52.1\% on behavior.
Furthermore, the highest-scoring model (LTX-2) on visual quality was ranked last by native annotators, while CultureScore's Behavior dimension shows the strongest positive correlation with human cultural judgment among the automatic metrics we evaluate, underscoring that cultural faithfulness is an essential criterion for equitable video generation. Data and code are publicly available.\footnote{\url{https://huggingface.co/datasets/ankurani/CultureScore}}

\end{abstract}

\section{Introduction}

Video Generation models like Veo3~\citep{GoogleDeepMind2025Veo3}, Sora~\citep{openai2024videoworldsimulators}, and Wan2.2~\citep{wan2025wanopenadvancedlargescale} are capable of simulating the physical world with high faithfulness. They are already used to generate advertisements, social media content, and short films~\citep{Hume2025Flow}. As adoption grows globally, a fundamental question arises about how faithfully these models represent the world's cultural diversity. A system trained predominantly on Western media will generate a handshake when prompted for a greeting, not a Namaste or a Salam. These systematic biases may marginalize the cultural norms of billions of people as video generation tools spread worldwide. Ensuring that these models can accurately depict diverse cultural realities is essential for equitable access to generative AI, with implications for education, media, policy communication, and social inclusion.

Research on cultural evaluation in generative AI has so far focused almost entirely on text-to-image (T2I) models. Benchmarks like CulturalFrames~\citep{nayak-etal-2025-culturalframes}, CuRE~\citep{rege2025cureculturalgapslong}, CUBE~\citep{kannenbeyondaesthetics}, and CultDiff~\citep{bayramli2025diffusionmodelsgloballens} have collectively revealed that T2I models frequently miss cultural cues and default to stereotypical depictions.
However, images are static, while culture is often expressed through actions in sequence, requiring evaluation frameworks that reason over motion sequences rather than single frames. A Namaste requires folding the hands and bowing in a specific order; a Salam carries its own gesture and rhythm. Images cannot capture these distinctions, and neither can evaluation frameworks designed for them. On the video side, the landscape is sparse. VideoScore~\citep{he2024videoscore} and UnifiedReward~\citep{wang2025unified} measure temporal consistency, visual quality, and factual grounding.
While useful for assessing general video quality, these metrics can be actively misleading when used for cultural faithfulness. For example, VideoScore assigns equal scores for videos that replace a Namaste with a handshake.

\begin{figure*}[t]
    \centering
    \includegraphics[width=1\textwidth]{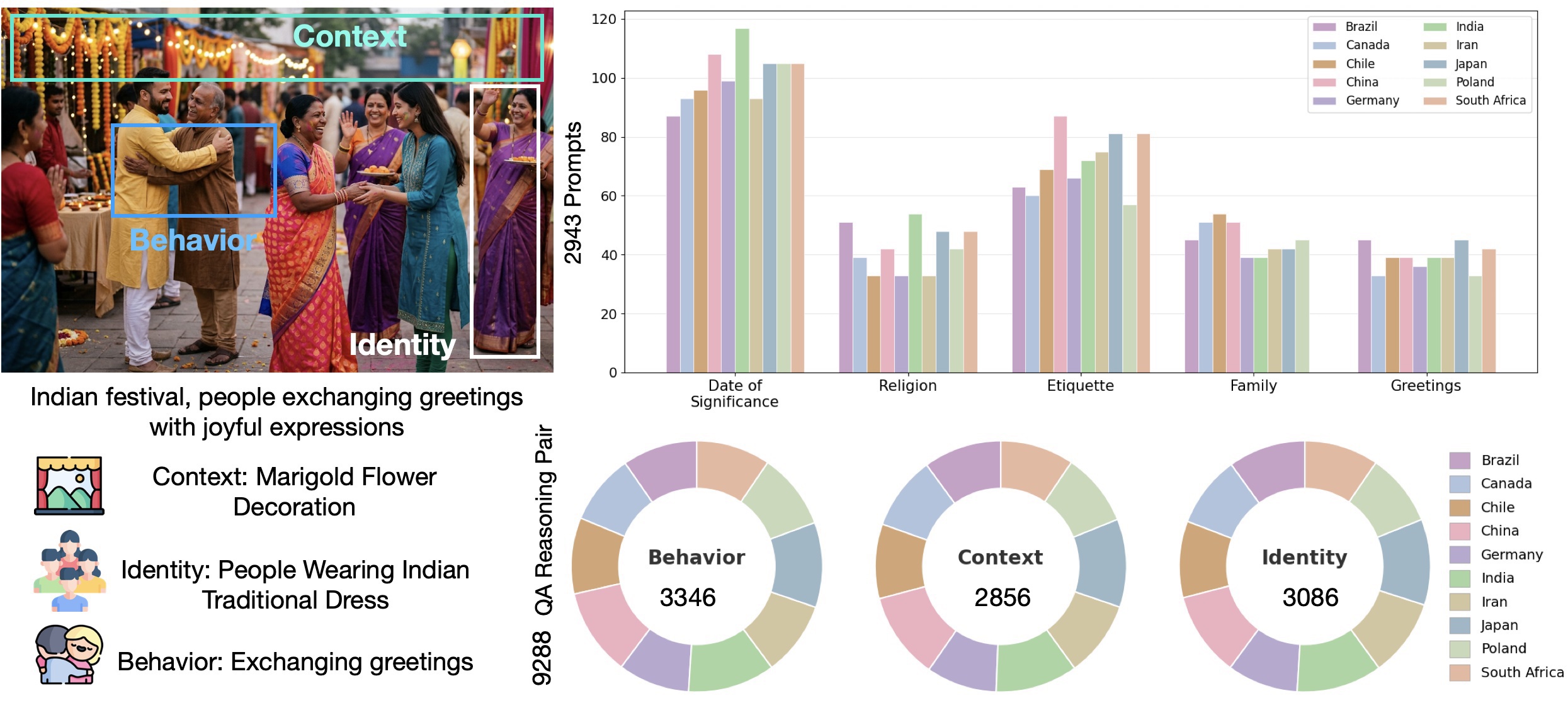}
    \caption{\name\ is a new compositional evaluation framework that decomposes cultural faithfulness into three dimensions: Identity (who is represented), Context (culturally localized background), and Behavior (normative gestures and expressivity). We operationalize this framework through an evaluation suite of 2,943 culturally validated prompts spanning 10 countries and 5 socio-cultural domains, yielding 6,174 generated videos and 9,288 reasoning and QA pairs across three state-of-the-art models: Veo 3.1 Fast \citep{GoogleDeepMind2025Veo3}, LTX-2 \citep{hacohen2024ltxvideorealtimevideolatent}, and Wan 2.2 \citep{wan2025wanopenadvancedlargescale}.}
    \vspace{-2mm}
    \label{fig:Intro_stats}
\end{figure*}

To address these limitations, we propose a compositional approach to evaluating cultural faithfulness in generated videos. Prior work in video understanding has shown that complex visual scenes can be more reliably analyzed when decomposed into semantically meaningful components like actions, scenes, and objects~\citep{ray2018scenes}. Inspired by compositionality, we propose an evaluation framework called \name\ that decomposes a textual prompt into three culturally grounded facets: identity (who is represented and how), behavior (culturally normative gestures, prosody, expressivity), and context (culturally situated settings and social conventions). Unlike holistic scores, this fine-grained metric exposes subtle cultural mismatches, such as an incorrect greeting gesture performed in the right setting, and identifies precisely which component failed. The result is not just a faithfulness score but an interpretable diagnostic that reveals where and how a model diverges from authentic cultural representation.

We operationalize this framework by building an evaluation dataset grounded in CulturalFrames~\citep{nayak-etal-2025-culturalframes}, a benchmark of 981 culturally validated settings spanning 10 countries and 5 socio-cultural domains. Each setting is decomposed into 9288 question-answer pairs across identity, behavior, and context components (see Figure \ref{fig:Intro_stats}).
These QA-pairs are used to evaluate the cultural faithfulness of state-of-the-art video generation models, producing component-level accuracy scores that aggregate into an overall \name. Our evaluation reveals a striking finding: models that score highest on perceptual metrics systematically score lowest on cultural faithfulness, and vice versa — a divergence that is reflected in native human preference rankings.

\paragraph{Contributions.}
To summarize, this work makes the following contributions to the study of cultural faithfulness in video generation models.
\begin{enumerate}[topsep=0pt,leftmargin=*,parsep=0pt,partopsep=0pt]
    \item We propose \textbf{\name}, a compositional evaluation framework that decomposes cultural faithfulness into Identity, Behavior, and Context dimensions, enabling fine-grained diagnosis of when models diverge from authentic cultural representation.

    \item We construct and publicly release a \textbf{culturally grounded evaluation suite} of 2,943 prompts spanning 10 countries and 5 socio-cultural domains. It yields 6,174 videos and 9,288 reasoning and QA pairs, providing a reusable benchmark for future work on cultural faithfulness in video generation.

    \item We present a comprehensive analysis of three state-of-the-art video generation models across 10 countries, identifying systematic cultural failure modes, including a striking \textbf{inverse correlation between current video generation metrics and cultural faithfulness}, corroborated by native human preference rankings.
    
\end{enumerate}

\section{Related Work}

\paragraph{Video generation models.} 
Early video generation models, such as Make-A-Video~\citep{singer2023makeavideo}, built on the success of text-to-image diffusion models by extending spatial U-Net architectures with temporal attention modules.
A significant architectural shift occurred in 2024 with the adoption of Diffusion Transformer (DiT) backbones~\citep{peebles2023scalablediffusionmodelstransformers, wang2024recipe, lee2024grid}. Further, OpenAI's Sora~\citep{openai2024videoworldsimulators} demonstrated that scaling DiT-based video models yields strong temporal consistency and semantic coherence. This catalyzed a wave of open-source DiT-based video generation models, including CogVideoX~\citep{yang2025cogvideox}, HunyuanVideo~\citep{kong2025hunyuanvideosystematicframeworklarge}, LTX-Video~\citep{hacohen2024ltxvideorealtimevideolatent}, and Wan~\citep{wan2025wanopenadvancedlargescale} that have made substantial progress on text-video alignment, complex motion generation, real-time video generation, and instruction following.
Despite these advances, video generation models are primarily evaluated on physical realism, temporal consistency, and general prompt adherence~\citep{he2024videoscore, wang2025unified}. Their capacity to faithfully represent different cultures and their nuances remains entirely unstudied. 

\paragraph{Cultural representation in generative AI.} Research into cultural representation in AI systems across a range of modalities, from language models ~\citep{chiu2025culturalbenchrobustdiversechallenging} to multimodal systems \citep{nayak-etal-2024-benchmarking, bhatia-etal-2024-local}, has shown that frontier models struggle with non-Western cultural knowledge. Later, as text-to-image (T2I) generation matured, researchers extended this inquiry to visual content creation. Benchmarks like CUBE~\citep{kannenbeyondaesthetics}, CuRE~\cite{rege2025cureculturalgapslong}, CulturalFrames~\citep{nayak-etal-2025-culturalframes}, and CultDiff ~\citep{bayramli2025diffusionmodelsgloballens} revealed that T2I models frequently miss cultural cues and default to stereotypical depictions. Image transcreation~\citep{khanuja-etal-2024-image} further highlighted the difficulty of the problem for image editing, where the goal was to adapt cultural content across regions. On the evaluation side, CULTIVate 's~\citep {malakouti2026culture} AHEaD metric measures cultural faithfulness across alignment, hallucination, exaggeration, and diversity dimensions. However, cultural expression in video dynamically unfolds over time through gestures and interactions that a single frame cannot capture. To the best of our knowledge, no existing benchmark or metric addresses cultural faithfulness in video generation models.

\paragraph{Video generation evaluation metrics.} Early metrics like FVD~\citep{unterthiner2019accurategenerativemodelsvideo} measured statistical similarity between generated and real video distributions, while more comprehensive benchmarks such as VBench~\citep{huang2024vbench} and VBench++~\citep{huang2024vbenchcomprehensiveversatilebenchmark} improved coverage by assessing perceptual dimensions like motion smoothness and temporal consistency. Yet none measured reliably whether generated content was semantically faithful to the prompt. MLLMs, with their ability to reason over both visual content and natural language, offer a more promising path, as shown by their success in text-to-image evaluation~\citep{ku2024viescoreexplainablemetricsconditional}. Building on this, VideoScore~\citep{he2024videoscore} and UnifiedReward~\citep{wang2025unified} trained VLMs on human preference data to produce scores across visual quality and semantic faithfulness. VFEval~\citep{song-etal-2025-vf} further confirmed that VLMs provide reliable feedback for evaluating video generation. Yet, none of these works evaluate for cultural faithfulness. In this work, we leverage VLMs to build a compositional, QA-based evaluation framework for evaluating cultural faithfulness, exposing failure modes that current metrics miss.

\section{\name: A Decomposed Evaluation Framework}

Evaluating cultural faithfulness in Video Generation models presents a unique technical challenge: while standard metrics like VideoScore \citep{he2024videoscore} or Unified Reward \citep{wang2025unified} assess visual quality, temporal consistency, dynamic degree, etc., they fail to capture the granular accuracy of cultural norms. 
Formally, given a cultural prompt $P$ representing a specific socio-cultural frame, the goal is to evaluate a video $V$ such that the identity, behavior, and context are all internally consistent with the target culture.
\begin{enumerate}[topsep=0pt,leftmargin=*,parsep=0pt,partopsep=0pt]
    \item \textbf{Identity} refers to who is represented and how they are depicted in the generated video. This dimension focuses on the characters’ physical appearance, attire, and demographic markers. For example, the visual representation of Japanese colleagues or Chinese students should reflect distinct cultural norms in professional or academic dress and physical characteristics.

    \item \textbf{Behavior} encompasses what the people are doing, specifically focusing on culturally normative gestures, and expressivity. Unlike static images, video must capture the specific "gesture and rhythm" of actions. A primary example is the cultural variation in greetings: while a model might default to a Western handshake, a culturally faithful generation would correctly depict a Namaste in India or a Salam, which require specific sequences of motion and posture.

    \item \textbf{Context} refers to the culturally localized background, including the physical setting, environmental details, and underlying social conventions. This dimension covers both the atmospheric decor, such as Islamic calligraphy or Persian rugs, and social arrangements, such as a family seated around a dastarkhwan (floor spread) rather than a Western-style dining table. Figure \ref{fig:Intro_stats} illustrates these dimensions through a comparison of generated content across different cultural prompts.
\end{enumerate}
The technical challenges of this problem are three-fold: (1) \textit{Granularity Gap}: Cultural errors are often subtle and \emph{hidden} within high-quality visual outputs, making holistic scores unreliable; (2) \textit{Label Dependency}: Models often over-rely on explicit geographic tokens (e.g., "India") rather than understanding the underlying cultural concepts; and (3) \textit{Temporal Complexity}: Culturally specific motions (e.g., behavior) can be difficult to generate.

\begin{figure*}
    \centering
    
    \includegraphics[width=1\textwidth]{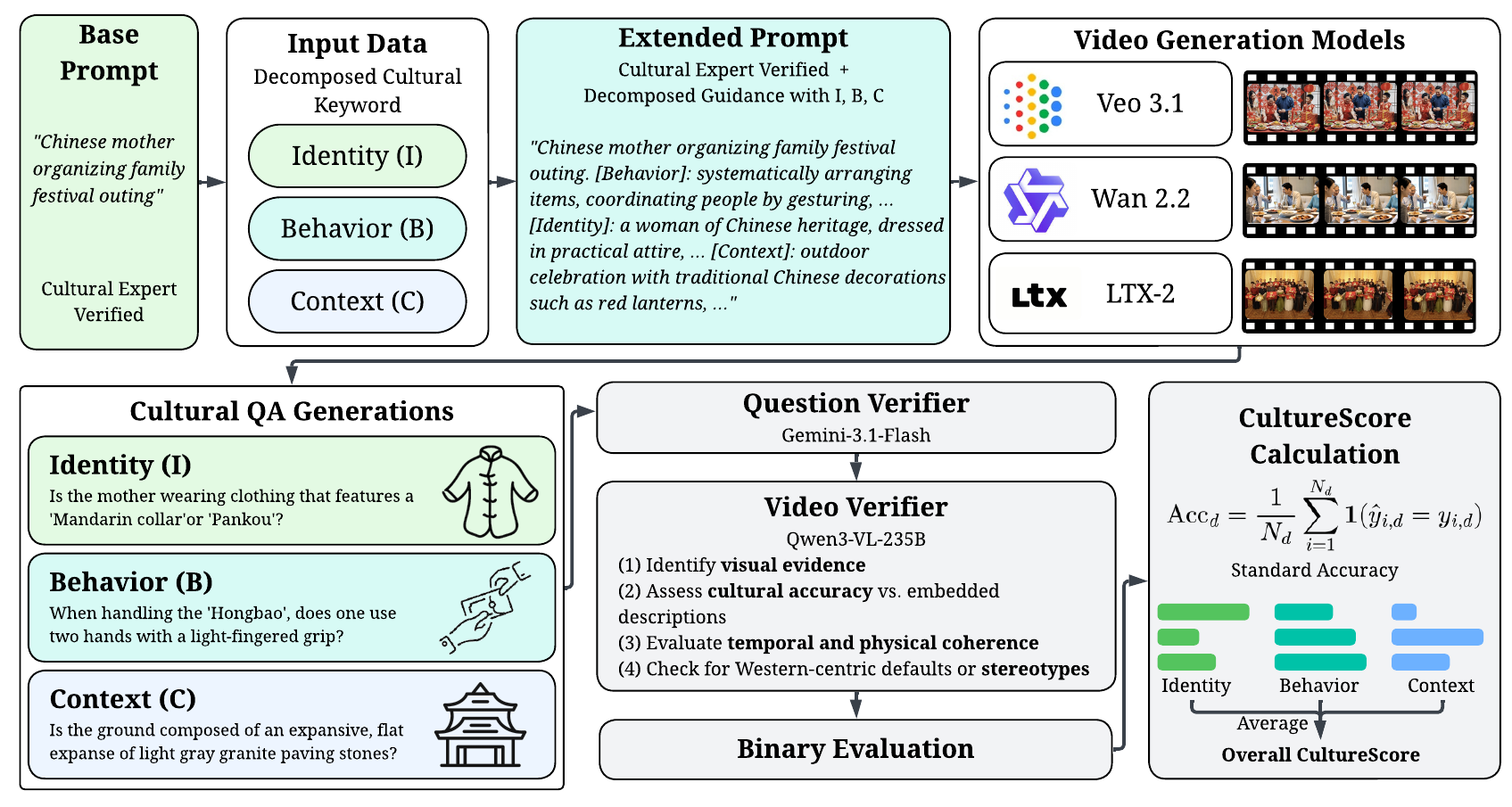}
    \caption{The \name\ evaluation framework. Base prompts are decomposed into Identity, Behavior, and Context dimensions and expanded with elaborated meaning. Videos generated by each model are evaluated via question-answer pairs, which are verified for accuracy by a VLM before scoring. Component-level accuracy scores across the three dimensions are averaged to produce an overall \name.}
    \label{fig:framework}
\end{figure*}

\subsection{Cultural Dimensions and Data Curation}

Our evaluation framework \name\ is grounded in CulturalFrames \citep{nayak-etal-2025-culturalframes}, a benchmark of 981 culturally validated prompts spanning 10 countries and 5 socio-cultural domains, originally developed for evaluating cultural representation in text-to-image models. We are the first to adapt these prompts for video generation evaluation, decomposing each into our three culturally grounded dimensions of Identity, Behavior, and Context to construct a balanced evaluation set across 10 geographic regions and 5 socio-cultural categories (see Appendix \ref{app:culture_atlas} for more details). Prompts were adapted using the gemini-3-flash-preview model \citep{gemini3flash2026} to ensure each prompt explicitly encodes geographic and cultural identifiers. Where a country name was absent from the original prompt, geographical adjectives were appended. For example, the prompt \emph{``Couple meeting at a German sports club gathering''} is decomposed as: \emph{German Couple} (Identity), \emph{German meeting} (Behavior), and \emph{German sports club gathering} (Context). Full prompt details are provided in Appendix section \ref{app:prompt}. We also provide details on how countries are similar to one another in Appendix section \ref{app:category_similarity}.

\subsection{Evaluation Framework}
Given these sourced cultural prompts, our approach consists of four main steps: (i) \textbf{Counterfactual prompt augmentation}, where systematic variations of cultural prompts are generated to probe model behavior; (ii) \textbf{video generation}, where augmented prompts are used to generate videos, (iii) \textbf{IBC decomposition}, where each identity, behavior, and context dimension are evaluated; and (iv) \textbf{VLM-based scoring}, where Vision-Language Models (VLMs) perform fine-grained, aspect-based question answering to quantify faithfulness in each dimension. Please refer to Figure \ref{fig:framework}.

\begin{figure*}[t]
    \centering
    \includegraphics[width=1\textwidth]{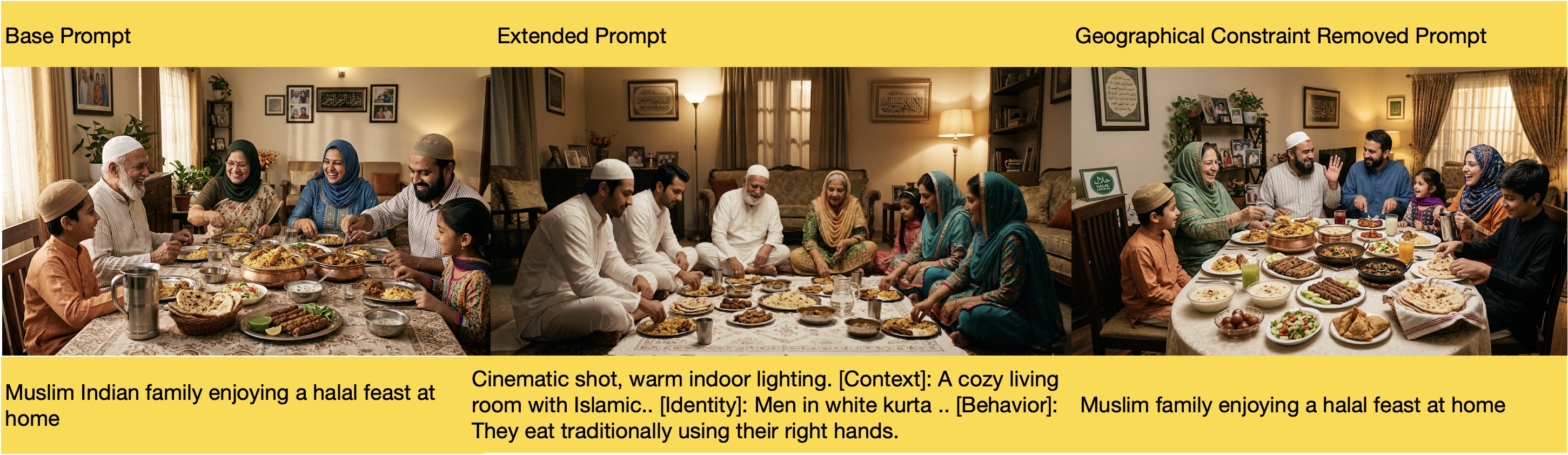}
    \caption{The \textbf{Base Prompt} (left) includes an explicit geographic identifier (\textit{``Muslim \textbf{Indian} family enjoying a halal feast at home''}), producing a culturally grounded scene. The \textbf{Extended Prompt} (center) augments the scene with decomposed Identity, Behavior, and Context descriptions, yielding richer cultural detail, such as Islamic d\'ecor, white kurta attire, etc. The \textbf{Geographical Constraint Removed Prompt} (right) strips the country identifier, retaining only \textit{``Muslim family enjoying a halal feast at home''}, to probe whether models have internalized cultural concepts independently of geographic triggers.}
    \label{fig:data_example}
\end{figure*}

\paragraph{Counterfactual prompt augmentation.}
To systematically probe how models respond to varying levels of cultural explicitness and to identify whether failures stem from missing explicit cues or absent implicit knowledge, we design three prompt variations for each cultural scenario.
\textbf{Base Prompt} includes the original prompt adapted from a previous work on culture \citep{nayak-etal-2025-culturalframes} that is verified by native people from that country. \textbf{Extended Prompt} expands the prompt into an explicit descriptive scene, mapping nuances into \textit{Identity}, \textit{Behavior}, and \textit{Context} by elaborating its meaning from the Oxford dictionary. The extended prompt is used in our evaluation framework. 
\textbf{Geographical Constraint Removed Prompt} removes specific country names (e.g., changing ``Muslim Indian family'' to ``Muslim family'') to test whether models have internalized cultural concepts as open-world knowledge or rely on geographic triggers to activate cultural associations.
An illustrative example of this strategy for a ``Halal Feast'' scenario is provided in Figure \ref{fig:data_example}. A sample of these prompts was evaluated by the native residents of the countries (refer to section \ref{sec:humanai}).

\paragraph{Video generation models.}
To ensure a comprehensive evaluation of cultural faithfulness, we select three state-of-the-art video generation models: \textbf{Wan 2.2}, \textbf{LTX-2}, and \textbf{Veo 3.1 Fast}. These models represent the current frontier in high-fidelity temporal synthesis and instruction following. For LTX-2 and Wan 2.2, we generate 5-second videos across three prompt variations: a base prompt, an extended prompt for structural guidance, and a version where the specific geographical country name is removed to test implicit cultural knowledge. For Veo 3.1 Fast, we test on a subset of ~10\% of the data because of the API costs and generate 4-second videos.   
More details on the model specifications are present in the Appendix section \ref{app:experiments}.

\paragraph{IBC decomposition.}
To evaluate the generated videos,
we use \texttt{gemini-3-flash-preview} \citep{gemini3flash2026} to automatically generate 9,288 culturally grounded questions and their corresponding ground-truth answers. The model is provided with human-validated prompts from Cultural Frames. We prompt the model to generate diverse reasoning pairs based on two key principles. First, we \textbf{embed visual descriptions}, where precise physical or spatial details are embedded directly into the question so the evaluator does not rely on implicit cultural knowledge. For example, instead of asking if a person is wearing a traditional Kimono, the model asks: Is the person wearing a traditional Kimono, characterized by left-over-right wrapped lapels and wide, square-cut sleeves?'' Second, we enforce \textbf{temporal grounding}, which explicitly probes the progression of movement across frames, such as: Does the pouring behavior begin with the vessel held low, smoothly rise to a higher elevation, and return low without breaking the stream?'' This approach ensures that the evaluation is rooted in specific, verifiable cultural markers rather than generic visual aesthetics. Details are in Appendix \ref{app:question_gen}. We also add a question verification stage that filters out irrelevant questions, leading to 8811 (94.87\%) valid questions; see Appendix \ref{app:question_verifier} for details. A sample of these verified questions was then annotated by native residents of the country to validate the relevance of the questions. Refer to section \ref{sec:humanai} for more details.

\paragraph{VLM-based scoring.} 
To evaluate the generated videos against our questions, we use \textbf{Qwen3-VL-235B-A22B-Instruct} \citep{bai2025qwen3}, a state-of-the-art vision-language model designed for complex visual understanding and temporal reasoning. Given answers generated by the model, \name\ averages model performance across the three decomposed dimensions of identity, behavior, and context. Prompt details present in Appendix section \ref{app:answer_generator}. Native residents of the country then annotated a sample of the dataset to validate answer accuracy. Refer to section \ref{sec:humanai} for more details.

\section{Experiments}

Our evaluation is guided by the following four research questions:

\textbf{RQ1:} To what extent do current state-of-the-art video generation models faithfully represent cultural Identity, Behavior, and Context, and does performance on existing perceptual quality metrics, such as VideoScore, align with or diverge from \name?

\textbf{RQ2:} Does providing decomposed, culturally explicit prompt guidance improve \name, and which dimensions benefit most?

\textbf{RQ3:} Do models rely on explicit geographic identifiers to activate cultural knowledge, or have they internalized cultural concepts as geography-independent representations?

\textbf{RQ4:} How do \name\ and VideoScore each align with the cultural preferences of native human evaluators, and which automated metric better reflects human judgment of cultural faithfulness?

\subsection{Results}

\begin{figure*}[t]
    \centering
    \includegraphics[width=0.8\textwidth]{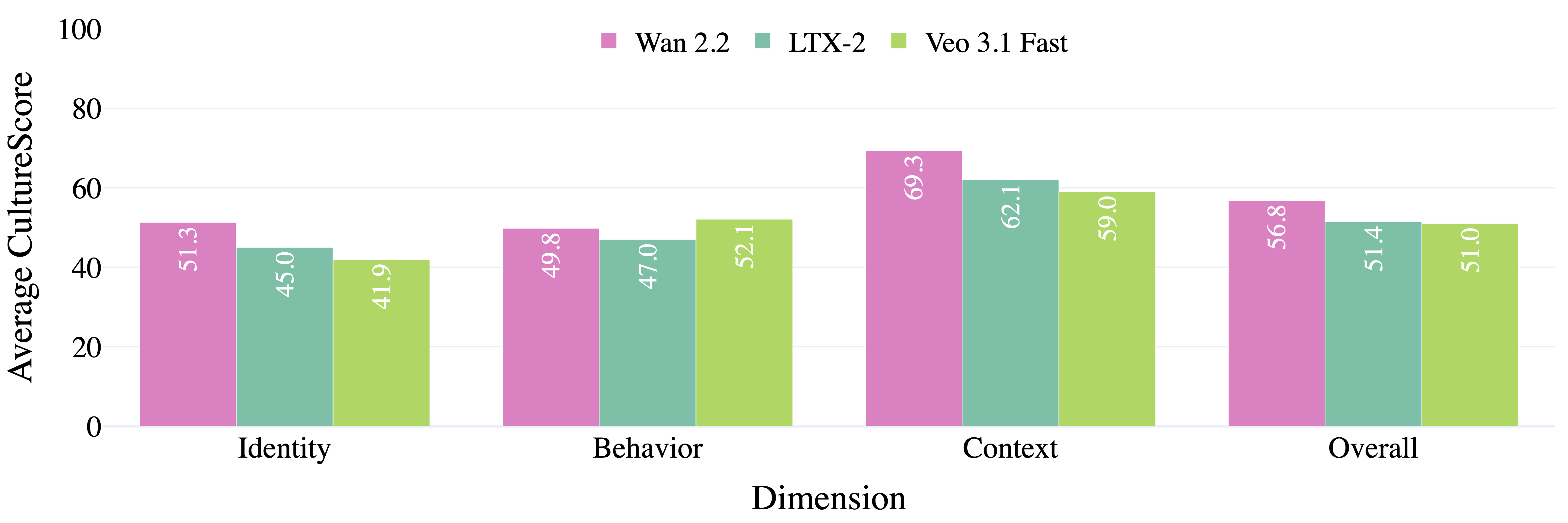}
    \caption{Average \name\ (\%) across Identity, Behavior, and Context dimensions for three video generation models under extended prompts. Context across all the models performs better than behavior and identity. Wan 2.2 outperforms the other two for Identity and Context.}
    \label{fig:Result_1}
\end{figure*}

\begin{figure*}[t]
    \centering
    \includegraphics[width=0.8\linewidth]{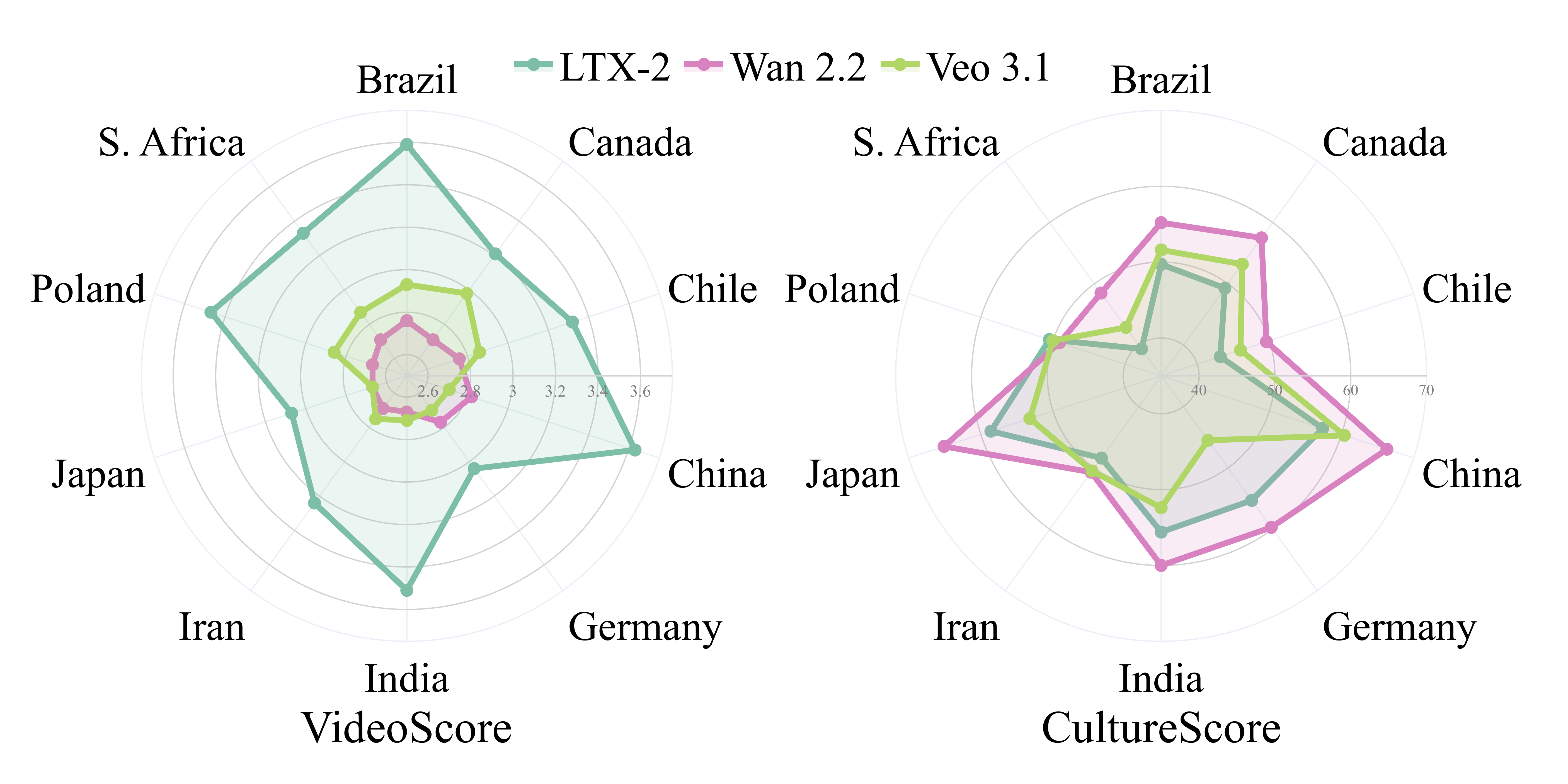}
    
    \caption{The inverse relationship between VideoScore and \name{}, averaged over all dimensions. The model perceived as most ``cinematic'' (LTX-2) does not lead on cultural faithfulness, while the most culturally accurate model (Wan 2.2) scores lowest on general quality metrics. Wan 2.2 leads on \name{} in 9 of 10 countries.}
    \label{fig:inverse_correlation}
\end{figure*}

For \textbf{RQ1}, we present results across two complementary views: (i) how \name\ varies across the Identity, Behavior, and Context dimensions under our evaluation framework (Figure~\ref{fig:Result_1}), and (ii) how VideoScore and \name\ diverge when compared across models and countries (Figure~\ref{fig:inverse_correlation}).

\paragraph{Dimension-level findings.} Context consistently yields the highest accuracy across all models: Wan 2.2 \citep{wan2025wanopenadvancedlargescale} achieves 69.3\%, LTX-2 \citep{hacohen2024ltxvideorealtimevideolatent} achieves 62.1\%, and Veo 3.1 Fast \citep{GoogleDeepMind2025Veo3} achieves 59.0\%.
Identity is the second-strongest dimension only for Wan 2.2 (51.3\%); for LTX-2 and Veo 3.1 Fast, Behavior ranks second (47.0\% and 52.1\%), with
Identity weakest for both (45.0\% and 41.9\%). Veo 3.1 Fast attains the
highest Behavior score of the three models, yet no model exceeds 52.1\% on
this dimension, well below the 69.3\% Context ceiling, indicating that culturally specific motion sequences represent a persistent failure mode that prompt enrichment alone cannot resolve. Refer to Figure \ref{fig:Result_1} for more details.

\paragraph{Inverse relationship between VideoScore vs.~\name.}  As shown in Figure~\ref{fig:inverse_correlation}, a striking and consistent pattern emerges across all three models. 
LTX-2 \citep{hacohen2024ltxvideorealtimevideolatent} secures the highest VideoScore (avg.~3.34 across the full suite) yet ranks second lowest in \name\ accuracy (avg. 51.4\%), with Veo 3.1 Fast scoring lowest (avg. 51.0\%). 
Wan 2.2 \citep{wan2025wanopenadvancedlargescale} achieves the highest cultural accuracy on average (avg.~56.8\%) while receiving the lowest VideoScore ratings (avg.~2.72), leading on \name\ in 9 of 10 countries, though only narrowly in Iran (50.7\% vs.\ 50.5\% for Veo 3.1 Fast \citep{GoogleDeepMind2025Veo3}).
The exception is Poland, where LTX-2 leads (50.5\%, vs.\ 50.0\% for Veo 3.1 Fast and 49.1\% for Wan 2.2), suggesting that optimizing for perceptual quality does not translate to cultural faithfulness.

\paragraph{Effect of extended prompt.} For \textbf{RQ2}, we compare \name\ under base prompts against extended prompts, which augment each scene with explicit Identity, Behavior, and Context descriptions derived from Oxford Dictionary definitions. This tests whether cultural failures stem from insufficient prompt specificity or from deeper gaps in model knowledge.

\begin{figure*}
    \centering
    \includegraphics[width=\linewidth]{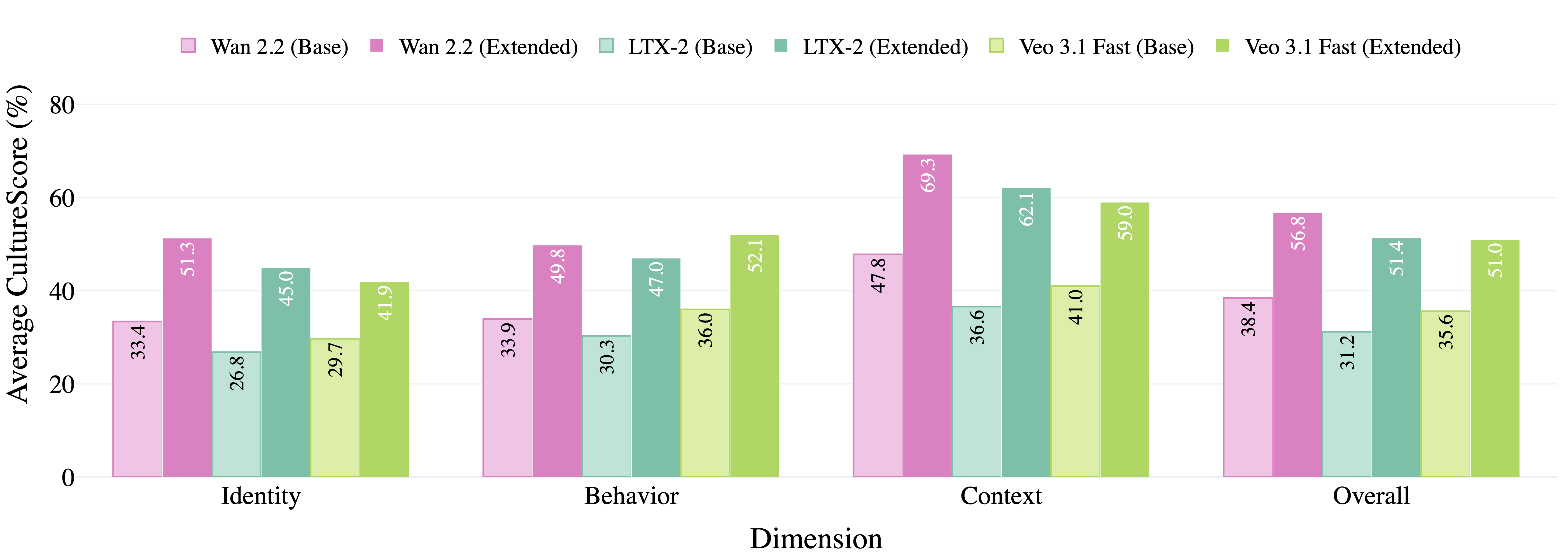}
    \vspace{-2mm}
    \caption{Average \name\ (\%) for base and extended prompts across the models. Extended prompts consistently outperform the base prompt across all dimensions, implying that extended prompting yields better output.}
    \label{fig:base_extended}
\end{figure*}

Extended prompts yield \textbf{consistent gains} across all three models (Veo 3.1 Fast \citep{GoogleDeepMind2025Veo3}, LTX-2 \citep{hacohen2024ltxvideorealtimevideolatent}, and Wan 2.2 \citep{wan2025wanopenadvancedlargescale}) and all three dimensions (Behavior, Identity, and Context), but the magnitude varies substantially by dimension. For Identity, LTX-2 benefits most (+18.2 percentage points
(pp), from 26.8\% to 45.0\%), closely followed by Wan 2.2 (+17.9 pp, from 33.4\% to 51.3\%) and Veo 3.1 Fast (+12.2 pp, from 29.7\% to 41.9\%).
The smaller gain for Veo 3.1 Fast suggests it already captures some object-level identity cues from base prompts. Refer to Figure \ref{fig:base_extended} and Appendix section \ref{app:experiments} Table \ref{tab:accuracy_variation_ltx} and \ref{tab:accuracy_variation_wan}. \textbf{The Context dimension shows the largest absolute gains across all models.} LTX-2 improves by +25.5 pp (36.6\% to 62.1\%), Wan 2.2 by +21.5 pp (47.8\% to 69.3\%), and Veo 3.1 Fast by +18.0 pp (41.0\% to 59.0\%), likely because scene-level descriptions (settings, decorations, spatial layouts) map more directly onto learnable visual features.

\paragraph{Behavior is the most resistant dimension to prompt enrichment.} Despite receiving the same structured guidance, behavior gains are comparable across all models: Wan 2.2 (+15.9 pp), Veo 3.1 Fast (+16.1 pp), and LTX-2 (+16.7 pp). Even under extended prompting, no model exceeds 52.1\% on Behavior, compared to 69.3\% on Context for the best-performing model (Wan 2.2), indicating that temporally coherent motion sequences remain a persistent failure mode that prompt enrichment alone cannot resolve.

We also evaluate a second VLM judge on extended prompts to test the sensitivity of our results to judge-model choice, and report approximately 82\% agreement between Qwen3-VL-235B-A22B \cite{bai2025qwen3} and
Gemini 2.5 Flash \cite{comanici2025gemini25pushingfrontier}. Gemini's judgments are released
publicly for reproducibility.

\begin{figure*}[t]
    \centering
    \includegraphics[width=1\linewidth]{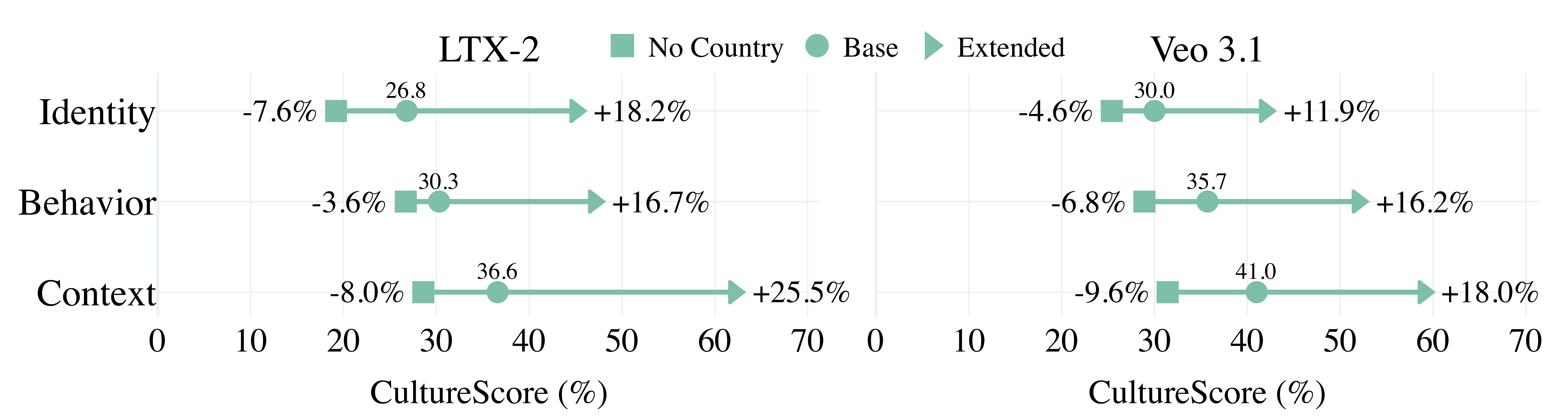}
    
    \caption{Average \name{} of LTX-2 and Veo 3.1 Fast across IBC dimensions with all three prompt variants, computed on the culturally unique subset (475 prompts for LTX-2, 43 for Veo 3.1 Fast) on which the geographical constraint removed condition is defined. Across all dimensions, the extended prompt performs the best, with the Context dimension and the LTX-2 model benefiting the most. The no country prompt performs worst throughout, and LTX-2 degrades more overall than Veo 3.1 Fast ($-6.3$ vs.\ $-4.7$ points on average). Relative to the base prompt, the dimension that suffers most differs by model: Identity for LTX-2 and Behavior for Veo 3.1 Fast, whose Identity score is nearly unchanged without an explicit geographic anchor.}
\label{fig:prompt_variants}
    \label{fig:placeholder}
\end{figure*}

\paragraph{Explicit vs. implicit cultural knowledge.} For \textbf{RQ3}, we study whether models have internalized cultural concepts as open-world knowledge or rely on geographic tokens as triggers. To answer this, we remove country names from base prompts while retaining cultural identifiers (e.g., ``Indian women greeting people with Namaste.'' to ``Women greeting people with Namaste''). We restrict this ablation to prompts whose scenes are distinctly tied to a specific culture, as judged by the classifier in Appendix~\ref{app: Geographical_constraint_verifier}, since prompts describing universal activities cannot reveal culture-specific failure. This yields 48.4\% of prompts (475 of 981) for LTX-2 and 44.8\% (43 of 96 sampled prompts) for Veo 3.1 Fast. All base, extended, and no-country scores in this analysis are computed on these matched subsets. Removing geographic identifiers causes a consistent and substantial drop.
For LTX-2, compared to extended prompts, removing geographic anchors causes accuracy to fall by 35.7 pp on Context (66.1\% $\rightarrow$ 30.4\%), 28.9 pp on Identity (47.9\% $\rightarrow$ 19.0\%), and 23.3 pp on Behavior (48.5\% $\rightarrow$ 25.2\%).
For Veo 3.1 Fast, compared to extended prompts, removing geographic anchors causes accuracy to fall by 36.0 pp on Context (67.2\% $\rightarrow$ 31.2\%), 22.3 pp on Behavior (50.4\% $\rightarrow$ 28.1\%), and 13.1 pp on Identity (36.5\% $\rightarrow$ 23.4\%).
This suggests that models rely heavily on explicit geographic tokens as cultural triggers rather than having internalized the underlying cultural concepts, and that cultural understanding in current video generation models remains largely surface-level.
The effect is most severe for Iranian and Chinese prompts, where LTX-2 shows the largest drops across all three dimensions (Iran: $-11.9$~pp avg, China: $-8.9$~pp avg) when geographic anchors are removed, suggesting that low-resource or visually distinctive cultures in training data are disproportionately affected. Notably, Canada and Chile show the smallest drops ($-1.6$~pp and $-1.3$~pp, respectively), possibly due to greater overlap with Western training data defaults.
Taken together, these findings reveal a consistent picture: cultural faithfulness in current video generation models is both prompt-sensitive and geographically dependent. Structured guidance improves context and identity considerably, but behavior remains stubbornly below the ceiling. We report the numbers in Figure~\ref{fig:prompt_variants}.

\subsection{Human Evaluation} \label{sec:humanai}

To validate whether the divergence between VideoScore and CultureScore reflects genuine human perception, we conducted a systematic human evaluation study on Prolific with N=45 native residents (5 per country) spanning 9 countries (all except Iran, given the current geopolitical situation). Each participant evaluated 3 video triplets (9 videos in total), yielding 135 rank observations per model. Participants were required to be verified current residents who also hold the nationality of the evaluated country on Prolific, ensuring cultural authenticity of judgments. Participants spanned diverse professional backgrounds including entrepreneurs, musicians, homemakers, policymakers, and data analysts, to avoid skew toward technical AI expertise.

For each triplet, participants completed a three-tier assessment: (i) \textbf{Question Relevance:} Judging whether the \name-generated evaluation question was a meaningful probe for the given cultural context, (ii) \textbf{Perceptual Accuracy:} Answering the Identity, Behavior, or Context question based strictly on visual evidence in the videos, and (iii) \textbf{Preference Ranking:} Ranking videos generated from all three models from 1st to 3rd based on overall cultural faithfulness.
Full annotation details are in Appendix~\ref{app:annotation}.

\begin{figure*}[t]
    \centering
    \includegraphics[width=0.8\textwidth]{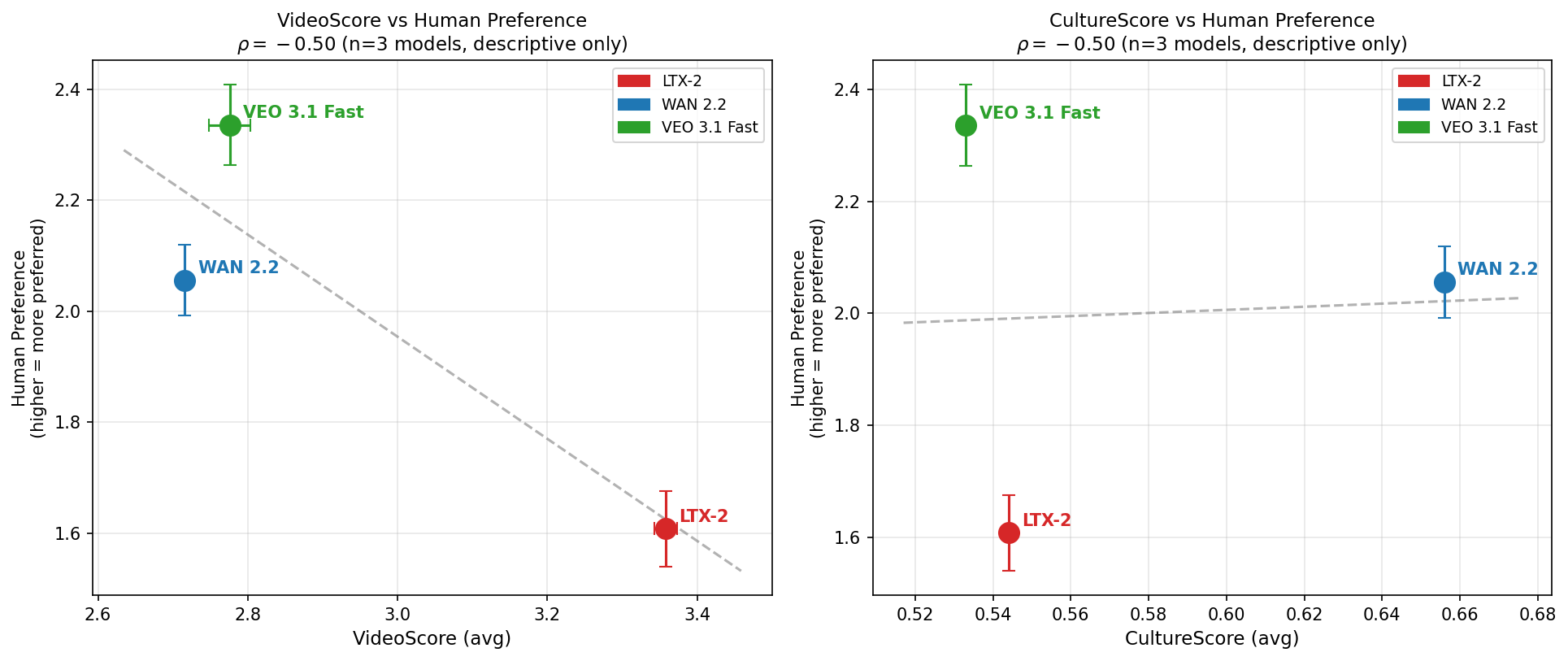}
    \vspace{-2mm}
    \caption {Correlation between \name{} and native human preference rankings, across all nine countries. VideoScore (left) is inversely correlated with human preference; the model receiving the highest video score (LTX-2, red) is consistently ranked lowest by human evaluators. \name{} (right) recovers the Wan~2.2 $>$ LTX-2 ordering that VideoScore reverses, with Wan~2.2 (blue) ranking highest on cultural accuracy; with $n = 3$ models both coefficients are descriptive only. Each point represents one model aggregated across all countries; error bars denote $\pm 1$ standard error.}
    \label{fig:Result_human_preference}
\end{figure*}

\paragraph{Inter-Annotator Agreement.}
Given the categorical nature of Question Relevance, we report Gwet's AC1~\citep{gwet2008computing} for question relevance, as it provides a more reliable agreement estimate under skewed marginal distributions. 
For Perceptual Accuracy, where annotators selected all videos for which a condition held true, we report average pairwise Jaccard similarity ~\citep{jaccard1912distribution} to account for the multi-label nature of responses. 
For preference ranking, we report Spearman's $\rho$ ~\citep{spearman1961proof} as a descriptive measure of ordering agreement between human preferences and evaluation metrics. Agreement scores for each task are reported below.

\paragraph{Question Relevance.} 
Across all nine countries, context and identity questions were judged highly relevant on average (84.2\% and 78.3\%, respectively), while behavior questions received notably lower ratings (69.0\%). Behavior relevance showed the greatest cross-cultural variability: German raters rated behavior questions most relevant (93.3\%), while Japanese raters gave the lowest rating (44.4\%). Indian raters showed a strong preference for context and identity (94.4\% each) over behavior (66.7\%), a pattern echoed in Japan (94.4\% context, 72.2\% identity vs.\ 44.4\% behavior). Inter-rater agreement was moderate on average (AC1\,$\approx 0.50$). Questions marked irrelevant by a majority of raters were excluded from further analysis.

\paragraph{Perceptual Accuracy.}
Identity questions yielded the highest answer accuracy across all nine countries (avg.\ 63.0\%), while behavior and context questions were comparably harder (53.7\% and 51.2\%, respectively), consistent with behavior's role as the weakest dimension in our automated evaluation. Identity accuracy was especially high for German and Indian raters (88.9\% each), whereas Canadian and Japanese raters found identity the most difficult (33.3\% and 44.4\%). Behavior accuracy showed the greatest cross-cultural variability, ranging from 33.3\% (Brazil, South Africa) to 77.8\% (Germany). Context accuracy diverged notably across cohorts: Indian and Polish raters scored highest (66.7\% each), whereas Brazilian and Chinese raters scored the lowest (33.3\% and 44.4\%). 
Since annotators selected a set of videos per question, we report inter-rater agreement via average pairwise Jaccard similarity; agreement was moderate overall ($J = 0.63$), ranging from $J = 0.47$ (South Africa) to $J = 0.78$ (India).

\paragraph{Preference Ranking and Metric Correlation.}

\noindent
For each prompt, annotators watched three AI-generated videos side-by-side and assigned a rank to each model (1 = most preferred, 3 = least preferred). We aggregate preferences by averaging per-model ranks across all raters and prompts within each country, then pooling across all nine countries ($n = 135$ rank observations per model). Human preference followed a clear ordering: VEO~3.1~Fast was most preferred (avg.\ rank $1.66 \pm 0.07$), followed by Wan~2.2 ($1.94 \pm 0.06$), and LTX-2 was ranked lowest ($2.39 \pm 0.07$).

\noindent
To assess whether automated metrics align with this ordering, we compare each model's aggregate \name\ and VideoScore against its average human preference rank. \name\ here is computed as the fraction of culturally grounded evaluation questions for which the model's generated answer matches the majority human ground truth; VideoScore is the average of five perceptual dimensions ~\citep{he2024videoscore}.

\noindent
\textbf{VideoScore} does not reflect human preference: LTX-2 received the highest VideoScore ($3.36$ on the subset) yet was ranked last by annotators, while Veo~3.1~Fast and Wan~2.2 scored substantially lower ($2.78$ and $2.72$, respectively) but were preferred by native human annotators ($\rho = -0.50$). This inversion confirms that perceptual quality metrics can actively mislead model selection. \textbf{\name}, by contrast, rewards Wan~2.2 most highly on the human-evaluation subset ($0.656$, ground truth determined by majority annotator vote across $n=27$ prompts), consistent with its second-place human preference ranking. Together, these results indicate that general-purpose video quality metrics are not merely insufficient for measuring cultural faithfulness, but can actively mislead model selection when cultural accuracy is the priority. We also conducted an additional experiment to compare \name\ with unified reward \cite{wang2025unified} and clipscore \cite{hessel2021clipscore} with human judgement. Each generated video was reduced to skip the blank or under-formed opening frames common in generated videos. More details can be found in the Appendix section \ref{app:comparison}.

\section{Conclusion}

We presented \name{}, a compositional framework that decomposes cultural faithfulness in video generation into Identity, Behavior, and Context dimensions. Across 2,943 prompts spanning 10 countries and 5 socio-cultural domains, existing quality metrics such as VideoScore actively mislead cultural evaluation: the model ranked highest on visual quality (LTX-2) was ranked lowest on cultural faithfulness by native human evaluators. Decomposed cultural guidance yields meaningful gains, with Context most responsive to prompt enrichment and Behavior the most persistent failure mode. 
Behavior shows the strongest correlation with human cultural preference, suggesting that culturally normative motion is what audiences notice first. We hope \name{} provides a reusable foundation for auditing cultural representation in generative AI.

\section*{Limitations} \label{app:limitations}

Our evaluation has three primary limitations.

\paragraph{Asymmetric model coverage.} Due to the high operational costs of the 
Veo~3.1 Fast API, we evaluate a stratified sample of 288 videos rather than the 
full 2,943-prompt suite used for Wan~2.2 and LTX-2. While the sample is designed 
to maintain cross-cultural and cross-category coverage, direct comparisons between 
Veo~3.1 and the other two models should be interpreted with this asymmetry in mind.

\paragraph{Geographic scope of human evaluation.} Our human evaluation is limited 
to native annotators from all the countries except Iran given the current geopolitical situation. We aim to include it in the future.

\paragraph{VLM-as-judge biases.} \name\ relies on Qwen3-VL to evaluate generated 
videos against culturally grounded questions. Although we include a question 
verification stage to reduce hallucinated or inaccurate evaluation criteria, the 
scoring model may itself carry cultural biases that systematically favor or penalize 
certain representations. We recommend treating \name\ as a complementary metric 
to be used alongside human evaluation, rather than as a standalone ground truth.

\section*{Ethics Statement}

This work involves human evaluation of culturally sensitive video content. Annotators were recruited from Prolific as native members of the cultures being evaluated and participated voluntarily. No personally identifiable information was collected during the annotation process.
The cultural prompts used in this study are grounded in the Cultural Atlas \citep{cultural_atlas_2016}, a publicly available resource developed collaboratively with community members. We acknowledge that cultural representation is inherently contested and that no benchmark can fully capture the diversity within any national or regional group. \name\ reflects cultural norms as documented in reference materials and validated by native evaluators; it should not be interpreted as a definitive or exhaustive account of any culture.

We further note that certain populations are at particular risk of under-evaluation in \name. Minority communities within a country, immigrant and diasporic groups whose practices diverge from those documented in the country of origin, and Indigenous communities whose cultures are not adequately covered by national framings may be underrepresented. Agreement scores and evaluation results should therefore be read as reflecting dominant or majority cultural norms within each country, rather than the full range of practices present in it. Extending \name\ to
these populations would require targeted recruitment, which we leave to future work.

We also acknowledge that the VLMs used in our automated evaluation pipeline, including Qwen3-VL and Gemini Flash, may themselves carry cultural biases that influence scoring. Relatedly, our operationalization of the Identity dimension, which is grounded in physical appearance, attire, and demographic markers, risks reinforcing stereotypes. In particular, Identity is a deeply complex concept extending well beyond visual surface features, encompassing language, lived experience, class, and history that a generated video cannot fully capture. Our choice to evaluate Identity through visually legible markers is a pragmatic constraint of the video evaluation setting, not a theoretical claim about what identity means.

Finally, we consider the dual-use potential of this work. While \name\ is intended to surface cultural fidelity gaps, the same signal could be optimized against to produce superficial or stereotyped cultural markers, or used to make generated content more persuasive for culturally
targeted advertising, propaganda, or misleading media. We release \name\ as a diagnostic benchmark rather than an optimization objective or a certification of cultural authenticity. We recommend that \name\ be used alongside, rather than as a replacement for, human judgment, particularly when evaluation findings are used to make claims about specific communities. The dataset and evaluation suite are released publicly to enable replication and extension by the broader research community.

\section*{LLM Usage Disclosure}
We used LLMs for minor writing assistance, including grammar correction and language
polishing. The core research
ideas, methodology, experimental design, implementation, analysis, and conclusions were developed and carried out by the authors. No LLM was used to generate research ideas, experimental analysis, figures, or evaluation results, beyond the models under study.

\section*{Acknowledgments}
The authors would like to thank the Media Lab Consortium and the MIT Tata Center Technology and Design Fellowship for supporting the
work. 

\bibliography{custom}

@inproceedings{nayak-etal-2025-culturalframes,
    title = "{C}ultural{F}rames: Assessing Cultural Expectation Alignment in Text-to-Image Models and Evaluation Metrics",
    author = "Nayak, Shravan  and
      Bhatia, Mehar  and
      Zhang, Xiaofeng  and
      Rieser, Verena  and
      Hendricks, Lisa Anne  and
      Steenkiste, Sjoerd Van  and
      Goyal, Yash  and
      Stanczak, Karolina  and
      Agrawal, Aishwarya",
    editor = "Christodoulopoulos, Christos  and
      Chakraborty, Tanmoy  and
      Rose, Carolyn  and
      Peng, Violet",
    booktitle = "Findings of the Association for Computational Linguistics: EMNLP 2025",
    month = nov,
    year = "2025",
    address = "Suzhou, China",
    publisher = "Association for Computational Linguistics",
    url = "https://aclanthology.org/2025.findings-emnlp.1141/",
    doi = "10.18653/v1/2025.findings-emnlp.1141",
    pages = "20918--20953",
    ISBN = "979-8-89176-335-7"
}

@inproceedings{khanuja-etal-2024-image,
    title = "An image speaks a thousand words, but can everyone listen? On image transcreation for cultural relevance",
    author = "Khanuja, Simran  and
      Ramamoorthy, Sathyanarayanan  and
      Song, Yueqi  and
      Neubig, Graham",
    editor = "Al-Onaizan, Yaser  and
      Bansal, Mohit  and
      Chen, Yun-Nung",
    booktitle = "Proceedings of the 2024 Conference on Empirical Methods in Natural Language Processing",
    month = nov,
    year = "2024",
    address = "Miami, Florida, USA",
    publisher = "Association for Computational Linguistics",
    url = "https://aclanthology.org/2024.emnlp-main.573/",
    doi = "10.18653/v1/2024.emnlp-main.573",
    pages = "10258--10279"
}

@inproceedings{he2024videoscore,
  title={Videoscore: Building automatic metrics to simulate fine-grained human feedback for video generation},
  author={He, Xuan and Jiang, Dongfu and Zhang, Ge and Ku, Max and Soni, Achint and Siu, Sherman and Chen, Haonan and Chandra, Abhranil and Jiang, Ziyan and Arulraj, Aaran and others},
  booktitle={Proceedings of the 2024 Conference on Empirical Methods in Natural Language Processing},
  pages={2105--2123},
  year={2024}
}

@article{wang2025unified,
  title={Unified reward model for multimodal understanding and generation},
  author={Wang, Yibin and Zang, Yuhang and Li, Hao and Jin, Cheng and Wang, Jiaqi},
  journal={arXiv preprint arXiv:2503.05236},
  year={2025}
}

@inproceedings{ray2018scenes,
  title={Scenes-objects-actions: A multi-task, multi-label video dataset},
  author={Ray, Jamie and Wang, Heng and Tran, Du and Wang, Yufei and Feiszli, Matt and Torresani, Lorenzo and Paluri, Manohar},
  booktitle={Proceedings of the European conference on computer vision (ECCV)},
  pages={635--651},
  year={2018}
}

@inproceedings{wang2024recipe,
  title={A recipe for scaling up text-to-video generation with text-free videos},
  author={Wang, Xiang and Zhang, Shiwei and Yuan, Hangjie and Qing, Zhiwu and Gong, Biao and Zhang, Yingya and Shen, Yujun and Gao, Changxin and Sang, Nong},
  booktitle={Proceedings of the IEEE/CVF Conference on Computer Vision and Pattern Recognition},
  pages={6572--6582},
  year={2024}
}

@inproceedings{lee2024grid,
  title={Grid diffusion models for text-to-video generation},
  author={Lee, Taegyeong and Kwon, Soyeong and Kim, Taehwan},
  booktitle={Proceedings of the IEEE/CVF Conference on Computer Vision and Pattern Recognition},
  pages={8734--8743},
  year={2024}
}

@misc{peebles2023scalablediffusionmodelstransformers,
      title={Scalable Diffusion Models with Transformers}, 
      author={William Peebles and Saining Xie},
      year={2023},
      eprint={2212.09748},
      archivePrefix={arXiv},
      primaryClass={cs.CV},
      url={https://arxiv.org/abs/2212.09748}, 
}

@misc{openai2024videoworldsimulators,
  author = {{OpenAI}},
  title  = {Video generation models as world simulators},
  year   = {2024},
  month  = {February},
  url    = {https://openai.com/index/video-generation-models-as-world-simulators/}
}

@inproceedings{yang2025cogvideox,
    title={CogVideoX: Text-to-Video Diffusion Models with An Expert Transformer},
    author={Zhuoyi Yang and Jiayan Teng and Wendi Zheng and Ming Ding and Shiyu Huang and Jiazheng Xu and Yuanming Yang and Wenyi Hong and Xiaohan Zhang and Guanyu Feng and Da Yin and Yuxuan Zhang and Weihan Wang and Yean Cheng and Bin Xu and Xiaotao Gu and Yuxiao Dong and Jie Tang},
    booktitle={The Thirteenth International Conference on Learning Representations},
    year={2025},
    url={https://openreview.net/forum?id=LQzN6TRFg9}
}

@misc{kong2025hunyuanvideosystematicframeworklarge,
      title={HunyuanVideo: A Systematic Framework For Large Video Generative Models}, 
      author={Weijie Kong and Qi Tian and Zijian Zhang and Rox Min and Zuozhuo Dai and Jin Zhou and Jiangfeng Xiong and Xin Li and Bo Wu and Jianwei Zhang and Kathrina Wu and Qin Lin and Junkun Yuan and Yanxin Long and Aladdin Wang and Andong Wang and Changlin Li and Duojun Huang and Fang Yang and Hao Tan and Hongmei Wang and Jacob Song and Jiawang Bai and Jianbing Wu and Jinbao Xue and Joey Wang and Kai Wang and Mengyang Liu and Pengyu Li and Shuai Li and Weiyan Wang and Wenqing Yu and Xinchi Deng and Yang Li and Yi Chen and Yutao Cui and Yuanbo Peng and Zhentao Yu and Zhiyu He and Zhiyong Xu and Zixiang Zhou and Zunnan Xu and Yangyu Tao and Qinglin Lu and Songtao Liu and Dax Zhou and Hongfa Wang and Yong Yang and Di Wang and Yuhong Liu and Jie Jiang and Caesar Zhong},
      year={2025},
      eprint={2412.03603},
      archivePrefix={arXiv},
      primaryClass={cs.CV},
      url={https://arxiv.org/abs/2412.03603}, 
}

@misc{hacohen2024ltxvideorealtimevideolatent,
      title={LTX-Video: Realtime Video Latent Diffusion}, 
      author={Yoav HaCohen and Nisan Chiprut and Benny Brazowski and Daniel Shalem and Dudu Moshe and Eitan Richardson and Eran Levin and Guy Shiran and Nir Zabari and Ori Gordon and Poriya Panet and Sapir Weissbuch and Victor Kulikov and Yaki Bitterman and Zeev Melumian and Ofir Bibi},
      year={2024},
      eprint={2501.00103},
      archivePrefix={arXiv},
      primaryClass={cs.CV},
      url={https://arxiv.org/abs/2501.00103}, 
}

@inproceedings{
singer2023makeavideo,
title={Make-A-Video: Text-to-Video Generation without Text-Video Data},
author={Uriel Singer and Adam Polyak and Thomas Hayes and Xi Yin and Jie An and Songyang Zhang and Qiyuan Hu and Harry Yang and Oron Ashual and Oran Gafni and Devi Parikh and Sonal Gupta and Yaniv Taigman},
booktitle={The Eleventh International Conference on Learning Representations },
year={2023},
url={https://openreview.net/forum?id=nJfylDvgzlq}
}

@techreport{GoogleDeepMind2025Veo3,
  author      = {{Google DeepMind}},
  title       = {Veo: a text-to-video generation system},
  year        = {2025},
  url         = {https://storage.googleapis.com/deepmind-media/veo/Veo-3-Tech-Report.pdf},
  type        = {Technical Report},
  note        = {Accessed: 2026-03-31}
}

@online{Hume2025Flow,
  author       = {Tom Hume and Matthew Carey and Thomas Iljic},
  title        = {Meet {Flow}: {AI}-powered filmmaking with {Veo 3}},
  year         = {2025},
  month        = {May},
  day          = {20},
  publisher    = {The Keyword (Google Blog)},
  url          = {https://blog.google/innovation-and-ai/products/google-flow-veo-ai-filmmaking-tool/}
}

@misc{rege2025cureculturalgapslong,
      title={CuRe: Cultural Gaps in the Long Tail of Text-to-Image Systems}, 
      author={Aniket Rege and Zinnia Nie and Mahesh Ramesh and Unmesh Raskar and Zhuoran Yu and Aditya Kusupati and Yong Jae Lee and Ramya Korlakai Vinayak},
      year={2025},
      eprint={2506.08071},
      archivePrefix={arXiv},
      primaryClass={cs.CV},
      url={https://arxiv.org/abs/2506.08071}, 
}

@inproceedings{kannenbeyondaesthetics,
author = {Kannen, Nithish and Ahmad, Arif and Andreetto, Marco and Prabhakaran, Vinodkumar and Prabhu, Utsav and Dieng, Adji Bousso and Bhattacharyya, Pushpak and Dave, Shachi},
title = {Beyond aesthetics: cultural competence in text-to-image models},
year = {2024},
isbn = {9798331314385},
publisher = {Curran Associates Inc.},
address = {Red Hook, NY, USA},
booktitle = {Proceedings of the 38th International Conference on Neural Information Processing Systems},
articleno = {439},
numpages = {32},
location = {Vancouver, BC, Canada},
series = {NIPS '24}
}

@misc{bayramli2025diffusionmodelsgloballens,
      title={Diffusion Models Through a Global Lens: Are They Culturally Inclusive?}, 
      author={Zahra Bayramli and Ayhan Suleymanzade and Na Min An and Huzama Ahmad and Eunsu Kim and Junyeong Park and James Thorne and Alice Oh},
      year={2025},
      eprint={2502.08914},
      archivePrefix={arXiv},
      primaryClass={cs.CV},
      url={https://arxiv.org/abs/2502.08914}, 
}

@inproceedings{
malakouti2026culture,
title={Culture in Action: Evaluating Text-to-Image Models through Social Activities},
author={Sina Malakouti and Boqing Gong and Adriana Kovashka},
booktitle={The Fourteenth International Conference on Learning Representations},
year={2026},
url={https://openreview.net/forum?id=opG4m2U0Oo}
}

@misc{cultural_atlas_2016,
  author = {{Cultural Atlas}},
  title = {About the Cultural Atlas},
  year = {2016},
  url = {https://culturalatlas.sbs.com.au/},
  note = {Accessed: 2026-03-31},
  publisher = {SBS, Mosaica \& Multicultural NSW}
}

@inproceedings{reimers-2019-sentence-bert,
    title = "Sentence-BERT: Sentence Embeddings using Siamese BERT-Networks",
    author = "Reimers, Nils and Gurevych, Iryna",
    booktitle = "Proceedings of the 2019 Conference on Empirical Methods in Natural Language Processing",
    month = "11",
    year = "2019",
    publisher = "Association for Computational Linguistics",
    url = "https://arxiv.org/abs/1908.10084",
}

@misc{gemini3flash2026,
  author       = {Google},
  title        = {Gemini 3 Flash Preview},
  year         = {2026},
  howpublished = {\url{https://ai.google.dev/gemini-api/docs/models/gemini-3-flash-preview}},
  note         = {Large language model; accessed March 31, 2026}
}

@misc{chiu2025culturalbenchrobustdiversechallenging,
      title={CulturalBench: A Robust, Diverse, and Challenging Cultural Benchmark by Human-AI CulturalTeaming}, 
      author={Yu Ying Chiu and Liwei Jiang and Bill Yuchen Lin and Chan Young Park and Shuyue Stella Li and Sahithya Ravi and Mehar Bhatia and Maria Antoniak and Yulia Tsvetkov and Vered Shwartz and Yejin Choi},
      year={2025},
      eprint={2410.02677},
      archivePrefix={arXiv},
      primaryClass={cs.CL},
      url={https://arxiv.org/abs/2410.02677}, 
}

@inproceedings{nayak-etal-2024-benchmarking,
    title = "Benchmarking Vision Language Models for Cultural Understanding",
    author = "Nayak, Shravan  and
      Jain, Kanishk  and
      Awal, Rabiul  and
      Reddy, Siva  and
      Steenkiste, Sjoerd Van  and
      Hendricks, Lisa Anne  and
      Stanczak, Karolina  and
      Agrawal, Aishwarya",
    editor = "Al-Onaizan, Yaser  and
      Bansal, Mohit  and
      Chen, Yun-Nung",
    booktitle = "Proceedings of the 2024 Conference on Empirical Methods in Natural Language Processing",
    month = nov,
    year = "2024",
    address = "Miami, Florida, USA",
    publisher = "Association for Computational Linguistics",
    url = "https://aclanthology.org/2024.emnlp-main.329/",
    doi = "10.18653/v1/2024.emnlp-main.329",
    pages = "5769--5790"
}

@inproceedings{bhatia-etal-2024-local,
    title = "From Local Concepts to Universals: Evaluating the Multicultural Understanding of Vision-Language Models",
    author = "Bhatia, Mehar  and
      Ravi, Sahithya  and
      Chinchure, Aditya  and
      Hwang, EunJeong  and
      Shwartz, Vered",
    editor = "Al-Onaizan, Yaser  and
      Bansal, Mohit  and
      Chen, Yun-Nung",
    booktitle = "Proceedings of the 2024 Conference on Empirical Methods in Natural Language Processing",
    month = nov,
    year = "2024",
    address = "Miami, Florida, USA",
    publisher = "Association for Computational Linguistics",
    url = "https://aclanthology.org/2024.emnlp-main.385/",
    doi = "10.18653/v1/2024.emnlp-main.385",
    pages = "6763--6782"
}

@inproceedings{huang2024vbench,
  title={Vbench: Comprehensive benchmark suite for video generative models},
  author={Huang, Ziqi and He, Yinan and Yu, Jiashuo and Zhang, Fan and Si, Chenyang and Jiang, Yuming and Zhang, Yuanhan and Wu, Tianxing and Jin, Qingyang and Chanpaisit, Nattapol and others},
  booktitle={Proceedings of the IEEE/CVF Conference on Computer Vision and Pattern Recognition},
  pages={21807--21818},
  year={2024}
}

@inproceedings{song-etal-2025-vf,
    title = "{VF}-Eval: Evaluating Multimodal {LLM}s for Generating Feedback on {AIGC} Videos",
    author = "Song, Tingyu  and
      Hu, Tongyan  and
      Gan, Guo  and
      Zhao, Yilun",
    editor = "Che, Wanxiang  and
      Nabende, Joyce  and
      Shutova, Ekaterina  and
      Pilehvar, Mohammad Taher",
    booktitle = "Proceedings of the 63rd Annual Meeting of the Association for Computational Linguistics (Volume 1: Long Papers)",
    month = jul,
    year = "2025",
    address = "Vienna, Austria",
    publisher = "Association for Computational Linguistics",
    url = "https://aclanthology.org/2025.acl-long.1027/",
    doi = "10.18653/v1/2025.acl-long.1027",
    pages = "21126--21146",
    ISBN = "979-8-89176-251-0"
}

@misc{huang2024vbenchcomprehensiveversatilebenchmark,
      title={VBench++: Comprehensive and Versatile Benchmark Suite for Video Generative Models}, 
      author={Ziqi Huang and Fan Zhang and Xiaojie Xu and Yinan He and Jiashuo Yu and Ziyue Dong and Qianli Ma and Nattapol Chanpaisit and Chenyang Si and Yuming Jiang and Yaohui Wang and Xinyuan Chen and Ying-Cong Chen and Limin Wang and Dahua Lin and Yu Qiao and Ziwei Liu},
      year={2024},
      eprint={2411.13503},
      archivePrefix={arXiv},
      primaryClass={cs.CV},
      url={https://arxiv.org/abs/2411.13503}, 
}

@misc{ku2024viescoreexplainablemetricsconditional,
      title={VIEScore: Towards Explainable Metrics for Conditional Image Synthesis Evaluation}, 
      author={Max Ku and Dongfu Jiang and Cong Wei and Xiang Yue and Wenhu Chen},
      year={2024},
      eprint={2312.14867},
      archivePrefix={arXiv},
      primaryClass={cs.CV},
      url={https://arxiv.org/abs/2312.14867}, 
}

@misc{unterthiner2019accurategenerativemodelsvideo,
      title={Towards Accurate Generative Models of Video: A New Metric and Challenges}, 
      author={Thomas Unterthiner and Sjoerd van Steenkiste and Karol Kurach and Raphael Marinier and Marcin Michalski and Sylvain Gelly},
      year={2019},
      eprint={1812.01717},
      archivePrefix={arXiv},
      primaryClass={cs.CV},
      url={https://arxiv.org/abs/1812.01717}, 
}

@article{gwet2008computing,
  title={Computing inter-rater reliability and its variance in the presence of high agreement},
  author={Gwet, Kilem Li},
  journal={British Journal of Mathematical and Statistical Psychology},
  volume={61},
  number={1},
  pages={29--48},
  year={2008},
  publisher={Wiley Online Library}
}

@article{jaccard1912distribution,
  title={The distribution of the flora in the alpine zone. 1},
  author={Jaccard, Paul},
  journal={New phytologist},
  volume={11},
  number={2},
  pages={37--50},
  year={1912},
  publisher={Wiley Online Library}
}

@article{spearman1961proof,
  title={The proof and measurement of association between two things.},
  author={Spearman, Charles},
  year={1961},
  publisher={Appleton-Century-Crofts}
}

@misc{comanici2025gemini25pushingfrontier,
      title={Gemini 2.5: Pushing the Frontier with Advanced Reasoning, Multimodality, Long Context, and Next Generation Agentic Capabilities}, 
      author={Gheorghe Comanici and Eric Bieber and Mike Schaekermann and Ice Pasupat and Noveen Sachdeva and Inderjit Dhillon and Marcel Blistein and Ori Ram and Dan Zhang and Evan Rosen and Luke Marris and Sam Petulla and Colin Gaffney and Asaf Aharoni and Nathan Lintz and Tiago Cardal Pais and Henrik Jacobsson and Idan Szpektor and Nan-Jiang Jiang and Krishna Haridasan and Ahmed Omran and Nikunj Saunshi and Dara Bahri and Gaurav Mishra and Eric Chu and Toby Boyd and Brad Hekman and Aaron Parisi and Chaoyi Zhang and Kornraphop Kawintiranon and Tania Bedrax-Weiss and Oliver Wang and Ya Xu and Ollie Purkiss and Uri Mendlovic and Ilaï Deutel and Nam Nguyen and Adam Langley and Flip Korn and Lucia Rossazza and Alexandre Ramé and Sagar Waghmare and Helen Miller and Nathan Byrd and Ashrith Sheshan and Raia Hadsell and Sangnie Bhardwaj and Pawel Janus and Tero Rissa and Dan Horgan and Alvin Abdagic and Lior Belenki and James Allingham and Anima Singh and Theo Guidroz and Srivatsan Srinivasan and Herman Schmit and Kristen Chiafullo and Andre Elisseeff and Nilpa Jha and Prateek Kolhar and Leonard Berrada and Frank Ding and Xiance Si and Shrestha Basu Mallick and Franz Och and Sofia Erell and Eric Ni and Tejasi Latkar and Sherry Yang and Petar Sirkovic and Ziqiang Feng and Robert Leland and Rachel Hornung and Gang Wu and Charles Blundell and Hamidreza Alvari and Po-Sen Huang and Cathy Yip and Sanja Deur and Li Liu and Gabriela Surita and Pablo Duque and Dima Damen and Johnson Jia and Arthur Guez and Markus Mircea and Animesh Sinha and Alberto Magni and Paweł Stradomski and Tal Marian and Vlado Galić and Wenhu Chen and Hisham Husain and Achintya Singhal and Dominik Grewe and François-Xavier Aubet and Shuang Song and Lorenzo Blanco and Leland Rechis and Lewis Ho and Rich Munoz and Kelvin Zheng and Jessica Hamrick and Kevin Mather and Hagai Taitelbaum and Eliza Rutherford and Yun Lei and Kuangyuan Chen and Anand Shukla and Erica Moreira and Eric Doi and Berivan Isik and Nir Shabat and Dominika Rogozińska and Kashyap Kolipaka and Jason Chang and Eugen Vušak and Srinivasan Venkatachary and Shadi Noghabi and Tarun Bharti and Younghoon Jun and Aleksandr Zaks and Simon Green and Jeshwanth Challagundla and William Wong and Muqthar Mohammad and Dean Hirsch and Yong Cheng and Iftekhar Naim and Lev Proleev and Damien Vincent and Aayush Singh and Maxim Krikun and Dilip Krishnan and Zoubin Ghahramani and Aviel Atias and Rajeev Aggarwal and Christo Kirov and Dimitrios Vytiniotis and Christy Koh and Alexandra Chronopoulou and Pawan Dogra and Vlad-Doru Ion and Gladys Tyen and Jason Lee and Felix Weissenberger and Trevor Strohman and Ashwin Balakrishna and Jack Rae and Marko Velic and Raoul de Liedekerke and Oded Elyada and Wentao Yuan and Canoee Liu and Lior Shani and Sergey Kishchenko and Bea Alessio and Yandong Li and Richard Song and Sam Kwei and Orion Jankowski and Aneesh Pappu and Youhei Namiki and Yenai Ma and Nilesh Tripuraneni and Colin Cherry and Marissa Ikonomidis and Yu-Cheng Ling and Colin Ji and Beka Westberg and Auriel Wright and Da Yu and David Parkinson and Swaroop Ramaswamy and Jerome Connor and Soheil Hassas Yeganeh and Snchit Grover and George Kenwright and Lubo Litchev and Chris Apps and Alex Tomala and Felix Halim and Alex Castro-Ros and Zefei Li and Anudhyan Boral and Pauline Sho and Michal Yarom and Eric Malmi and David Klinghoffer and Rebecca Lin and Alan Ansell and Pradeep Kumar S and Shubin Zhao and Siqi Zuo and Adam Santoro and Heng-Tze Cheng and Solomon Demmessie and Yuchi Liu and Nicole Brichtova and Allie Culp and Nathaniel Braun and Dan Graur and Will Ng and Nikhil Mehta and Aaron Phillips and Patrik Sundberg and Varun Godbole and Fangyu Liu and Yash Katariya and David Rim and Mojtaba Seyedhosseini and Sean Ammirati and Jonas Valfridsson and Mahan Malihi and Timothy Knight and Andeep Toor and Thomas Lampe and Abe Ittycheriah and Lewis Chiang and Chak Yeung and Alexandre Fréchette and Jinmeng Rao and Huisheng Wang and Himanshu Srivastava and Richard Zhang and Rocky Rhodes and Ariel Brand and Dean Weesner and Ilya Figotin and Felix Gimeno and Rachana Fellinger and Pierre Marcenac and José Leal and Eyal Marcus and Victor Cotruta and Rodrigo Cabrera and Sheryl Luo and Dan Garrette and Vera Axelrod and Sorin Baltateanu and David Barker and Dongkai Chen and Horia Toma and Ben Ingram and Jason Riesa and Chinmay Kulkarni and Yujing Zhang and Hongbin Liu and Chao Wang and Martin Polacek and Will Wu and Kai Hui and Adrian N Reyes and Yi Su and Megan Barnes and Ishaan Malhi and Anfal Siddiqui and Qixuan Feng and Mihai Damaschin and Daniele Pighin and Andreas Steiner and Samuel Yang and Ramya Sree Boppana and Simeon Ivanov and Arun Kandoor and Aditya Shah and Asier Mujika and Da Huang and Christopher A. Choquette-Choo and Mohak Patel and Tianhe Yu and Toni Creswell and Jerry and Liu and Catarina Barros and Yasaman Razeghi and Aurko Roy and Phil Culliton and Binbin Xiong and Jiaqi Pan and Thomas Strohmann and Tolly Powell and Babi Seal and Doug DeCarlo and Pranav Shyam and Kaan Katircioglu and Xuezhi Wang and Cassidy Hardin and Immanuel Odisho and Josef Broder and Oscar Chang and Arun Nair and Artem Shtefan and Maura O'Brien and Manu Agarwal and Sahitya Potluri and Siddharth Goyal and Amit Jhindal and Saksham Thakur and Yury Stuken and James Lyon and Kristina Toutanova and Fangxiaoyu Feng and Austin Wu and Ben Horn and Alek Wang and Alex Cullum and Gabe Taubman and Disha Shrivastava and Chongyang Shi and Hamish Tomlinson and Roma Patel and Tao Tu and Ada Maksutaj Oflazer and Francesco Pongetti and Mingyao Yang and Adrien Ali Taïga and Vincent Perot and Nuo Wang Pierse and Feng Han and Yoel Drori and Iñaki Iturrate and Ayan Chakrabarti and Legg Yeung and Dave Dopson and Yi-ting Chen and Apoorv Kulshreshtha and Tongfei Guo and Philip Pham and Tal Schuster and Junquan Chen and Alex Polozov and Jinwei Xing and Huanjie Zhou and Praneeth Kacham and Doron Kukliansky and Antoine Miech and Sergey Yaroshenko and Ed Chi and Sholto Douglas and Hongliang Fei and Mathieu Blondel and Preethi Myla and Lior Madmoni and Xing Wu and Daniel Keysers and Kristian Kjems and Isabela Albuquerque and Lijun Yu and Joel D'sa and Michelle Plantan and Vlad Ionescu and Jaume Sanchez Elias and Abhirut Gupta and Manish Reddy Vuyyuru and Fred Alcober and Tong Zhou and Kaiyang Ji and Florian Hartmann and Subha Puttagunta and Hugo Song and Ehsan Amid and Anca Stefanoiu and Andrew Lee and Paul Pucciarelli and Emma Wang and Amit Raul and Slav Petrov and Isaac Tian and Valentin Anklin and Nana Nti and Victor Gomes and Max Schumacher and Grace Vesom and Alex Panagopoulos and Konstantinos Bousmalis and Daniel Andor and Josh Jacob and Yuan Zhang and Bill Rosgen and Matija Kecman and Matthew Tung and Alexandra Belias and Noah Goodman and Paul Covington and Brian Wieder and Nikita Saxena and Elnaz Davoodi and Muhuan Huang and Sharath Maddineni and Vincent Roulet and Folawiyo Campbell-Ajala and Pier Giuseppe Sessa and Xintian and Wu and Guangda Lai and Paul Collins and Alex Haig and Vytenis Sakenas and Xiaowei Xu and Marissa Giustina and Laurent El Shafey and Pichi Charoenpanit and Shefali Garg and Joshua Ainslie and Boone Severson and Montse Gonzalez Arenas and Shreya Pathak and Sujee Rajayogam and Jie Feng and Michiel Bakker and Sheng Li and Nevan Wichers and Jamie Rogers and Xinyang Geng and Yeqing Li and Rolf Jagerman and Chao Jia and Nadav Olmert and David Sharon and Matthew Mauger and Sandeep Mariserla and Hongxu Ma and Megha Mohabey and Kyuyeun Kim and Alek Andreev and Scott Pollom and Juliette Love and Vihan Jain and Priyanka Agrawal and Yannick Schroecker and Alisa Fortin and Manfred Warmuth and Ji Liu and Andrew Leach and Irina Blok and Ganesh Poomal Girirajan and Roee Aharoni and Benigno Uria and Andrei Sozanschi and Dan Goldberg and Lucian Ionita and Marco Tulio Ribeiro and Martin Zlocha and Vighnesh Birodkar and Sami Lachgar and Liangzhe Yuan and Himadri Choudhury and Matt Ginsberg and Fei Zheng and Gregory Dibb and Emily Graves and Swachhand Lokhande and Gabriel Rasskin and George-Cristian Muraru and Corbin Quick and Sandeep Tata and Pierre Sermanet and Aditya Chawla and Itay Karo and Yan Wang and Susan Zhang and Orgad Keller and Anca Dragan and Guolong Su and Ian Chou and Xi Liu and Yiqing Tao and Shruthi Prabhakara and Marc Wilson and Ruibo Liu and Shibo Wang and Georgie Evans and David Du and Alfonso Castaño and Gautam Prasad and Mona El Mahdy and Sebastian Gerlach and Machel Reid and Jarrod Kahn and Amir Zait and Thanumalayan Sankaranarayana Pillai and Thatcher Ulrich and Guanyu Wang and Jan Wassenberg and Efrat Farkash and Kiran Yalasangi and Congchao Wang and Maria Bauza and Simon Bucher and Ting Liu and Jun Yan and Gary Leung and Vikas Sindhwani and Parker Barnes and Avi Singh and Ivan Jurin and Jichuan Chang and Niket Kumar Bhumihar and Sivan Eiger and Gui Citovsky and Ben Withbroe and Zhang Li and Siyang Xue and Niccolò Dal Santo and Georgi Stoyanov and Yves Raimond and Steven Zheng and Yilin Gao and Vít Listík and Sławek Kwasiborski and Rachel Saputro and Adnan Ozturel and Ganesh Mallya and Kushal Majmundar and Ross West and Paul Caron and Jinliang Wei and Lluis Castrejon and Sharad Vikram and Deepak Ramachandran and Nikhil Dhawan and Jiho Park and Sara Smoot and George van den Driessche and Yochai Blau and Chase Malik and Wei Liang and Roy Hirsch and Cicero Nogueira dos Santos and Eugene Weinstein and Aäron van den Oord and Sid Lall and Nicholas FitzGerald and Zixuan Jiang and Xuan Yang and Dale Webster and Ali Elqursh and Aedan Pope and Georges Rotival and David Raposo and Wanzheng Zhu and Jeff Dean and Sami Alabed and Dustin Tran and Arushi Gupta and Zach Gleicher and Jessica Austin and Edouard Rosseel and Megh Umekar and Dipanjan Das and Yinghao Sun and Kai Chen and Karolis Misiunas and Xiang Zhou and Yixian Di and Alyssa Loo and Josh Newlan and Bo Li and Vinay Ramasesh and Ying Xu and Alex Chen and Sudeep Gandhe and Radu Soricut and Nikita Gupta and Shuguang Hu and Seliem El-Sayed and Xavier Garcia and Idan Brusilovsky and Pu-Chin Chen and Andrew Bolt and Lu Huang and Alex Gurney and Zhiying Zhang and Alexander Pritzel and Jarek Wilkiewicz and Bryan Seybold and Bhargav Kanagal Shamanna and Felix Fischer and Josef Dean and Karan Gill and Ross Mcilroy and Abhishek Bhowmick and Jeremy Selier and Antoine Yang and Derek Cheng and Vladimir Magay and Jie Tan and Dhriti Varma and Christian Walder and Tomas Kocisky and Ryo Nakashima and Paul Natsev and Mike Kwong and Ionel Gog and Chiyuan Zhang and Sander Dieleman and Thomas Jimma and Andrey Ryabtsev and Siddhartha Brahma and David Steiner and Dayou Du and Ante Žužul and Mislav Žanić and Mukund Raghavachari and Willi Gierke and Zeyu Zheng and Dessie Petrova and Yann Dauphin and Yuchuan Liu and Ido Kessler and Steven Hand and Chris Duvarney and Seokhwan Kim and Hyo Lee and Léonard Hussenot and Jeffrey Hui and Josh Smith and Deepali Jain and Jiawei Xia and Gaurav Singh Tomar and Keyvan Amiri and Du Phan and Fabian Fuchs and Tobias Weyand and Nenad Tomasev and Alexandra Cordell and Xin Liu and Jonathan Mallinson and Pankaj Joshi and Andy Crawford and Arun Suggala and Steve Chien and Nick Fernando and Mariella Sanchez-Vargas and Duncan Williams and Phil Crone and Xiyang Luo and Igor Karpov and Jyn Shan and Terry Thurk and Robin Strudel and Paul Voigtlaender and Piyush Patil and Tim Dozat and Ali Khodaei and Sahil Singla and Piotr Ambroszczyk and Qiyin Wu and Yifan Chang and Brian Roark and Chaitra Hegde and Tianli Ding and Angelos Filos and Zhongru Wu and André Susano Pinto and Shuang Liu and Saarthak Khanna and Aditya Pandey and Siobhan Mcloughlin and Qiujia Li and Sam Haves and Allan Zhou and Elena Buchatskaya and Isabel Leal and Peter de Boursac and Nami Akazawa and Nina Anderson and Terry Chen and Krishna Somandepalli and Chen Liang and Sheela Goenka and Stephanie Winkler and Alexander Grushetsky and Yifan Ding and Jamie Smith and Fan Ye and Jordi Pont-Tuset and Eric Li and Ruichao Li and Tomer Golany and Dawid Wegner and Tao Jiang and Omer Barak and Yuan Shangguan and Eszter Vértes and Renee Wong and Jörg Bornschein and Alex Tudor and Michele Bevilacqua and Tom Schaul and Ankit Singh Rawat and Yang Zhao and Kyriakos Axiotis and Lei Meng and Cory McLean and Jonathan Lai and Jennifer Beattie and Nate Kushman and Yaxin Liu and Blair Kutzman and Fiona Lang and Jingchen Ye and Praneeth Netrapalli and Pushkar Mishra and Myriam Khan and Megha Goel and Rob Willoughby and David Tian and Honglei Zhuang and JD Chen and Zak Tsai and Tasos Kementsietsidis and Arjun Khare and James Keeling and Keyang Xu and Nathan Waters and Florent Altché and Ashok Popat and Bhavishya Mittal and David Saxton and Dalia El Badawy and Michael Mathieu and Zheng Zheng and Hao Zhou and Nishant Ranka and Richard Shin and Qingnan Duan and Tim Salimans and Ioana Mihailescu and Uri Shaham and Ming-Wei Chang and Yannis Assael and Nishanth Dikkala and Martin Izzard and Vincent Cohen-Addad and Cat Graves and Vlad Feinberg and Grace Chung and DJ Strouse and Danny Karmon and Sahand Sharifzadeh and Zoe Ashwood and Khiem Pham and Jon Blanton and Alex Vasiloff and Jarred Barber and Mark Geller and Aurick Zhou and Fedir Zubach and Tzu-Kuo Huang and Lei Zhang and Himanshu Gupta and Matt Young and Julia Proskurnia and Ronny Votel and Valentin Gabeur and Gabriel Barcik and Aditya Tripathi and Hongkun Yu and Geng Yan and Beer Changpinyo and Filip Pavetić and Amy Coyle and Yasuhisa Fujii and Jorge Gonzalez Mendez and Tianhao Zhou and Harish Rajamani and Blake Hechtman and Eddie Cao and Da-Cheng Juan and Yi-Xuan Tan and Valentin Dalibard and Yilun Du and Natalie Clay and Kaisheng Yao and Wenhao Jia and Dimple Vijaykumar and Yuxiang Zhou and Xinyi Bai and Wei-Chih Hung and Steven Pecht and Georgi Todorov and Nikhil Khadke and Pramod Gupta and Preethi Lahoti and Arnaud Autef and Karthik Duddu and James Lee-Thorp and Alexander Bykovsky and Tautvydas Misiunas and Sebastian Flennerhag and Santhosh Thangaraj and Jed McGiffin and Zack Nado and Markus Kunesch and Andreas Noever and Amir Hertz and Marco Liang and Victor Stone and Evan Palmer and Samira Daruki and Arijit Pramanik and Siim Põder and Austin Kyker and Mina Khan and Evgeny Sluzhaev and Marvin Ritter and Avraham Ruderman and Wenlei Zhou and Chirag Nagpal and Kiran Vodrahalli and George Necula and Paul Barham and Ellie Pavlick and Jay Hartford and Izhak Shafran and Long Zhao and Maciej Mikuła and Tom Eccles and Hidetoshi Shimokawa and Kanav Garg and Luke Vilnis and Hanwen Chen and Ilia Shumailov and Kuang-Huei Lee and Abdelrahman Abdelhamed and Meiyan Xie and Vered Cohen and Ester Hlavnova and Dan Malkin and Chawin Sitawarin and James Lottes and Pauline Coquinot and Tianli Yu and Sandeep Kumar and Jingwei Zhang and Aroma Mahendru and Zafarali Ahmed and James Martens and Tao Chen and Aviel Boag and Daiyi Peng and Coline Devin and Arseniy Klimovskiy and Mary Phuong and Danny Vainstein and Jin Xie and Bhuvana Ramabhadran and Nathan Howard and Xinxin Yu and Gitartha Goswami and Jingyu Cui and Sam Shleifer and Mario Pinto and Chih-Kuan Yeh and Ming-Hsuan Yang and Sara Javanmardi and Dan Ethier and Chace Lee and Jordi Orbay and Suyog Kotecha and Carla Bromberg and Pete Shaw and James Thornton and Adi Gerzi Rosenthal and Shane Gu and Matt Thomas and Ian Gemp and Aditya Ayyar and Asahi Ushio and Aarush Selvan and Joel Wee and Chenxi Liu and Maryam Majzoubi and Weiren Yu and Jake Abernethy and Tyler Liechty and Renke Pan and Hoang Nguyen and Qiong and Hu and Sarah Perrin and Abhinav Arora and Emily Pitler and Weiyi Wang and Kaushik Shivakumar and Flavien Prost and Ben Limonchik and Jing Wang and Yi Gao and Timothee Cour and Shyamal Buch and Huan Gui and Maria Ivanova and Philipp Neubeck and Kelvin Chan and Lucy Kim and Huizhong Chen and Naman Goyal and Da-Woon Chung and Lu Liu and Yao Su and Anastasia Petrushkina and Jiajun Shen and Armand Joulin and Yuanzhong Xu and Stein Xudong Lin and Yana Kulizhskaya and Ciprian Chelba and Shobha Vasudevan and Eli Collins and Vasilisa Bashlovkina and Tony Lu and Doug Fritz and Jongbin Park and Yanqi Zhou and Chen Su and Richard Tanburn and Mikhail Sushkov and Mitchelle Rasquinha and Jinning Li and Jennifer Prendki and Yiming Li and Pallavi LV and Shriya Sharma and Hen Fitoussi and Hui Huang and Andrew Dai and Phuong Dao and Mike Burrows and Henry Prior and Danfeng Qin and Golan Pundak and Lars Lowe Sjoesund and Art Khurshudov and Zhenkai Zhu and Albert Webson and Elizabeth Kemp and Tat Tan and Saurabh Agrawal and Susie Sargsyan and Liqun Cheng and Jim Stephan and Tom Kwiatkowski and David Reid and Arunkumar Byravan and Assaf Hurwitz Michaely and Nicolas Heess and Luowei Zhou and Sonam Goenka and Viral Carpenter and Anselm Levskaya and Bo Wang and Reed Roberts and Rémi Leblond and Sharat Chikkerur and Stav Ginzburg and Max Chang and Robert Riachi and Chuqiao and Xu and Zalán Borsos and Michael Pliskin and Julia Pawar and Morgane Lustman and Hannah Kirkwood and Ankit Anand and Aditi Chaudhary and Norbert Kalb and Kieran Milan and Sean Augenstein and Anna Goldie and Laurel Prince and Karthik Raman and Yanhua Sun and Vivian Xia and Aaron Cohen and Zhouyuan Huo and Josh Camp and Seher Ellis and Lukas Zilka and David Vilar Torres and Lisa Patel and Sho Arora and Betty Chan and Jonas Adler and Kareem Ayoub and Jacky Liang and Fayaz Jamil and Jiepu Jiang and Simon Baumgartner and Haitian Sun and Yael Karov and Yaroslav Akulov and Hui Zheng and Irene Cai and Claudio Fantacci and James Rubin and Alex Rav Acha and Mengchao Wang and Nina D'Souza and Rohit Sathyanarayana and Shengyang Dai and Simon Rowe and Andrey Simanovsky and Omer Goldman and Yuheng Kuang and Xiaoyue Pan and Andrew Rosenberg and Tania Rojas-Esponda and Praneet Dutta and Amy Zeng and Irina Jurenka and Greg Farquhar and Yamini Bansal and Shariq Iqbal and Becca Roelofs and Ga-Young Joung and Parker Beak and Changwan Ryu and Ryan Poplin and Yan Wu and Jean-Baptiste Alayrac and Senaka Buthpitiya and Olaf Ronneberger and Caleb Habtegebriel and Wei Li and Paul Cavallaro and Aurora Wei and Guy Bensky and Timo Denk and Harish Ganapathy and Jeff Stanway and Pratik Joshi and Francesco Bertolini and Jessica Lo and Olivia Ma and Zachary Charles and Geta Sampemane and Himanshu Sahni and Xu Chen and Harry Askham and David Gaddy and Peter Young and Jiewen Tan and Matan Eyal and Arthur Bražinskas and Li Zhong and Zhichun Wu and Mark Epstein and Kai Bailey and Andrew Hard and Kamyu Lee and Sasha Goldshtein and Alex Ruiz and Mohammed Badawi and Matthias Lochbrunner and JK Kearns and Ashley Brown and Fabio Pardo and Theophane Weber and Haichuan Yang and Pan-Pan Jiang and Berkin Akin and Zhao Fu and Marcus Wainwright and Chi Zou and Meenu Gaba and Pierre-Antoine Manzagol and Wendy Kan and Yang Song and Karina Zainullina and Rui Lin and Jeongwoo Ko and Salil Deshmukh and Apoorv Jindal and James Svensson and Divya Tyam and Heri Zhao and Christine Kaeser-Chen and Scott Baird and Pooya Moradi and Jamie Hall and Qiuchen Guo and Vincent Tsang and Bowen Liang and Fernando Pereira and Suhas Ganesh and Ivan Korotkov and Jakub Adamek and Sridhar Thiagarajan and Vinh Tran and Charles Chen and Chris Tar and Sanil Jain and Ishita Dasgupta and Taylan Bilal and David Reitter and Kai Zhao and Giulia Vezzani and Yasmin Gehman and Pulkit Mehta and Lauren Beltrone and Xerxes Dotiwalla and Sergio Guadarrama and Zaheer Abbas and Stefani Karp and Petko Georgiev and Chun-Sung Ferng and Marc Brockschmidt and Liqian Peng and Christoph Hirnschall and Vikas Verma and Yingying Bi and Ying Xiao and Avigail Dabush and Kelvin Xu and Phil Wallis and Randall Parker and Qifei Wang and Yang Xu and Ilkin Safarli and Dinesh Tewari and Yin Zhang and Seungyeon Kim and Andrea Gesmundo and Mackenzie Thomas and Sergey Levi and Ahmed Chowdhury and Kanishka Rao and Peter Garst and Sam Conway-Rahman and Helen Ran and Kay McKinney and Zhisheng Xiao and Wenhao Yu and Rohan Agrawal and Axel Stjerngren and Catalin Ionescu and Jingjing Chen and Vivek Sharma and Justin Chiu and Fei Liu and Ken Franko and Clayton Sanford and Xingyu Cai and Paul Michel and Sanjay Ganapathy and Jane Labanowski and Zachary Garrett and Ben Vargas and Sean Sun and Bryan Gale and Thomas Buschmann and Guillaume Desjardins and Nimesh Ghelani and Palak Jain and Mudit Verma and Chulayuth Asawaroengchai and Julian Eisenschlos and Jitendra Harlalka and Hideto Kazawa and Don Metzler and Joshua Howland and Ying Jian and Jake Ades and Viral Shah and Tynan Gangwani and Seungji Lee and Roman Ring and Steven M. Hernandez and Dean Reich and Amer Sinha and Ashutosh Sathe and Joe Kovac and Ashleah Gill and Ajay Kannan and Andrea D'olimpio and Martin Sevenich and Jay Whang and Been Kim and Khe Chai Sim and Jilin Chen and Jiageng Zhang and Shuba Lall and Yossi Matias and Bill Jia and Abe Friesen and Sara Nasso and Ashish Thapliyal and Bryan Perozzi and Ting Yu and Anna Shekhawat and Safeen Huda and Peter Grabowski and Eric Wang and Ashwin Sreevatsa and Hilal Dib and Mehadi Hassen and Parker Schuh and Vedrana Milutinovic and Chris Welty and Michael Quinn and Ali Shah and Bangju Wang and Gabe Barth-Maron and Justin Frye and Natalie Axelsson and Tao Zhu and Yukun Ma and Irene Giannoumis and Hanie Sedghi and Chang Ye and Yi Luan and Kevin Aydin and Bilva Chandra and Vivek Sampathkumar and Ronny Huang and Victor Lavrenko and Ahmed Eleryan and Zhi Hong and Steven Hansen and Sara Mc Carthy and Bidisha Samanta and Domagoj Ćevid and Xin Wang and Fangtao Li and Michael Voznesensky and Matt Hoffman and Andreas Terzis and Vikash Sehwag and Gil Fidel and Luheng He and Mu Cai and Yanzhang He and Alex Feng and Martin Nikoltchev and Samrat Phatale and Jason Chase and Rory Lawton and Ming Zhang and Tom Ouyang and Manuel Tragut and Mehdi Hafezi Manshadi and Arjun Narayanan and Jiaming Shen and Xu Gao and Tolga Bolukbasi and Nick Roy and Xin Li and Daniel Golovin and Liviu Panait and Zhen Qin and Guangxing Han and Thomas Anthony and Sneha Kudugunta and Viorica Patraucean and Aniket Ray and Xinyun Chen and Xiaochen Yang and Tanuj Bhatia and Pranav Talluri and Alex Morris and Andrija Ražnatović and Bethanie Brownfield and James An and Sheng Peng and Patrick Kane and Ce Zheng and Nico Duduta and Joshua Kessinger and James Noraky and Siqi Liu and Keran Rong and Petar Veličković and Keith Rush and Alex Goldin and Fanny Wei and Shiva Mohan Reddy Garlapati and Caroline Pantofaru and Okwan Kwon and Jianmo Ni and Eric Noland and Julia Di Trapani and Françoise Beaufays and Abhijit Guha Roy and Yinlam Chow and Aybuke Turker and Geoffrey Cideron and Lantao Mei and Jon Clark and Qingyun Dou and Matko Bošnjak and Ralph Leith and Yuqing Du and Amir Yazdanbakhsh and Milad Nasr and Chester Kwak and Suraj Satishkumar Sheth and Alex Kaskasoli and Ankesh Anand and Balaji Lakshminarayanan and Sammy Jerome and David Bieber and Chun-Te Chu and Alexandre Senges and Tianxiao Shen and Mukund Sridhar and Ndaba Ndebele and Benjamin Beyret and Shakir Mohamed and Mia Chen and Markus Freitag and Jiaxian Guo and Luyang Liu and Paul Roit and Heng Chen and Shen Yan and Tom Stone and JD Co-Reyes and Jeremy Cole and Salvatore Scellato and Shekoofeh Azizi and Hadi Hashemi and Alicia Jin and Anand Iyer and Marcella Valentine and András György and Arun Ahuja and Daniel Hernandez Diaz and Chen-Yu Lee and Nathan Clement and Weize Kong and Drew Garmon and Ishaan Watts and Kush Bhatia and Khyatti Gupta and Matt Miecnikowski and Hugo Vallet and Ankur Taly and Edward Loper and Saket Joshi and James Atwood and Jo Chick and Mark Collier and Fotis Iliopoulos and Ryan Trostle and Beliz Gunel and Ramiro Leal-Cavazos and Arnar Mar Hrafnkelsson and Michael Guzman and Xiaoen Ju and Andy Forbes and Jesse Emond and Kushal Chauhan and Ben Caine and Li Xiao and Wenjun Zeng and Alexandre Moufarek and Daniel Murphy and Maya Meng and Nitish Gupta and Felix Riedel and Anil Das and Elijah Lawal and Shashi Narayan and Tiberiu Sosea and James Swirhun and Linda Friso and Behnam Neyshabur and Jing Lu and Sertan Girgin and Michael Wunder and Edouard Yvinec and Aroonalok Pyne and Victor Carbune and Shruti Rijhwani and Yang Guo and Tulsee Doshi and Anton Briukhov and Max Bain and Ayal Hitron and Xuanhui Wang and Ashish Gupta and Ke Chen and Cosmo Du and Weiyang Zhang and Dhruv Shah and Arjun Akula and Max Dylla and Ashyana Kachra and Weicheng Kuo and Tingting Zou and Lily Wang and Luyao Xu and Jifan Zhu and Justin Snyder and Sachit Menon and Orhan Firat and Igor Mordatch and Yuan Yuan and Natalia Ponomareva and Rory Blevins and Lawrence Moore and Weijun Wang and Phil Chen and Martin Scholz and Artur Dwornik and Jason Lin and Sicheng Li and Diego Antognini and Te I and Xiaodan Song and Matt Miller and Uday Kalra and Adam Raveret and Oscar Akerlund and Felix Wu and Andrew Nystrom and Namrata Godbole and Tianqi Liu and Hannah DeBalsi and Jewel Zhao and Buhuang Liu and Avi Caciularu and Lauren Lax and Urvashi Khandelwal and Victoria Langston and Eric Bailey and Silvio Lattanzi and Yufei Wang and Neel Kovelamudi and Sneha Mondal and Guru Guruganesh and Nan Hua and Ofir Roval and Paweł Wesołowski and Rishikesh Ingale and Jonathan Halcrow and Tim Sohn and Christof Angermueller and Bahram Raad and Eli Stickgold and Eva Lu and Alec Kosik and Jing Xie and Timothy Lillicrap and Austin Huang and Lydia Lihui Zhang and Dominik Paulus and Clement Farabet and Alex Wertheim and Bing Wang and Rishabh Joshi and Chu-ling Ko and Yonghui Wu and Shubham Agrawal and Lily Lin and XiangHai Sheng and Peter Sung and Tyler Breland-King and Christina Butterfield and Swapnil Gawde and Sumeet Singh and Qiao Zhang and Raj Apte and Shilpa Shetty and Adrian Hutter and Tao Li and Elizabeth Salesky and Federico Lebron and Jonni Kanerva and Michela Paganini and Arthur Nguyen and Rohith Vallu and Jan-Thorsten Peter and Sarmishta Velury and David Kao and Jay Hoover and Anna Bortsova and Colton Bishop and Shoshana Jakobovits and Alessandro Agostini and Alekh Agarwal and Chang Liu and Charles Kwong and Sasan Tavakkol and Ioana Bica and Alex Greve and Anirudh GP and Jake Marcus and Le Hou and Tom Duerig and Rivka Moroshko and Dave Lacey and Andy Davis and Julien Amelot and Guohui Wang and Frank Kim and Theofilos Strinopoulos and Hui Wan and Charline Le Lan and Shankar Krishnan and Haotian Tang and Peter Humphreys and Junwen Bai and Idan Heimlich Shtacher and Diego Machado and Chenxi Pang and Ken Burke and Dangyi Liu and Renga Aravamudhan and Yue Song and Ed Hirst and Abhimanyu Singh and Brendan Jou and Liang Bai and Francesco Piccinno and Chuyuan Kelly Fu and Robin Alazard and Barak Meiri and Daniel Winter and Charlie Chen and Mingda Zhang and Jens Heitkaemper and John Lambert and Jinhyuk Lee and Alexander Frömmgen and Sergey Rogulenko and Pranav Nair and Paul Niemczyk and Anton Bulyenov and Bibo Xu and Hadar Shemtov and Morteza Zadimoghaddam and Serge Toropov and Mateo Wirth and Hanjun Dai and Sreenivas Gollapudi and Daniel Zheng and Alex Kurakin and Chansoo Lee and Kalesha Bullard and Nicolas Serrano and Ivana Balazevic and Yang Li and Johan Schalkwyk and Mark Murphy and Mingyang Zhang and Kevin Sequeira and Romina Datta and Nishant Agrawal and Charles Sutton and Nithya Attaluri and Mencher Chiang and Wael Farhan and Gregory Thornton and Kate Lin and Travis Choma and Hung Nguyen and Kingshuk Dasgupta and Dirk Robinson and Iulia Comşa and Michael Riley and Arjun Pillai and Basil Mustafa and Ben Golan and Amir Zandieh and Jean-Baptiste Lespiau and Billy Porter and David Ross and Sujeevan Rajayogam and Mohit Agarwal and Subhashini Venugopalan and Bobak Shahriari and Qiqi Yan and Hao Xu and Taylor Tobin and Pavel Dubov and Hongzhi Shi and Adrià Recasens and Anton Kovsharov and Sebastian Borgeaud and Lucio Dery and Shanthal Vasanth and Elena Gribovskaya and Linhai Qiu and Mahdis Mahdieh and Wojtek Skut and Elizabeth Nielsen and CJ Zheng and Adams Yu and Carrie Grimes Bostock and Shaleen Gupta and Aaron Archer and Chris Rawles and Elinor Davies and Alexey Svyatkovskiy and Tomy Tsai and Yoni Halpern and Christian Reisswig and Bartek Wydrowski and Bo Chang and Joan Puigcerver and Mor Hazan Taege and Jian Li and Eva Schnider and Xinjian Li and Dragos Dena and Yunhan Xu and Umesh Telang and Tianze Shi and Heiga Zen and Kyle Kastner and Yeongil Ko and Neesha Subramaniam and Aviral Kumar and Pete Blois and Zhuyun Dai and John Wieting and Yifeng Lu and Yoel Zeldes and Tian Xie and Anja Hauth and Alexandru Ţifrea and Yuqi Li and Sam El-Husseini and Dan Abolafia and Howard Zhou and Wen Ding and Sahra Ghalebikesabi and Carlos Guía and Andrii Maksai and Ágoston Weisz and Sercan Arik and Nick Sukhanov and Aga Świetlik and Xuhui Jia and Luo Yu and Weiyue Wang and Mark Brand and Dawn Bloxwich and Sean Kirmani and Zhe Chen and Alec Go and Pablo Sprechmann and Nithish Kannen and Alen Carin and Paramjit Sandhu and Isabel Edkins and Leslie Nooteboom and Jai Gupta and Loren Maggiore and Javad Azizi and Yael Pritch and Pengcheng Yin and Mansi Gupta and Danny Tarlow and Duncan Smith and Desi Ivanov and Mohammad Babaeizadeh and Ankita Goel and Satish Kambala and Grace Chu and Matej Kastelic and Michelle Liu and Hagen Soltau and Austin Stone and Shivani Agrawal and Min Kim and Kedar Soparkar and Srinivas Tadepalli and Oskar Bunyan and Rachel Soh and Arvind Kannan and DY Kim and Blake JianHang Chen and Afief Halumi and Sudeshna Roy and Yulong Wang and Olcan Sercinoglu and Gena Gibson and Sijal Bhatnagar and Motoki Sano and Daniel von Dincklage and Qingchun Ren and Blagoj Mitrevski and Mirek Olšák and Jennifer She and Carl Doersch and Jilei and Wang and Bingyuan Liu and Qijun Tan and Tamar Yakar and Tris Warkentin and Alex Ramirez and Carl Lebsack and Josh Dillon and Rajiv Mathews and Tom Cobley and Zelin Wu and Zhuoyuan Chen and Jon Simon and Swaroop Nath and Tara Sainath and Alexei Bendebury and Ryan Julian and Bharath Mankalale and Daria Ćurko and Paulo Zacchello and Adam R. Brown and Kiranbir Sodhia and Heidi Howard and Sergi Caelles and Abhinav Gupta and Gareth Evans and Anna Bulanova and Lesley Katzen and Roman Goldenberg and Anton Tsitsulin and Joe Stanton and Benoit Schillings and Vitaly Kovalev and Corey Fry and Rushin Shah and Kuo Lin and Shyam Upadhyay and Cheng Li and Soroush Radpour and Marcello Maggioni and Jing Xiong and Lukas Haas and Jenny Brennan and Aishwarya Kamath and Nikolay Savinov and Arsha Nagrani and Trevor Yacovone and Ryan Kappedal and Kostas Andriopoulos and Li Lao and YaGuang Li and Grigory Rozhdestvenskiy and Kazuma Hashimoto and Andrew Audibert and Sophia Austin and Daniel Rodriguez and Anian Ruoss and Garrett Honke and Deep Karkhanis and Xi Xiong and Qing Wei and James Huang and Zhaoqi Leng and Vittal Premachandran and Stan Bileschi and Georgios Evangelopoulos and Thomas Mensink and Jay Pavagadhi and Denis Teplyashin and Paul Chang and Linting Xue and Garrett Tanzer and Sally Goldman and Kaushal Patel and Shixin Li and Jeremy Wiesner and Ivy Zheng and Ian Stewart-Binks and Jie Han and Zhi Li and Liangchen Luo and Karel Lenc and Mario Lučić and Fuzhao Xue and Ryan Mullins and Alexey Guseynov and Chung-Ching Chang and Isaac Galatzer-Levy and Adam Zhang and Garrett Bingham and Grace Hu and Ale Hartman and Yue Ma and Jordan Griffith and Alex Irpan and Carey Radebaugh and Summer Yue and Lijie Fan and Victor Ungureanu and Christina Sorokin and Hannah Teufel and Peiran Li and Rohan Anil and Dimitris Paparas and Todd Wang and Chu-Cheng Lin and Hui Peng and Megan Shum and Goran Petrovic and Demetra Brady and Richard Nguyen and Klaus Macherey and Zhihao Li and Harman Singh and Madhavi Yenugula and Mariko Iinuma and Xinyi Chen and Kavya Kopparapu and Alexey Stern and Shachi Dave and Chandu Thekkath and Florence Perot and Anurag Kumar and Fangda Li and Yang Xiao and Matthew Bilotti and Mohammad Hossein Bateni and Isaac Noble and Lisa Lee and Amelio Vázquez-Reina and Julian Salazar and Xiaomeng Yang and Boyu Wang and Ela Gruzewska and Anand Rao and Sindhu Raghuram and Zheng Xu and Eyal Ben-David and Jieru Mei and Sid Dalmia and Zhaoyi Zhang and Yuchen Liu and Gagan Bansal and Helena Pankov and Steven Schwarcz and Andrea Burns and Christine Chan and Sumit Sanghai and Ricky Liang and Ethan Liang and Antoine He and Amy Stuart and Arun Narayanan and Yukun Zhu and Christian Frank and Bahar Fatemi and Amit Sabne and Oran Lang and Indro Bhattacharya and Shane Settle and Maria Wang and Brendan McMahan and Andrea Tacchetti and Livio Baldini Soares and Majid Hadian and Serkan Cabi and Timothy Chung and Nikita Putikhin and Gang Li and Jeremy Chen and Austin Tarango and Henryk Michalewski and Mehran Kazemi and Hussain Masoom and Hila Sheftel and Rakesh Shivanna and Archita Vadali and Ramona Comanescu and Doug Reid and Joss Moore and Arvind Neelakantan and Michaël Sander and Jonathan Herzig and Aviv Rosenberg and Mostafa Dehghani and JD Choi and Michael Fink and Reid Hayes and Eric Ge and Shitao Weng and Chia-Hua Ho and John Karro and Kalpesh Krishna and Lam Nguyen Thiet and Amy Skerry-Ryan and Daniel Eppens and Marco Andreetto and Navin Sarma and Silvano Bonacina and Burcu Karagol Ayan and Megha Nawhal and Zhihao Shan and Mike Dusenberry and Shantanu Thakoor and Sagar Gubbi and Duc Dung Nguyen and Reut Tsarfaty and Samuel Albanie and Jovana Mitrović and Meet Gandhi and Bo-Juen Chen and Alessandro Epasto and Georgi Stephanov and Ye Jin and Samuel Gehman and Aida Amini and Jack Weber and Feryal Behbahani and Shawn Xu and Miltos Allamanis and Xi Chen and Myle Ott and Claire Sha and Michal Jastrzebski and Hang Qi and David Greene and Xinyi Wu and Abodunrinwa Toki and Daniel Vlasic and Jane Shapiro and Ragha Kotikalapudi and Zhe Shen and Takaaki Saeki and Sirui Xie and Albin Cassirer and Shikhar Bharadwaj and Tatsuya Kiyono and Srinadh Bhojanapalli and Elan Rosenfeld and Sam Ritter and Jieming Mao and João Gabriel Oliveira and Zoltan Egyed and Bernd Bandemer and Emilio Parisotto and Keisuke Kinoshita and Juliette Pluto and Petros Maniatis and Steve Li and Yaohui Guo and Golnaz Ghiasi and Jean Tarbouriech and Srimon Chatterjee and Julie Jin and Katrina and Xu and Jennimaria Palomaki and Séb Arnold and Madhavi Sewak and Federico Piccinini and Mohit Sharma and Ben Albrecht and Sean Purser-haskell and Ashwin Vaswani and Chongyan Chen and Matheus Wisniewski and Qin Cao and John Aslanides and Nguyet Minh Phu and Maximilian Sieb and Lauren Agubuzu and Anne Zheng and Daniel Sohn and Marco Selvi and Anders Andreassen and Krishan Subudhi and Prem Eruvbetine and Oliver Woodman and Tomas Mery and Sebastian Krause and Xiaoqi Ren and Xiao Ma and Jincheng Luo and Dawn Chen and Wei Fan and Henry Griffiths and Christian Schuler and Alice Li and Shujian Zhang and Jean-Michel Sarr and Shixin Luo and Riccardo Patana and Matthew Watson and Dani Naboulsi and Michael Collins and Sailesh Sidhwani and Emiel Hoogeboom and Sharon Silver and Emily Caveness and Xiaokai Zhao and Mikel Rodriguez and Maxine Deines and Libin Bai and Patrick Griffin and Marco Tagliasacchi and Emily Xue and Spandana Raj Babbula and Bo Pang and Nan Ding and Gloria Shen and Elijah Peake and Remi Crocker and Shubha Srinivas Raghvendra and Danny Swisher and Woohyun Han and Richa Singh and Ling Wu and Vladimir Pchelin and Tsendsuren Munkhdalai and Dana Alon and Geoff Bacon and Efren Robles and Jannis Bulian and Melvin Johnson and George Powell and Felipe Tiengo Ferreira and Yaoyiran Li and Frederik Benzing and Mihajlo Velimirović and Hubert Soyer and William Kong and Tony and Nguyên and Zhen Yang and Jeremiah Liu and Joost van Amersfoort and Daniel Gillick and Baochen Sun and Nathalie Rauschmayr and Katie Zhang and Serena Zhan and Tao Zhou and Alexey Frolov and Chengrun Yang and Denis Vnukov and Louis Rouillard and Hongji Li and Amol Mandhane and Nova Fallen and Rajesh Venkataraman and Clara Huiyi Hu and Jennifer Brennan and Jenny Lee and Jerry Chang and Martin Sundermeyer and Zhufeng Pan and Rosemary Ke and Simon Tong and Alex Fabrikant and William Bono and Jindong Gu and Ryan Foley and Yiran Mao and Manolis Delakis and Dhruva Bhaswar and Roy Frostig and Nick Li and Avital Zipori and Cath Hope and Olga Kozlova and Swaroop Mishra and Josip Djolonga and Craig Schiff and Majd Al Merey and Eleftheria Briakou and Peter Morgan and Andy Wan and Avinatan Hassidim and RJ Skerry-Ryan and Kuntal Sengupta and Mary Jasarevic and Praveen Kallakuri and Paige Kunkle and Hannah Brennan and Tom Lieber and Hassan Mansoor and Julian Walker and Bing Zhang and Annie Xie and Goran Žužić and Adaeze Chukwuka and Alex Druinsky and Donghyun Cho and Rui Yao and Ferjad Naeem and Shiraz Butt and Eunyoung Kim and Zhipeng Jia and Mandy Jordan and Adam Lelkes and Mark Kurzeja and Sophie Wang and James Zhao and Andrew Over and Abhishek Chakladar and Marcel Prasetya and Neha Jha and Sriram Ganapathy and Yale Cong and Prakash Shroff and Carl Saroufim and Sobhan Miryoosefi and Mohamed Hammad and Tajwar Nasir and Weijuan Xi and Yang Gao and Young Maeng and Ben Hora and Chin-Yi Cheng and Parisa Haghani and Yoad Lewenberg and Caden Lu and Martin Matysiak and Naina Raisinghani and Huiyu Wang and Lexi Baugher and Rahul Sukthankar and Minh Giang and John Schultz and Noah Fiedel and Minmin Chen and Cheng-Chun Lee and Tapomay Dey and Hao Zheng and Shachi Paul and Celine Smith and Andy Ly and Yicheng Wang and Rishabh Bansal and Bartek Perz and Susanna Ricco and Stasha Blank and Vaishakh Keshava and Deepak Sharma and Marvin Chow and Kunal Lad and Komal Jalan and Simon Osindero and Craig Swanson and Jacob Scott and Anastasija Ilić and Xiaowei Li and Siddhartha Reddy Jonnalagadda and Afzal Shama Soudagar and Yan Xiong and Bat-Orgil Batsaikhan and Daniel Jarrett and Naveen Kumar and Maulik Shah and Matt Lawlor and Austin Waters and Mark Graham and Rhys May and Sabela Ramos and Sandra Lefdal and Zeynep Cankara and Nacho Cano and Brendan O'Donoghue and Jed Borovik and Frederick Liu and Jordan Grimstad and Mahmoud Alnahlawi and Katerina Tsihlas and Tom Hudson and Nikolai Grigorev and Yiling Jia and Terry Huang and Tobenna Peter Igwe and Sergei Lebedev and Xiaodan Tang and Igor Krivokon and Frankie Garcia and Melissa Tan and Eric Jia and Peter Stys and Shikhar Vashishth and Yu Liang and Balaji Venkatraman and Chenjie Gu and Anastasios Kementsietsidis and Chen Zhu and Junehyuk Jung and Yunfei Bai and Mohammad Javad Hosseini and Faruk Ahmed and Aditya Gupta and Xin Yuan and Shereen Ashraf and Shitij Nigam and Gautam Vasudevan and Pranjal Awasthi and Adi Mayrav Gilady and Zelda Mariet and Ramy Eskander and Haiguang Li and Hexiang Hu and Guillermo Garrido and Philippe Schlattner and George Zhang and Rohun Saxena and Petar Dević and Kritika Muralidharan and Ashwin Murthy and Yiqian Zhou and Min Choi and Arissa Wongpanich and Zhengdong Wang and Premal Shah and Yuntao Xu and Yiling Huang and Stephen Spencer and Alice Chen and James Cohan and Junjie Wang and Jonathan Tompson and Junru Wu and Ruba Haroun and Haiqiong Li and Blanca Huergo and Fan Yang and Tongxin Yin and James Wendt and Michael Bendersky and Rahma Chaabouni and Javier Snaider and Johan Ferret and Abhishek Jindal and Tara Thompson and Andrew Xue and Will Bishop and Shubham Milind Phal and Archit Sharma and Yunhsuan Sung and Prabakar Radhakrishnan and Mo Shomrat and Reeve Ingle and Roopali Vij and Justin Gilmer and Mihai Dorin Istin and Sam Sobell and Yang Lu and Emily Nottage and Dorsa Sadigh and Jeremiah Willcock and Tingnan Zhang and Steve Xu and Sasha Brown and Katherine Lee and Gary Wang and Yun Zhu and Yi Tay and Cheolmin Kim and Audrey Gutierrez and Abhanshu Sharma and Yongqin Xian and Sungyong Seo and Claire Cui and Elena Pochernina and Cip Baetu and Krzysztof Jastrzębski and Mimi Ly and Mohamed Elhawaty and Dan Suh and Eren Sezener and Pidong Wang and Nancy Yuen and George Tucker and Jiahao Cai and Zuguang Yang and Cindy Wang and Alex Muzio and Hai Qian and Jae Yoo and Derek Lockhart and Kevin R. McKee and Mandy Guo and Malika Mehrotra and Artur Mendonça and Sanket Vaibhav Mehta and Sherry Ben and Chetan Tekur and Jiaqi Mu and Muye Zhu and Victoria Krakovna and Hongrae Lee and AJ Maschinot and Sébastien Cevey and HyunJeong Choe and Aijun Bai and Hansa Srinivasan and Derek Gasaway and Nick Young and Patrick Siegler and Dan Holtmann-Rice and Vihari Piratla and Kate Baumli and Roey Yogev and Alex Hofer and Hado van Hasselt and Svetlana Grant and Yuri Chervonyi and David Silver and Andrew Hogue and Ayushi Agarwal and Kathie Wang and Preeti Singh and Four Flynn and Josh Lipschultz and Robert David and Lizzetth Bellot and Yao-Yuan Yang and Long Le and Filippo Graziano and Kate Olszewska and Kevin Hui and Akanksha Maurya and Nikos Parotsidis and Weijie Chen and Tayo Oguntebi and Joe Kelley and Anirudh Baddepudi and Johannes Mauerer and Gregory Shaw and Alex Siegman and Lin Yang and Shravya Shetty and Subhrajit Roy and Yunting Song and Wojciech Stokowiec and Ryan Burnell and Omkar Savant and Robert Busa-Fekete and Jin Miao and Samrat Ghosh and Liam MacDermed and Phillip Lippe and Mikhail Dektiarev and Zach Behrman and Fabian Mentzer and Kelvin Nguyen and Meng Wei and Siddharth Verma and Chris Knutsen and Sudeep Dasari and Zhipeng Yan and Petr Mitrichev and Xingyu Wang and Virat Shejwalkar and Jacob Austin and Srinivas Sunkara and Navneet Potti and Yan Virin and Christian Wright and Gaël Liu and Oriana Riva and Etienne Pot and Greg Kochanski and Quoc Le and Gargi Balasubramaniam and Arka Dhar and Yuguo Liao and Adam Bloniarz and Divyansh Shukla and Elizabeth Cole and Jong Lee and Sheng Zhang and Sushant Kafle and Siddharth Vashishtha and Parsa Mahmoudieh and Grace Chen and Raphael Hoffmann and Pranesh Srinivasan and Agustin Dal Lago and Yoav Ben Shalom and Zi Wang and Michael Elabd and Anuj Sharma and Junhyuk Oh and Suraj Kothawade and Maigo Le and Marianne Monteiro and Shentao Yang and Kaiz Alarakyia and Robert Geirhos and Diana Mincu and Håvard Garnes and Hayato Kobayashi and Soroosh Mariooryad and Kacper Krasowiak and Zhixin and Lai and Shibl Mourad and Mingqiu Wang and Fan Bu and Ophir Aharoni and Guanjie Chen and Abhimanyu Goyal and Vadim Zubov and Ankur Bapna and Elahe Dabir and Nisarg Kothari and Kay Lamerigts and Nicola De Cao and Jeremy Shar and Christopher Yew and Nitish Kulkarni and Dre Mahaarachchi and Mandar Joshi and Zhenhai Zhu and Jared Lichtarge and Yichao Zhou and Hannah Muckenhirn and Vittorio Selo and Oriol Vinyals and Peter Chen and Anthony Brohan and Vaibhav Mehta and Sarah Cogan and Ruth Wang and Ty Geri and Wei-Jen Ko and Wei Chen and Fabio Viola and Keshav Shivam and Lisa Wang and Madeleine Clare Elish and Raluca Ada Popa and Sébastien Pereira and Jianqiao Liu and Raphael Koster and Donnie Kim and Gufeng Zhang and Sayna Ebrahimi and Partha Talukdar and Yanyan Zheng and Petra Poklukar and Ales Mikhalap and Dale Johnson and Anitha Vijayakumar and Mark Omernick and Matt Dibb and Ayush Dubey and Qiong Hu and Apurv Suman and Vaibhav Aggarwal and Ilya Kornakov and Fei Xia and Wing Lowe and Alexey Kolganov and Ted Xiao and Vitaly Nikolaev and Steven Hemingray and Bonnie Li and Joana Iljazi and Mikołaj Rybiński and Ballie Sandhu and Peggy Lu and Thang Luong and Rodolphe Jenatton and Vineetha Govindaraj and Hui and Li and Gabriel Dulac-Arnold and Wonpyo Park and Henry Wang and Abhinit Modi and Jean Pouget-Abadie and Kristina Greller and Rahul Gupta and Robert Berry and Prajit Ramachandran and Jinyu Xie and Liam McCafferty and Jianling Wang and Kilol Gupta and Hyeontaek Lim and Blaž Bratanič and Andy Brock and Ilia Akolzin and Jim Sproch and Dan Karliner and Duhyeon Kim and Adrian Goedeckemeyer and Noam Shazeer and Cordelia Schmid and Daniele Calandriello and Parul Bhatia and Krzysztof Choromanski and Ceslee Montgomery and Dheeru Dua and Ana Ramalho and Helen King and Yue Gao and Lynn Nguyen and David Lindner and Divya Pitta and Oleaser Johnson and Khalid Salama and Diego Ardila and Michael Han and Erin Farnese and Seth Odoom and Ziyue Wang and Xiangzhuo Ding and Norman Rink and Ray Smith and Harshal Tushar Lehri and Eden Cohen and Neera Vats and Tong He and Parthasarathy Gopavarapu and Adam Paszke and Miteyan Patel and Wouter Van Gansbeke and Lucia Loher and Luis Castro and Maria Voitovich and Tamara von Glehn and Nelson George and Simon Niklaus and Zach Eaton-Rosen and Nemanja Rakićević and Erik Jue and Sagi Perel and Carrie Zhang and Yuval Bahat and Angéline Pouget and Zhi Xing and Fantine Huot and Ashish Shenoy and Taylor Bos and Vincent Coriou and Bryan Richter and Natasha Noy and Yaqing Wang and Santiago Ontanon and Siyang Qin and Gleb Makarchuk and Demis Hassabis and Zhuowan Li and Mandar Sharma and Kumaran Venkatesan and Iurii Kemaev and Roxanne Daniel and Shiyu Huang and Saloni Shah and Octavio Ponce and Warren and Chen and Manaal Faruqui and Jialin Wu and Slavica Andačić and Szabolcs Payrits and Daniel McDuff and Tom Hume and Yuan Cao and MH Tessler and Qingze Wang and Yinan Wang and Ivor Rendulic and Eirikur Agustsson and Matthew Johnson and Tanya Lando and Andrew Howard and Sri Gayatri Sundara Padmanabhan and Mayank Daswani and Andrea Banino and Michael Kilgore and Jonathan Heek and Ziwei Ji and Alvaro Caceres and Conglong Li and Nora Kassner and Alexey Vlaskin and Zeyu Liu and Alex Grills and Yanhan Hou and Roykrong Sukkerd and Gowoon Cheon and Nishita Shetty and Larisa Markeeva and Piotr Stanczyk and Tejas Iyer and Yuan Gong and Shawn Gao and Keerthana Gopalakrishnan and Tim Blyth and Malcolm Reynolds and Avishkar Bhoopchand and Misha Bilenko and Dero Gharibian and Vicky Zayats and Aleksandra Faust and Abhinav Singh and Min Ma and Hongyang Jiao and Sudheendra Vijayanarasimhan and Lora Aroyo and Vikas Yadav and Sarah Chakera and Ashwin Kakarla and Vilobh Meshram and Karol Gregor and Gabriela Botea and Evan Senter and Dawei Jia and Geza Kovacs and Neha Sharma and Sebastien Baur and Kai Kang and Yifan He and Lin Zhuo and Marija Kostelac and Itay Laish and Songyou Peng and Louis O'Bryan and Daniel Kasenberg and Girish Ramchandra Rao and Edouard Leurent and Biao Zhang and Sage Stevens and Ana Salazar and Ye Zhang and Ivan Lobov and Jake Walker and Allen Porter and Morgan Redshaw and Han Ke and Abhishek Rao and Alex Lee and Hoi Lam and Michael Moffitt and Jaeyoun Kim and Siyuan Qiao and Terry Koo and Robert Dadashi and Xinying Song and Mukund Sundararajan and Peng Xu and Chizu Kawamoto and Yan Zhong and Clara Barbu and Apoorv Reddy and Mauro Verzetti and Leon Li and George Papamakarios and Hanna Klimczak-Plucińska and Mary Cassin and Koray Kavukcuoglu and Rigel Swavely and Alain Vaucher and Jeffrey Zhao and Ross Hemsley and Michael Tschannen and Heming Ge and Gaurav Menghani and Yang Yu and Natalie Ha and Wei He and Xiao Wu and Maggie Song and Rachel Sterneck and Stefan Zinke and Dan A. Calian and Annie Marsden and Alejandro Cruzado Ruiz and Matteo Hessel and Almog Gueta and Benjamin Lee and Brian Farris and Manish Gupta and Yunjie Li and Mohammad Saleh and Vedant Misra and Kefan Xiao and Piermaria Mendolicchio and Gavin Buttimore and Varvara Krayvanova and Nigamaa Nayakanti and Matthew Wiethoff and Yash Pande and Azalia Mirhoseini and Ni Lao and Jasmine Liu and Yiqing Hua and Angie Chen and Yury Malkov and Dmitry Kalashnikov and Shubham Gupta and Kartik Audhkhasi and Yuexiang Zhai and Sudhindra Kopalle and Prateek Jain and Eran Ofek and Clemens Meyer and Khuslen Baatarsukh and Hana Strejček and Jun Qian and James Freedman and Ricardo Figueira and Michal Sokolik and Olivier Bachem and Raymond Lin and Dia Kharrat and Chris Hidey and Pingmei Xu and Dennis Duan and Yin Li and Muge Ersoy and Richard Everett and Kevin Cen and Rebeca Santamaria-Fernandez and Amir Taubenfeld and Ian Mackinnon and Linda Deng and Polina Zablotskaia and Shashank Viswanadha and Shivanker Goel and Damion Yates and Yunxiao Deng and Peter Choy and Mingqing Chen and Abhishek Sinha and Alex Mossin and Yiming Wang and Arthur Szlam and Susan Hao and Paul Kishan Rubenstein and Metin Toksoz-Exley and Miranda Aperghis and Yin Zhong and Junwhan Ahn and Michael Isard and Olivier Lacombe and Florian Luisier and Chrysovalantis Anastasiou and Yogesh Kalley and Utsav Prabhu and Emma Dunleavy and Shaan Bijwadia and Justin Mao-Jones and Kelly Chen and Rama Pasumarthi and Emily Wood and Adil Dostmohamed and Nate Hurley and Jiri Simsa and Alicia Parrish and Mantas Pajarskas and Matt Harvey and Ondrej Skopek and Yony Kochinski and Javier Rey and Verena Rieser and Denny Zhou and Sun Jae Lee and Trilok Acharya and Guowang Li and Joe Jiang and Xiaofan Zhang and Bryant Gipson and Ethan Mahintorabi and Marco Gelmi and Nima Khajehnouri and Angel Yeh and Kayi Lee and Loic Matthey and Leslie Baker and Trang Pham and Han Fu and Alex Pak and Prakhar Gupta and Cristina Vasconcelos and Adam Sadovsky and Brian Walker and Sissie Hsiao and Patrik Zochbauer and Andreea Marzoca and Noam Velan and Junhao Zeng and Gilles Baechler and Danny Driess and Divya Jain and Yanping Huang and Lizzie Tao and John Maggs and Nir Levine and Jon Schneider and Erika Gemzer and Samuel Petit and Shan Han and Zach Fisher and Dustin Zelle and Courtney Biles and Eugene Ie and Asya Fadeeva and Casper Liu and Juliana Vicente Franco and Adrian Collister and Hao Zhang and Renshen Wang and Ruizhe Zhao and Leandro Kieliger and Kurt Shuster and Rui Zhu and Boqing Gong and Lawrence Chan and Ruoxi Sun and Sujoy Basu and Roland Zimmermann and Jamie Hayes and Abhishek Bapna and Jasper Snoek and Weel Yang and Puranjay Datta and Jad Al Abdallah and Kevin Kilgour and Lu Li and SQ Mah and Yennie Jun and Morgane Rivière and Abhijit Karmarkar and Tammo Spalink and Tao Huang and Lucas Gonzalez and Duc-Hieu Tran and Averi Nowak and John Palowitch and Martin Chadwick and Ellie Talius and Harsh Mehta and Thibault Sellam and Philipp Fränken and Massimo Nicosia and Kyle He and Aditya Kini and David Amos and Sugato Basu and Harrison Jobe and Eleni Shaw and Qiantong Xu and Colin Evans and Daisuke Ikeda and Chaochao Yan and Larry Jin and Lun Wang and Sachin Yadav and Ilia Labzovsky and Ramesh Sampath and Ada Ma and Candice Schumann and Aditya Siddhant and Rohin Shah and John Youssef and Rishabh Agarwal and Natalie Dabney and Alessio Tonioni and Moran Ambar and Jing Li and Isabelle Guyon and Benny Li and David Soergel and Boya Fang and Georgi Karadzhov and Cristian Udrescu and Trieu Trinh and Vikas Raunak and Seb Noury and Dee Guo and Sonal Gupta and Mara Finkelstein and Denis Petek and Lihao Liang and Greg Billock and Pei Sun and David Wood and Yiwen Song and Xiaobin Yu and Tatiana Matejovicova and Regev Cohen and Kalyan Andra and David D'Ambrosio and Zhiwei Deng and Vincent Nallatamby and Ebrahim Songhori and Rumen Dangovski and Andrew Lampinen and Pankil Botadra and Adam Hillier and Jiawei Cao and Nagabhushan Baddi and Adhi Kuncoro and Toshihiro Yoshino and Ankit Bhagatwala and Marcáurelio Ranzato and Rylan Schaeffer and Tianlin Liu and Shuai Ye and Obaid Sarvana and John Nham and Chenkai Kuang and Isabel Gao and Jinoo Baek and Shubham Mittal and Ayzaan Wahid and Anita Gergely and Bin Ni and Josh Feldman and Carrie Muir and Pascal Lamblin and Wolfgang Macherey and Ethan Dyer and Logan Kilpatrick and Víctor Campos and Mukul Bhutani and Stanislav Fort and Yanif Ahmad and Aliaksei Severyn and Kleopatra Chatziprimou and Oleksandr Ferludin and Mason Dimarco and Aditya Kusupati and Joe Heyward and Dan Bahir and Kevin Villela and Katie Millican and Dror Marcus and Sanaz Bahargam and Caglar Unlu and Nicholas Roth and Zichuan Wei and Siddharth Gopal and Deepanway Ghoshal and Edward Lee and Sharon Lin and Jennie Lees and Dayeong Lee and Anahita Hosseini and Connie Fan and Seth Neel and Marcus Wu and Yasemin Altun and Honglong Cai and Enrique Piqueras and Josh Woodward and Alessandro Bissacco and Salem Haykal and Mahyar Bordbar and Prasha Sundaram and Sarah Hodkinson and Daniel Toyama and George Polovets and Austin Myers and Anu Sinha and Tomer Levinboim and Kashyap Krishnakumar and Rachita Chhaparia and Tatiana Sholokhova and Nitesh Bharadwaj Gundavarapu and Ganesh Jawahar and Haroon Qureshi and Jieru Hu and Nikola Momchev and Matthew Rahtz and Renjie Wu and Aishwarya P S and Kedar Dhamdhere and Meiqi Guo and Umang Gupta and Ali Eslami and Mariano Schain and Michiel Blokzijl and David Welling and Dave Orr and Levent Bolelli and Nicolas Perez-Nieves and Mikhail Sirotenko and Aman Prasad and Arjun Kar and Borja De Balle Pigem and Tayfun Terzi and Gellért Weisz and Dipankar Ghosh and Aditi Mavalankar and Dhruv Madeka and Kaspar Daugaard and Hartwig Adam and Viraj Shah and Dana Berman and Maggie Tran and Steven Baker and Ewa Andrejczuk and Grishma Chole and Ganna Raboshchuk and Mahdi Mirzazadeh and Thais Kagohara and Shimu Wu and Christian Schallhart and Bernett Orlando and Chen Wang and Alban Rrustemi and Hao Xiong and Hao Liu and Arpi Vezer and Nolan Ramsden and Shuo-yiin Chang and Sidharth Mudgal and Yan Li and Nino Vieillard and Yedid Hoshen and Farooq Ahmad and Ambrose Slone and Amy Hua and Natan Potikha and Mirko Rossini and Jon Stritar and Sushant Prakash and Zifeng Wang and Xuanyi Dong and Alireza Nazari and Efrat Nehoran and Kaan Tekelioglu and Yinxiao Li and Kartikeya Badola and Tom Funkhouser and Yuanzhen Li and Varun Yerram and Ramya Ganeshan and Daniel Formoso and Karol Langner and Tian Shi and Huijian Li and Yumeya Yamamori and Amayika Panda and Alaa Saade and Angelo Scorza Scarpati and Chris Breaux and CJ Carey and Zongwei Zhou and Cho-Jui Hsieh and Sophie Bridgers and Alena Butryna and Nishesh Gupta and Vaibhav Tulsyan and Sanghyun Woo and Evgenii Eltyshev and Will Grathwohl and Chanel Parks and Seth Benjamin and Rina Panigrahy and Shenil Dodhia and Daniel De Freitas and Chris Sauer and Will Song and Ferran Alet and Jackson Tolins and Cosmin Paduraru and Xingyi Zhou and Brian Albert and Zizhao Zhang and Lei Shu and Mudit Bansal and Sarah Nguyen and Amir Globerson and Owen Xiao and James Manyika and Tom Hennigan and Rong Rong and Josip Matak and Anton Bakalov and Ankur Sharma and Danila Sinopalnikov and Andrew Pierson and Stephen Roller and Geoff Brown and Mingcen Gao and Toshiyuki Fukuzawa and Amin Ghafouri and Kenny Vassigh and Iain Barr and Zhicheng Wang and Anna Korsun and Rajesh Jayaram and Lijie Ren and Tim Zaman and Samira Khan and Yana Lunts and Dan Deutsch and Dave Uthus and Nitzan Katz and Masha Samsikova and Amr Khalifa and Nikhil Sethi and Jiao Sun and Luming Tang and Uri Alon and Xianghong Luo and Dian Yu and Abhishek Nayyar and Bryce Petrini and Will Truong and Vincent Hellendoorn and Nikolai Chinaev and Chris Alberti and Wei Wang and Jingcao Hu and Vahab Mirrokni and Ananth Balashankar and Avia Aharon and Aahil Mehta and Ahmet Iscen and Joseph Kready and Lucas Manning and Anhad Mohananey and Yuankai Chen and Anshuman Tripathi and Allen Wu and Igor Petrovski and Dawsen Hwang and Martin Baeuml and Shreyas Chandrakaladharan and Yuan Liu and Rey Coaguila and Maxwell Chen and Sally Ma and Pouya Tafti and Susheel Tatineni and Terry Spitz and Jiayu Ye and Paul Vicol and Mihaela Rosca and Adrià Puigdomènech and Zohar Yahav and Sanjay Ghemawat and Hanzhao Lin and Phoebe Kirk and Zaid Nabulsi and Sergey Brin and Bernd Bohnet and Ken Caluwaerts and Aditya Srikanth Veerubhotla and Dan Zheng and Zihang Dai and Petre Petrov and Yichong Xu and Ramin Mehran and Zhuo Xu and Luisa Zintgraf and Jiho Choi and Spurthi Amba Hombaiah and Romal Thoppilan and Sashank Reddi and Lukasz Lew and Li Li and Kellie Webster and KP Sawhney and Lampros Lamprou and Siamak Shakeri and Mayank Lunayach and Jianmin Chen and Sumit Bagri and Alex Salcianu and Ying Chen and Yani Donchev and Charlotte Magister and Signe Nørly and Vitor Rodrigues and Tomas Izo and Hila Noga and Joe Zou and Thomas Köppe and Wenxuan Zhou and Kenton Lee and Xiangzhu Long and Danielle Eisenbud and Anthony Chen and Connor Schenck and Chi Ming To and Peilin Zhong and Emanuel Taropa and Minh Truong and Omer Levy and Danilo Martins and Zhiyuan Zhang and Christopher Semturs and Kelvin Zhang and Alex Yakubovich and Pol Moreno and Lara McConnaughey and Di Lu and Sam Redmond and Lotte Weerts and Yonatan Bitton and Tiziana Refice and Nicolas Lacasse and Arthur Conmy and Corentin Tallec and Julian Odell and Hannah Forbes-Pollard and Arkadiusz Socala and Jonathan Hoech and Pushmeet Kohli and Alanna Walton and Rui Wang and Mikita Sazanovich and Kexin Zhu and Andrei Kapishnikov and Rich Galt and Matthew Denton and Ben Murdoch and Caitlin Sikora and Kareem Mohamed and Wei Wei and Uri First and Tim McConnell and Luis C. Cobo and James Qin and Thi Avrahami and Daniel Balle and Yu Watanabe and Annie Louis and Adam Kraft and Setareh Ariafar and Yiming Gu and Eugénie Rives and Charles Yoon and Andrei Rusu and James Cobon-Kerr and Chris Hahn and Jiaming Luo and Yuvein and Zhu and Niharika Ahuja and Rodrigo Benenson and Raphaël Lopez Kaufman and Honglin Yu and Lloyd Hightower and Junlin Zhang and Darren Ni and Lisa Anne Hendricks and Gabby Wang and Gal Yona and Lalit Jain and Pablo Barrio and Surya Bhupatiraju and Siva Velusamy and Allan Dafoe and Sebastian Riedel and Tara Thomas and Zhe Yuan and Mathias Bellaiche and Sheena Panthaplackel and Klemen Kloboves and Sarthak Jauhari and Canfer Akbulut and Todor Davchev and Evgeny Gladchenko and David Madras and Aleksandr Chuklin and Tyrone Hill and Quan Yuan and Mukundan Madhavan and Luke Leonhard and Dylan Scandinaro and Qihang Chen and Ning Niu and Arthur Douillard and Bogdan Damoc and Yasumasa Onoe and Fabian Pedregosa and Fred Bertsch and Chas Leichner and Joseph Pagadora and Jonathan Malmaud and Sameera Ponda and Andy Twigg and Oleksii Duzhyi and Jingwei Shen and Miaosen Wang and Roopal Garg and Jing Chen and Utku Evci and Jonathan Lee and Leon Liu and Koji Kojima and Masa Yamaguchi and Arunkumar Rajendran and AJ Piergiovanni and Vinodh Kumar Rajendran and Marco Fornoni and Gabriel Ibagon and Harry Ragan and Sadh MNM Khan and John Blitzer and Andrew Bunner and Guan Sun and Takahiro Kosakai and Scott Lundberg and Ndidi Elue and Kelvin Guu and SK Park and Jane Park and Arunachalam Narayanaswamy and Chengda Wu and Jayaram Mudigonda and Trevor Cohn and Hairong Mu and Ravi Kumar and Laura Graesser and Yichi Zhang and Richard Killam and Vincent Zhuang and Mai Giménez and Wael Al Jishi and Ruy Ley-Wild and Alex Zhai and Kazuki Osawa and Diego Cedillo and Jialu Liu and Mayank Upadhyay and Marcin Sieniek and Roshan Sharma and Tom Paine and Anelia Angelova and Sravanti Addepalli and Carolina Parada and Kingshuk Majumder and Avery Lamp and Sanjiv Kumar and Xiang Deng and Artiom Myaskovsky and Tea Sabolić and Jeffrey Dudek and Sarah York and Félix de Chaumont Quitry and Jiazhong Nie and Dee Cattle and Alok Gunjan and Bilal Piot and Waleed Khawaja and Seojin Bang and Simon Wang and Siavash Khodadadeh and Raghavender R and Praynaa Rawlani and Richard Powell and Kevin Lee and Johannes Griesser and GS Oh and Cesar Magalhaes and Yujia Li and Simon Tokumine and Hadas Natalie Vogel and Dennis Hsu and Arturo BC and Disha Jindal and Matan Cohen and Zi Yang and Junwei Yuan and Dario de Cesare and Tony Bruguier and Jun Xu and Monica Roy and Alon Jacovi and Dan Belov and Rahul Arya and Phoenix Meadowlark and Shlomi Cohen-Ganor and Wenting Ye and Patrick Morris-Suzuki and Praseem Banzal and Gan Song and Pranavaraj Ponnuramu and Fred Zhang and George Scrivener and Salah Zaiem and Alif Raditya Rochman and Kehang Han and Badih Ghazi and Kate Lee and Shahar Drath and Daniel Suo and Antonious Girgis and Pradeep Shenoy and Duy Nguyen and Douglas Eck and Somit Gupta and Le Yan and Joao Carreira and Anmol Gulati and Ruoxin Sang and Daniil Mirylenka and Emma Cooney and Edward Chou and Mingyang Ling and Cindy Fan and Ben Coleman and Guilherme Tubone and Ravin Kumar and Jason Baldridge and Felix Hernandez-Campos and Angeliki Lazaridou and James Besley and Itay Yona and Neslihan Bulut and Quentin Wellens and AJ Pierigiovanni and Jasmine George and Richard Green and Pu Han and Connie Tao and Geoff Clark and Chong You and Abbas Abdolmaleki and Justin Fu and Tongzhou Chen and Ashwin Chaugule and Angad Chandorkar and Altaf Rahman and Will Thompson and Penporn Koanantakool and Mike Bernico and Jie Ren and Andrey Vlasov and Sergei Vassilvitskii and Maciej Kula and Yizhong Liang and Dahun Kim and Yangsibo Huang and Chengxi Ye and Dmitry Lepikhin and Wesley Helmholz},
      year={2025},
      eprint={2507.06261},
      archivePrefix={arXiv},
      primaryClass={cs.CL},
      url={https://arxiv.org/abs/2507.06261}, 
}

@inproceedings{hessel2021clipscore,
  title={Clipscore: A reference-free evaluation metric for image captioning},
  author={Hessel, Jack and Holtzman, Ari and Forbes, Maxwell and Le Bras, Ronan and Choi, Yejin},
  booktitle={Proceedings of the 2021 conference on empirical methods in natural language processing},
  pages={7514--7528},
  year={2021}
}

@article{bai2025qwen3,
  title={Qwen3-vl technical report},
  author={Bai, Shuai and Cai, Yuxuan and Chen, Ruizhe and Chen, Keqin and Chen, Xionghui and Cheng, Zesen and Deng, Lianghao and Ding, Wei and Gao, Chang and Ge, Chunjiang and others},
  journal={arXiv preprint arXiv:2511.21631},
  year={2025}
}

@misc{bai2025qwen25vltechnicalreport,
      title={Qwen2.5-VL Technical Report}, 
      author={Shuai Bai and Keqin Chen and Xuejing Liu and Jialin Wang and Wenbin Ge and Sibo Song and Kai Dang and Peng Wang and Shijie Wang and Jun Tang and Humen Zhong and Yuanzhi Zhu and Mingkun Yang and Zhaohai Li and Jianqiang Wan and Pengfei Wang and Wei Ding and Zheren Fu and Yiheng Xu and Jiabo Ye and Xi Zhang and Tianbao Xie and Zesen Cheng and Hang Zhang and Zhibo Yang and Haiyang Xu and Junyang Lin},
      year={2025},
      eprint={2502.13923},
      archivePrefix={arXiv},
      primaryClass={cs.CV},
      url={https://arxiv.org/abs/2502.13923}, 
}

@misc{wan2025wanopenadvancedlargescale,
      title={Wan: Open and Advanced Large-Scale Video Generative Models}, 
      author={Team Wan and Ang Wang and Baole Ai and Bin Wen and Chaojie Mao and Chen-Wei Xie and Di Chen and Feiwu Yu and Haiming Zhao and Jianxiao Yang and Jianyuan Zeng and Jiayu Wang and Jingfeng Zhang and Jingren Zhou and Jinkai Wang and Jixuan Chen and Kai Zhu and Kang Zhao and Keyu Yan and Lianghua Huang and Mengyang Feng and Ningyi Zhang and Pandeng Li and Pingyu Wu and Ruihang Chu and Ruili Feng and Shiwei Zhang and Siyang Sun and Tao Fang and Tianxing Wang and Tianyi Gui and Tingyu Weng and Tong Shen and Wei Lin and Wei Wang and Wei Wang and Wenmeng Zhou and Wente Wang and Wenting Shen and Wenyuan Yu and Xianzhong Shi and Xiaoming Huang and Xin Xu and Yan Kou and Yangyu Lv and Yifei Li and Yijing Liu and Yiming Wang and Yingya Zhang and Yitong Huang and Yong Li and You Wu and Yu Liu and Yulin Pan and Yun Zheng and Yuntao Hong and Yupeng Shi and Yutong Feng and Zeyinzi Jiang and Zhen Han and Zhi-Fan Wu and Ziyu Liu},
      year={2025},
      eprint={2503.20314},
      archivePrefix={arXiv},
      primaryClass={cs.CV},
      url={https://arxiv.org/abs/2503.20314}, 
}

\appendix
\setcounter{section}{0}

\section*{Appendix}\label{sec:appendix}
This section provides additional examples to assist in the understanding and interpretation of the research work presented.

\noindent
Section~ \ref{app:data}: Dataset Statistics
\vspace{1mm}

\noindent
Section~ \ref{app:annotation}: Annotation 
\vspace{1mm}

\noindent
Section~ \ref{app:prompt}: Prompts
\vspace{1mm}

\noindent
Section~ \ref{app:experiments}: Experiments
\vspace{1mm}

\section{Dataset Statistics and Analysis} \label{app:data}
\vspace{1mm}

\begin{table*}[ht]
\centering
\vspace{4pt}
\resizebox{\textwidth}{!}{
\begin{tabular}{ll cccccccccc}
\toprule
\textbf{Dimension} & \textbf{Category} & \textbf{Brazil} & \textbf{Canada} & \textbf{Chile} & \textbf{China} & \textbf{Germany} & \textbf{India} & \textbf{Iran} & \textbf{Japan} & \textbf{Poland} & \textbf{S.~Africa} \\
\midrule
\multirow{5}{*}{Identity} & Dates of Sig. & $19$ & $23$ & $20$ & $23$ & $21$ & $28$ & $19$ & $22$ & $17$ & $23$ \\
 & Etiquette & $11$ & $15$ & $12$ & $13$ & $17$ & $18$ & $19$ & $16$ & $13$ & $18$ \\
 & Family & $13$ & $14$ & $12$ & $14$ & $12$ & $11$ & $11$ & $11$ & $10$ & $0$ \\
 & Greetings & $11$ & $10$ & $10$ & $10$ & $9$ & $12$ & $12$ & $13$ & $11$ & $12$ \\
 & Religion & $16$ & $12$ & $10$ & $12$ & $8$ & $17$ & $10$ & $15$ & $7$ & $16$ \\
\midrule
\multirow{5}{*}{Context} & Dates of Sig. & $28$ & $31$ & $32$ & $36$ & $33$ & $37$ & $30$ & $35$ & $34$ & $35$ \\
 & Etiquette & $20$ & $20$ & $22$ & $28$ & $21$ & $24$ & $25$ & $27$ & $19$ & $26$ \\
 & Family & $15$ & $17$ & $17$ & $17$ & $13$ & $14$ & $14$ & $14$ & $15$ & $0$ \\
 & Greetings & $14$ & $11$ & $12$ & $12$ & $12$ & $13$ & $13$ & $15$ & $11$ & $14$ \\
 & Religion & $17$ & $13$ & $11$ & $14$ & $11$ & $18$ & $11$ & $16$ & $14$ & $16$ \\
\midrule
\multirow{5}{*}{Behavior} & Dates of Sig. & $26$ & $27$ & $29$ & $33$ & $32$ & $37$ & $28$ & $33$ & $32$ & $34$ \\
 & Etiquette & $21$ & $20$ & $23$ & $27$ & $22$ & $23$ & $24$ & $25$ & $19$ & $26$ \\
 & Family & $15$ & $15$ & $17$ & $17$ & $12$ & $14$ & $13$ & $14$ & $14$ & $0$ \\
 & Greetings & $14$ & $11$ & $12$ & $13$ & $11$ & $13$ & $13$ & $13$ & $11$ & $14$ \\
 & Religion & $16$ & $12$ & $11$ & $13$ & $10$ & $18$ & $10$ & $14$ & $11$ & $15$ \\
\bottomrule
\end{tabular}
}
\caption{Number of unique Identity, Context, and Behavior entries per country and category. South Africa contains no Family entries, reflecting the absence of Family category data points in the source CulturalFrames dataset.}
\label{tab:app_dataset_category_breakdown}
\end{table*}

\subsection{Cultural Atlas} \label{app:culture_atlas}

The \textbf{Cultural Atlas} is a landmark collaborative initiative developed through a partnership between Mosaica, SBS, and Multicultural NSW. Established in 2016, the project serves as a comprehensive educational resource aimed at enhancing cross-cultural literacy. It provides nuanced insights into the attitudes, social norms, and communication styles of Australia's diverse populations.
By synthesizing qualitative cultural observations with contemporary demographic statistics and settlement trends, the Atlas equips individuals and organizations with the tools necessary to navigate a pluralistic society. Ultimately, the project seeks to strengthen social cohesion and improve outcomes for all participants in Australia's multicultural landscape \cite{cultural_atlas_2016}.

To ensure a holistic understanding of each culture, the Atlas categorizes information into the following key domains:

\begin{description}
    \item[Dates of Significance] Identifying pivotal religious, national, and cultural observances. This section provides historical context for community celebrations and aids in understanding the timing of significant cultural milestones.
    
    \item[Religion] An exploration of the spiritual frameworks and belief systems that shape a culture’s worldview. This includes how faith influences daily life, ethics, and community structures.
    
    \item[Etiquette] Guidelines on the social protocols and ``unwritten rules'' of interaction. This domain covers body language, gift-giving, and social taboos to facilitate respectful engagement.
    
    \item[Family] An analysis of the foundational social unit, focusing on kinship structures, gender roles, and authority dynamics. It examines the balance between individualistic and collectivistic values.
    
    \item[Greetings] A summary of appropriate verbal and non-verbal salutations. This includes the use of honorifics, physical gestures, and the level of formality required during initial contact.
\end{description}

\subsection{Similarity between countries on different categories} \label{app:category_similarity}

To study the similarity between countries and different categories,
We obtained cultural reference texts. We developed a Python web scraper that programmatically downloads cultural descriptions from the Cultural Atlas (culturalatlas.sbs.com.au), a publicly available cultural information resource. The script systematically retrieves articles for 10 countries (Brazil, Canada, Chile, China, Germany, India, Iran, Japan, Poland, and South Africa) across five cultural categories: greetings, etiquette, family, religion, and dates of significanceyielding up to 50 country–category text documents. For each page, the script fetches the raw HTML, strips non-content elements (e.g., navigation, scripts, and styling), and extracts the main article text from paragraph, list, and heading elements using BeautifulSoup. The extracted texts are saved as individual plain-text files, with metadata including the source country, category, and URL.

To quantify cultural similarity across nations, we computed semantic embeddings of the Cultural Atlas ground-truth texts using the all-MiniLM-L6-v2 \cite{reimers-2019-sentence-bert} sentence transformer model. Each country's cultural descriptionsspanning greetings, etiquette, family, religion, and dates of significancewere encoded into dense vector representations and averaged to produce a single composite embedding per country. Pairwise cosine similarity was then computed across all 10 countries to construct a country-by-country similarity matrix, visualized as a heatmap. Additionally, per-category cross-country similarity matrices were generated to examine how cultural proximity between nations varies across specific domains. This embedding-based approach enables a data-driven comparison of cultural textual content, revealing which countries share the most and least similar cultural profiles according to the Cultural Atlas reference material.

The embedding-based cross-country analysis revealed that cultural similarity varies substantially across domains. Greetings emerged as the most universally convergent category, with the highest average pairwise similarity (0.608), as cultures broadly describe similar conventions around handshakes, eye contact, and formal titles. In contrast, religion was the most divisive category (mean 0.445), exhibiting the widest similarity range (0.255–0.651), reflecting the fundamental diversity of religious traditions across the sampled nations. Family structures (mean 0.558) also showed relatively high cross-cultural similarity, as most cultures describe patriarchal structures, elder respect, and extended family systems, whereas etiquette (0.415) diverged sharply, suggesting that everyday social manners are more culturally specific than familial organization.

At the country-pair level, India and Japan recorded the single highest cross-country similarity score in any category (0.732 in greetings), driven by shared conventions such as bowing, hierarchy-based greeting depth, honorific suffixes, and the avoidance of physical contact with strangers. Notably, large similarity swings were observed within the same country pair across categories. For instance, Brazil and Canada scored 0.646 in greetings but only 0.299 in etiquette, a swing of 0.348, indicating that while both cultures greet similarly through handshakes and informality, their broader social norms diverge considerably. Similarly, Chile and India scored 0.656 in family but only 0.322 in religion, reflecting shared extended family structures alongside completely divergent religious traditions.

\begin{figure*}
    \centering
    
    \begin{subfigure}{0.48\textwidth}
        \centering
        \includegraphics[width=\textwidth]{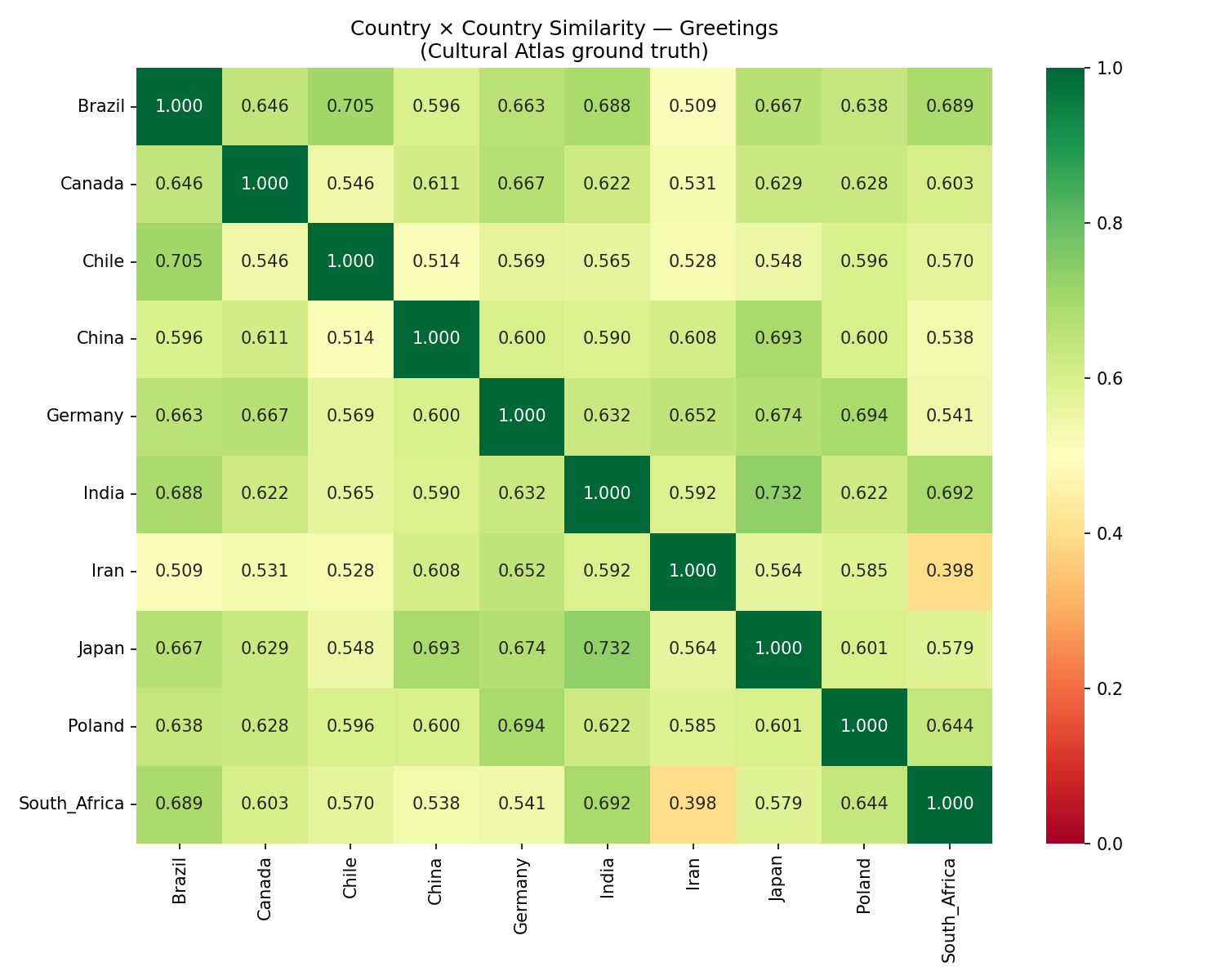}
        \caption{Greetings}
        \label{fig:heatmap_greetings}
    \end{subfigure}
    \hfill
    \begin{subfigure}{0.48\textwidth}
        \centering
        \includegraphics[width=\textwidth]{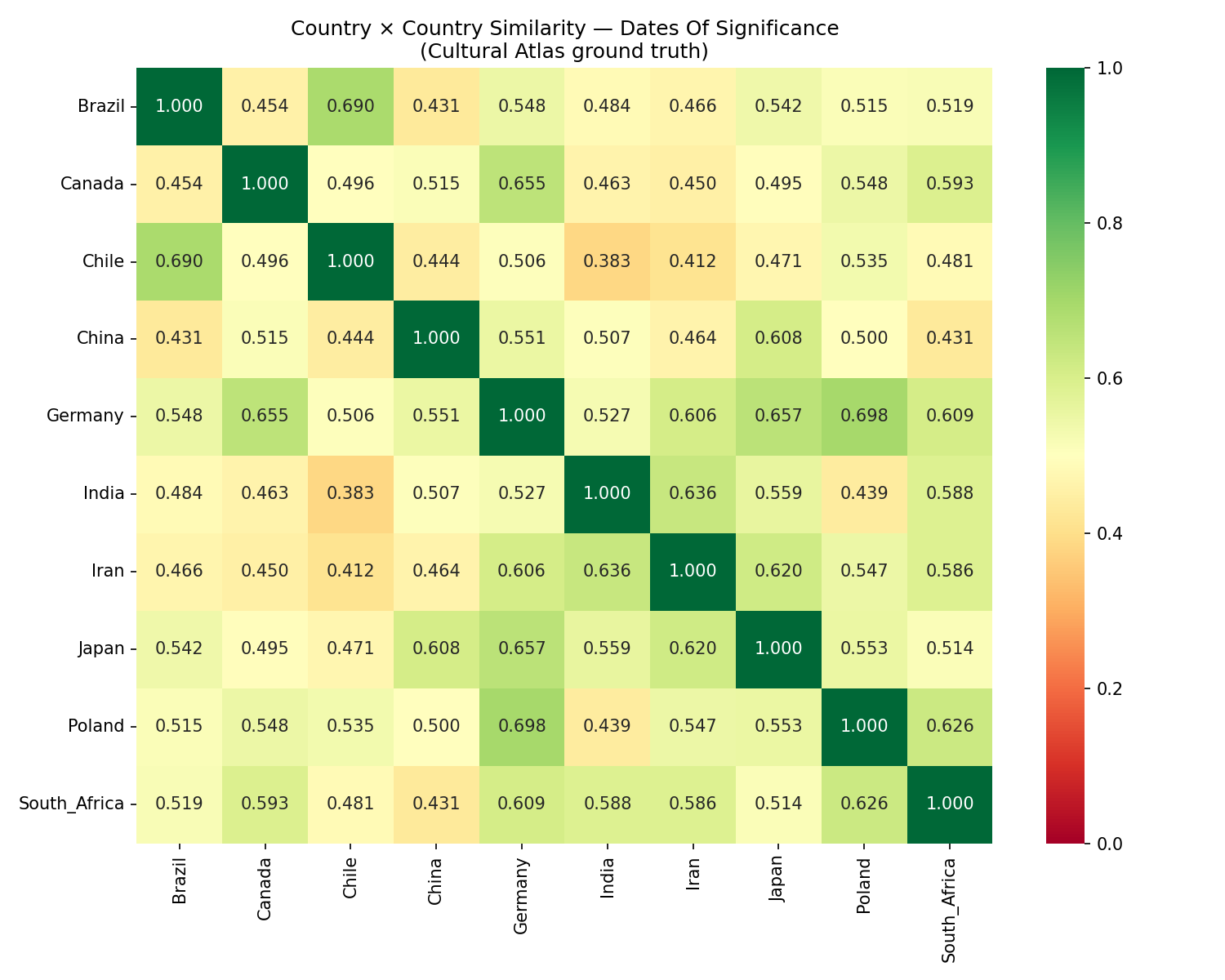}
        \caption{Dates of Significance}
        \label{fig:heatmap_date_of_significance}
    \end{subfigure}

    \vspace{1em} 

    \begin{subfigure}{0.48\textwidth}
        \centering
        \includegraphics[width=\textwidth]{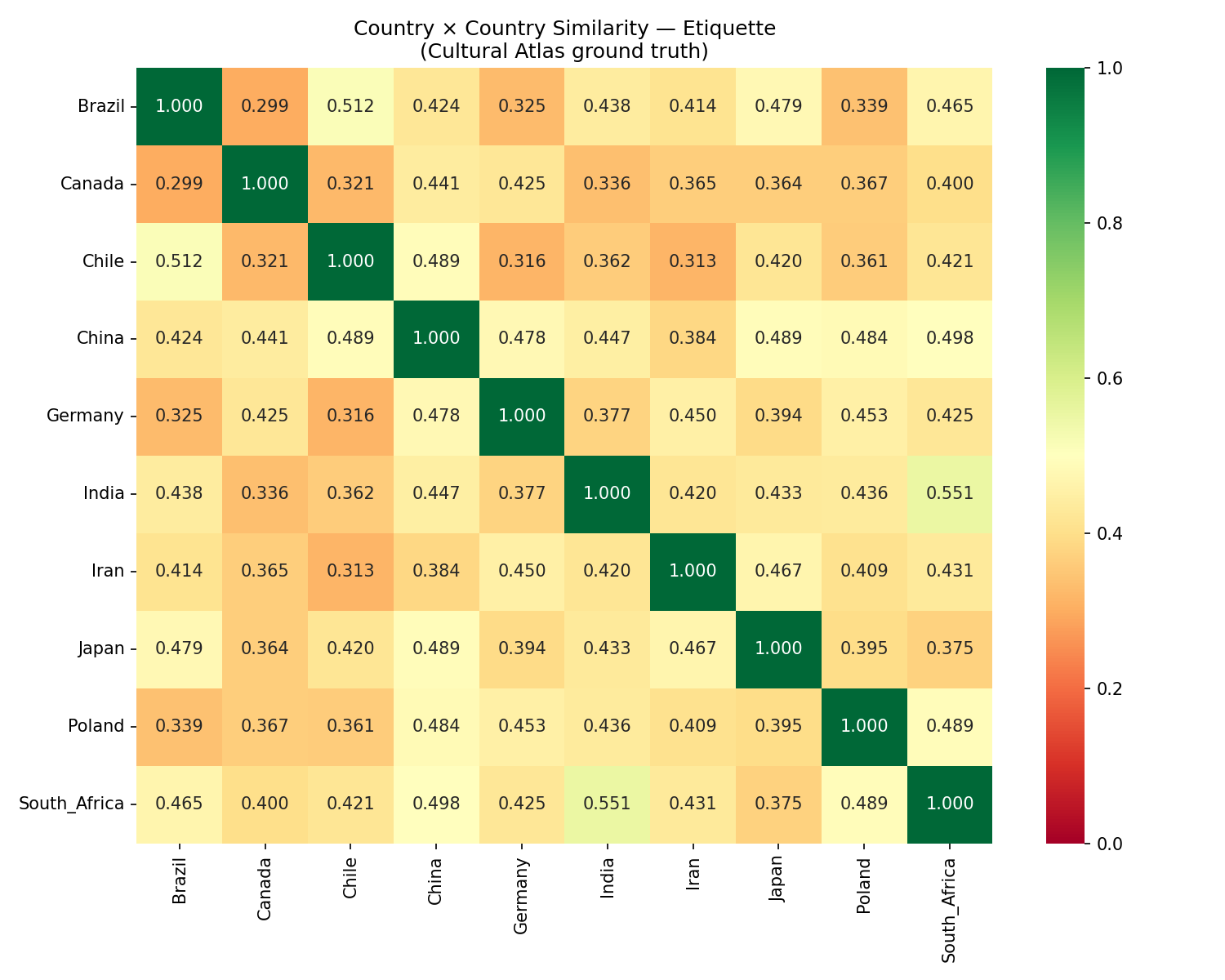}
        \caption{Etiquette}
        \label{fig:heatmap_etiquette}
    \end{subfigure}
    \hfill
    \begin{subfigure}{0.48\textwidth}
        \centering
        \includegraphics[width=\textwidth]{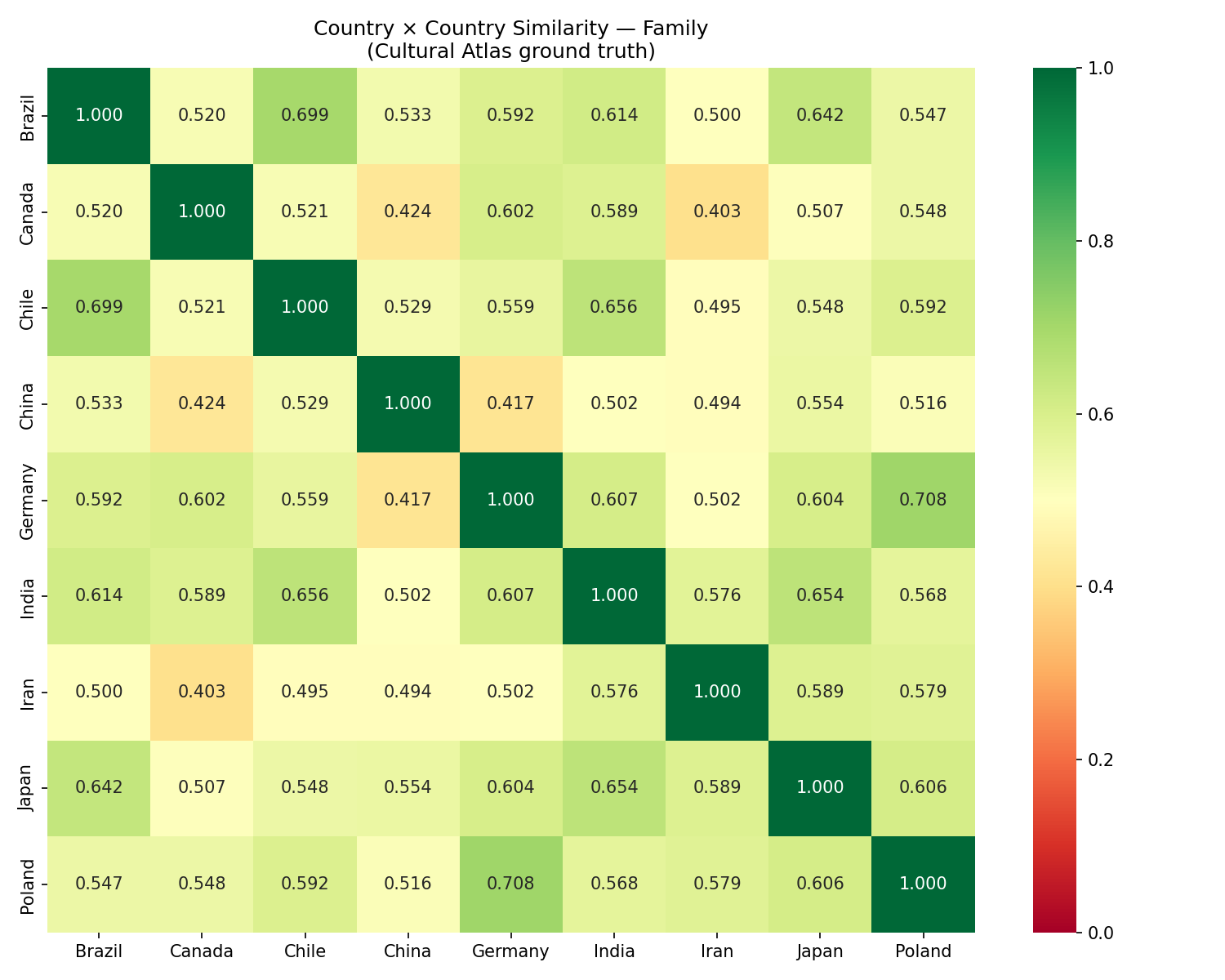}
        \caption{Family}
        \label{fig:heatmap_family}
    \end{subfigure}

    \vspace{1em}

    \begin{subfigure}{0.4\textwidth}
        \centering
        \includegraphics[width=\textwidth]{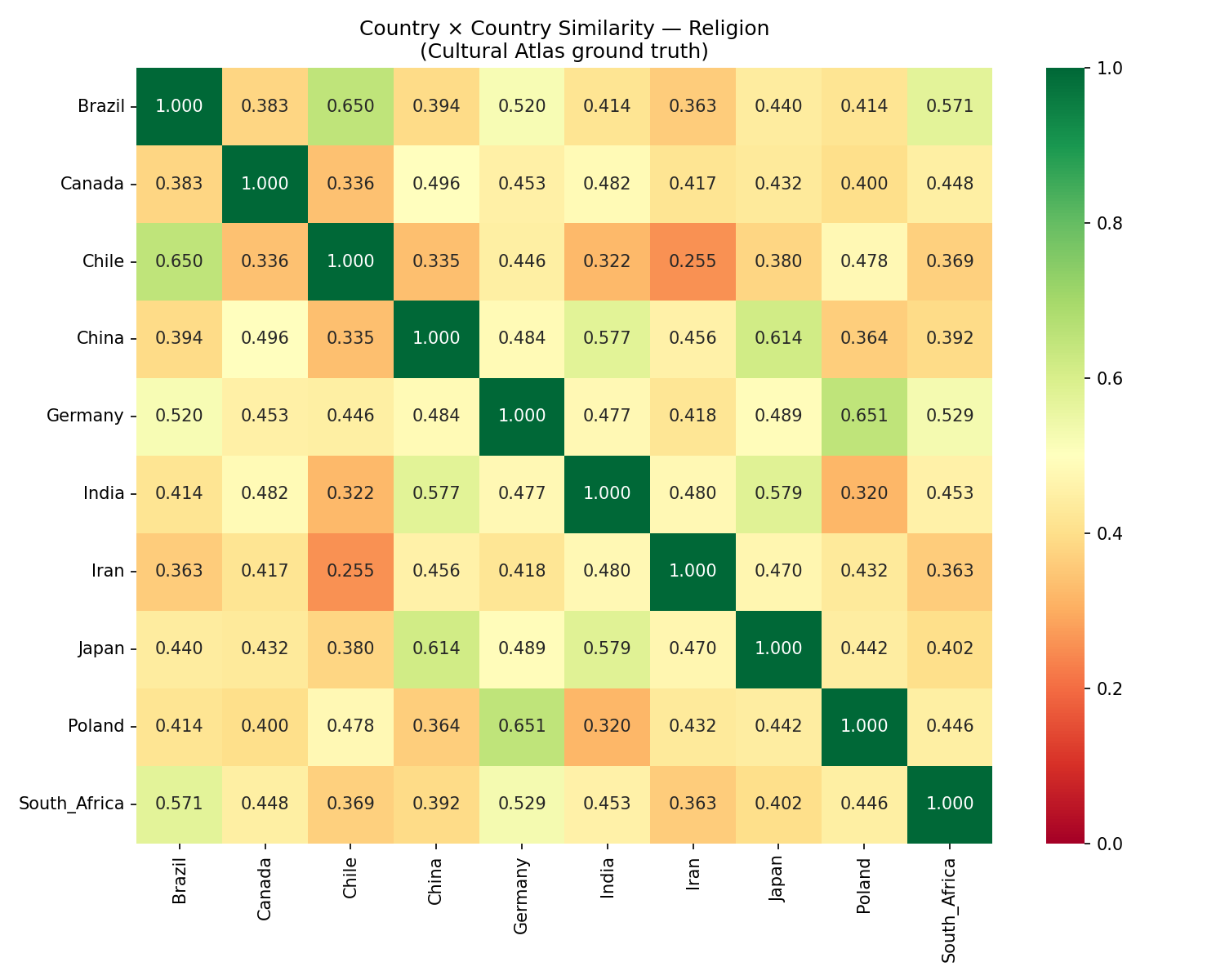}
        \caption{Religion}
        \label{fig:heatmap_religion}
    \end{subfigure}

    \caption{Heatmaps showing similarity between countries across different cultural categories.}
    \label{fig:all_heatmaps}
\end{figure*}

Several country-level patterns emerged as particularly noteworthy. Canada, despite its multicultural identity, was not a cultural middleman but rather an outlier in etiquette, recording the lowest average similarity to all other countries (0.369) in that category. Its informal, first-name-basis culture proved genuinely unique rather than a blend of other cultural norms. Chile stood out as the most culturally distinctive nation in both religion and dates of significance, likely driven by its Easter Island traditions such as the Tapati festival and Rapa Nui culture, combined with its specific Catholic-indigenous syncretism. Conversely, Germany emerged as the most central country in religion and dates of significance, with its cultural descriptions serving as a kind of baseline against which other nations are measured, likely because the Cultural Atlas frames German culture in broadly Western terms that partially overlap with many other cultures. Finally, the intra-region similarity gap was largest for religion, where pairs such as Germany–Poland, Brazil–Chile, and China–Japan shared religious traditions within their respective regions, while cross-region pairs like Chile–Iran (0.255) were almost entirely dissimilar. Please refer to Figure \ref{fig:all_heatmaps}.

\section{Annotation}
\label{app:annotation}

\subsection{Participant Recruitment}
We recruited a total of N = 45 native evaluators (5 each) across 9 countries (India, China, South Africa, Japan, Brazil, Chile, Canada, Poland, and Germany). Participants were sourced through Prolific, with the explicit requirement that each participant be a native resident or citizen of the country they were evaluating. Participants spanned diverse professional backgrounds, including entrepreneurs, musicians, homemakers, policymakers, and data analysts, to ensure that cultural judgments were not systematically skewed by technical familiarity with generative AI systems. Participation was voluntary and anonymous. Participants were paid 5USD for completing the study. No personally identifiable information was collected at any stage; all responses were used solely for academic research purposes.

\subsection{Annotation Platform and Setup}
Annotations were collected through the Prolific and Qualtrics platforms. Participants were required to complete the study on a laptop with a stable internet connection to ensure reliable video playback, and were instructed to work in a quiet environment where they could focus on visual details. The estimated completion time was 15 minutes. Each survey began with an informed consent question; participants were required to confirm that they understood and agreed to the terms of participation before proceeding.
A critical instruction was provided at the outset: participants were explicitly told to base all their judgments on video frames only, and not to consider audio in their assessments. This ensured that all evaluations were grounded purely in the visual cultural content generated by each model, which is the modality \name measures.

\subsection{Annotation Task Design}
\begin{figure*}
    \centering
    \includegraphics[width=1\textwidth]{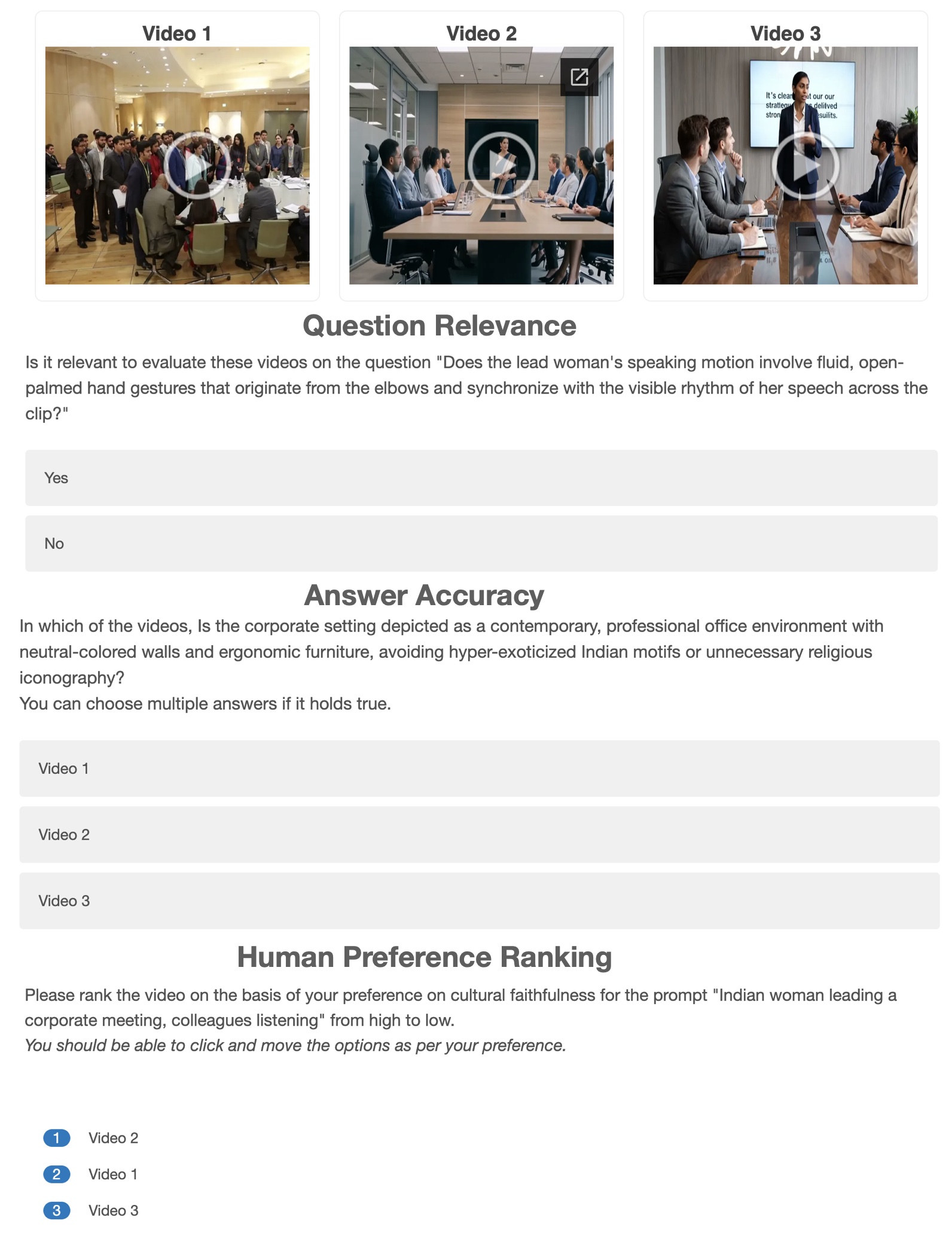}
    \caption{Annotation Platform-Humans annotated samples based on Question Relevance, Answer Accuracy, and Human Preference Ranking.}
    \label{fig:Annotation_Platform}
\end{figure*}

Each participant was assigned 3 video triplets, with each triplet corresponding to a single cultural prompt. Within each triplet, the three videos were generated by LTX-2, Wan 2.2, and Veo 3.1 Fast, respectively, and presented in randomized order to prevent position bias.
For each triplet, participants completed a three-tier assessment in the following fixed order, as illustrated in Figure \ref{fig:Annotation_Platform}:
(i) Question Relevance. Participants were shown a \name generated evaluation question drawn from the Identity, Behavior, or Context dimension alongside all three videos. They judged whether the question constituted a meaningful and appropriate probe of cultural faithfulness for the given prompt, responding with a binary Yes/No. Questions marked as irrelevant by a majority of raters within a cohort were excluded from further analysis.
(ii) Answer Accuracy. Participants were then asked to answer the same cultural question based strictly on visual evidence in the videos, selecting all videos for which the answer was affirmatively Yes. Multiple selections were permitted if the condition held true for more than one video. This tier assesses whether the cultural marker in question is visually legible to a native observer, independently of overall aesthetic preference.
(iii) Human Preference Ranking. Finally, participants ranked all three videos from 1st to 3rd based on overall cultural faithfulness to the given prompt, using a drag-and-drop interface. Participants were explicitly instructed to rank based on cultural accuracy rather than visual quality or cinematic appeal.
Participants were required to complete all three tiers before proceeding to the next triplet.
Figure \ref{fig:Annotation_Platform} presents a screenshot from the annotation platform.

\section{Prompt Specifications}
\label{app:prompt}

In the following, we provide details on all the prompts used in this research.

\subsection{Identity, Behavior, Context }\label{app:IBC}

The first stage decomposes each base cultural prompt into its three constituent dimensions: Identity, Behavior, and Context. This decomposition is a prerequisite for all downstream stages, as it provides the structured inputs used in extended prompt construction.

\begin{tcolorbox}[colback=gray!5!white, colframe=gray!75!black,
                  title={System Prompt: IBC Extractor}, arc=3mm, breakable]

Analyze the image prompt and extract Identity, Behavior, and Context using these rules:
\begin{enumerate}
    \item \textbf{Identity}: The primary human or living subject(s).
    \item \textbf{Behavior}: The physical activity, gesture, or movement.
    \item \textbf{Context}: The setting, atmosphere, and background details (including what is being watched).
\end{enumerate}

\textbf{EXAMPLES:} \\
Prompt: ``Indian family greeting elder with feet-touching gesture at home'' \\
Output: \{\texttt{"Identity"}: ``Indian family'', \texttt{"Behavior"}: ``Greeting elder with feet-touching gesture'', \texttt{"Context"}: ``Indian home''\}

\vspace{0.5em}
Prompt: ``Germans watching New Year's concert on television'' \\
Output: \{\texttt{"Identity"}: ``Germans'', \texttt{"Behavior"}: ``Watching'', \texttt{"Context"}: ``New Year's concert on television''\}

\vspace{1em}
\textbf{TASK:} \\
Prompt: ``\{prompt\_text\}'' \\
Return ONLY a JSON object with keys \texttt{"Identity"}, \texttt{"Behavior"}, and \texttt{"Context"}.
\end{tcolorbox}

\subsection{Extended Prompting}\label{app:Extended}
This stage constructs the Extended Prompt variant used as the primary evaluation condition in our experiments. Extended prompts augment the base prompt with explicit, dictionary-grounded visual descriptions for each of the three IBC dimensions, giving video generation models detailed guidance on what Identity, Behavior, and Context should look like in the generated video.

\begin{tcolorbox}[colback=gray!5!white, colframe=gray!75!black,
                  title={System Prompt: Extended Prompting}, arc=3mm, breakable]
\textbf{TASK:} \\
Create a detailed video generation prompt. You must extend the `Behavior',
`Identity', and `Context' using their formal definitions grounded in the
Oxford Dictionary.

\vspace{0.5em}
\textbf{INPUTS:}
\begin{itemize}
    \item Prompt: \{\texttt{prompt\_text}\}
    \item Behavior: \{\texttt{Behavior}\}
    \item Identity: \{\texttt{Identity}\}
    \item Context: \{\texttt{Context}\}
\end{itemize}

\vspace{0.5em}
\textbf{FORMAT:} \\
``Cinematic shot. \{\texttt{prompt\_text}\}. [Behavior]: [Detailed visual
description of the Behavior: \{\texttt{Behavior}\} from Oxford Dictionary],
[Identity]: [Detailed visual description of the Identity:
\{\texttt{Identity}\} from Oxford Dictionary], and [Context]: [Detailed visual
description of the Context: \{\texttt{Context}\} from Oxford Dictionary].''

\vspace{1em}
\textbf{OUTPUT:} \\
Return ONLY the final string following the format above. Do not include labels
like `Output:' or extra commentary.
\end{tcolorbox}

\subsection {Geographical Constraint Removed Prompting}\label{app:no_ctry}

This stage constructs the Geographical Constraint Removed variant, used exclusively in the implicit cultural knowledge ablation (RQ3). This prompt variant is produced by stripping explicit country-level identifiers from the base prompt while preserving all cultural, religious, and lifestyle descriptors that do not name a specific nation.

\begin{tcolorbox}[
    colback=gray!5!white,
    colframe=gray!75!black,
    title={System Prompt: Geographical Constraint Removed Prompting},
    arc=3mm,
    fonttitle=\bfseries,
    breakable
]
\textbf{TASK:} \\
Rewrite the following prompt to remove any mention of specific countries
(e.g., India, Indian, Germany, German, USA, American, etc.).

\vspace{0.5em}
\textbf{IMPORTANT:} Keep religious, cultural, or lifestyle descriptors
(e.g., Muslim, Halal, Festive, Buddhist, Vegan).

\vspace{0.5em}
\textbf{EXAMPLES:} \\
\textbf{Input:} ``Indian Muslim family enjoying a halal feast at home'' \\
\textbf{Output:} ``Muslim family enjoying a halal feast at home''

\vspace{0.5em}
\textbf{Input:} ``German elderly couple walking in a park'' \\
\textbf{Output:} ``Elderly couple walking in a park''

\vspace{0.5em}
\textbf{Input:} ``A traditional Japanese tea ceremony in Tokyo'' \\
\textbf{Output:} ``A traditional tea ceremony in Tokyo''

\vspace{1em}
\textbf{INPUT:} \\
``\{prompt\_text\}'' \\
Return ONLY the rewritten string. No explanations.
\end{tcolorbox}

\subsection{Question Generator}\label{app:question_gen}

This stage generates the culturally grounded question-answer pairs used to evaluate generated videos. For each extended prompt, a language model acting as a cultural anthropologist and visual evaluator produces a set of Yes/No evaluation questions decomposed across the Identity, Behavior, and Context dimensions. In total, this stage produces 9,288 question-answer pairs across all prompts and models.

\begin{tcolorbox}[colback=gray!5!white, colframe=gray!75!black,
                  title={System Prompt: Question Generation}, arc=3mm, breakable]
You are an expert cultural anthropologist, visual evaluator, and prompt
engineer. Your task is to analyze a given video generation prompt and
systematically generate an exhaustive evaluation framework to measure the
cultural faithfulness of the resulting generated video.

Decompose the prompt into three categories: Context, Identity, and Behavior.

\vspace{0.5em}
\textbf{Strict Guidelines:}
\begin{enumerate}
    \item \textbf{Yes/No Format:} Every question must be structured so that a
    ``Yes'' indicates cultural faithfulness and a ``No'' indicates a failure or
    cultural inaccuracy.
    \item \textbf{Embedded Visual Descriptions:} Embed precise visual,
    physical, or spatial descriptions directly into the question. Do not assume
    the evaluating model has implicit cultural knowledge. (e.g., instead of
    ``Is the person wearing a traditional Kimono?'', ask ``Is the person
    wearing a traditional Kimono, characterized by left-over-right wrapped
    lapels and wide, square-cut sleeves?'').
    \item \textbf{Temporal Grounding (Crucial for Behaviors):} Questions MUST
    explicitly probe the sequence, duration, physics, or progression of the
    movement across frames. (e.g., ``Does the pouring behavior begin with the
    vessel held low, smoothly rise to a higher elevation, and return low
    without breaking the stream?'').
    \item \textbf{Explicit vs. Implicit:} Generate questions for explicitly
    requested elements, and mandatory implicit elements required for cultural
    authenticity. Actively avoid Western-centric stereotypes (e.g., implicitly
    checking that a traditional daily context avoids hyper-exoticized or
    religious backdrops unless the context or the prompt requires them).
    \item \textbf{Weighting Strategy (1--10 Scale):}
    \begin{itemize}
        \item \textbf{8--10:} Critical explicit elements or absolute cultural
        boundaries.
        \item \textbf{4--7:} Expected contextual elements and secondary
        identities that enhance authenticity.
        \item \textbf{1--3:} Minor background details or high-fidelity nuances.
    \end{itemize}
\end{enumerate}
\end{tcolorbox}

\subsection{Question Verifier}\label{app:question_verifier}

This stage filters out questions that, despite being generated from culturally validated prompts, make claims about cultural elements that are factually inaccurate or not genuinely grounded in the target culture. This step is necessary because the question generator in Stage \ref{app:question_gen} can occasionally produce questions that embed incorrect or hallucinated cultural details. For example, attributing a practice to a country where it does not originate would corrupt the evaluation signal.

\begin{tcolorbox}[colback=gray!5!white, colframe=gray!75!black,
                  title={System Prompt: Question Verifier}, arc=3mm, breakable]
\textbf{TASK:} \\
The following question describes a visual context. Verify if the cultural
elements mentioned (identities, practices, symbols, customs) are grounded in
actual \{\texttt{culture}\} cultural practices or facts:

\vspace{0.5em}
Question: ``\{\texttt{question}\}''

\vspace{0.5em}
Focus on whether the specific cultural elements mentioned are real and accurate
for \{\texttt{culture}\} culture.

\vspace{1em}
\textbf{OUTPUT FORMAT (Strict JSON, no extra text before or after):}

{\footnotesize
\begin{verbatim}
{
  "grounded": "Yes" or "No",
  "evidence": "one-sentence summary
     confirming or denying the
     cultural element is authentic"
}
\end{verbatim}
}
\end{tcolorbox}

\subsection{Answer Generator}\label{app:answer_generator}

This stage evaluates each generated video against the verified question set from Stage \ref{app:question_verifier}. A vision-language model (Qwen3-VL-235B) is provided with a generated video and a single cultural question, and is asked to reason carefully before producing a binary Yes/No answer indicating whether the video is culturally faithful with respect to what the question asks.

\begin{tcolorbox}[colback=gray!5!white, colframe=gray!75!black,
                  title={System Prompt: Answer Generator}, arc=3mm, breakable]
You are an expert cultural anthropologist and visual evaluator assessing the
cultural faithfulness of a generated video.

When answering each question, you MUST reason within \texttt{<think>}
\texttt{</think>} tags following these steps:
\begin{enumerate}
    \item \textbf{Identify the visual evidence:} Describe exactly what you
    observe in the video frames---specific identities, clothing details,
    spatial arrangements, architectural elements, lighting, and colors.
    \item \textbf{Assess cultural accuracy:} Compare your observations against
    the culturally specific visual descriptions embedded in the question. Do
    not rely on implicit cultural knowledge---only evaluate what the question
    explicitly describes.
    \item \textbf{Evaluate temporal and physical coherence (for Behavior
    questions):} Examine the sequence, duration, physics, and progression of
    movements across frames. Note whether behaviors follow the temporal
    grounding specified in the question.
    \item \textbf{Check for stereotyping or inauthenticity:} Flag if the video
    substitutes Western-centric defaults, hyper-exoticized elements, or generic
    representations in place of the specific cultural markers described in the
    question.
\end{enumerate}

After your reasoning, provide the final answer as either Yes or No. ``Yes''
means the video is culturally faithful for what the question asks. ``No'' means
it fails or is culturally inaccurate.

The final answer MUST BE put in a box. For example: $\boxed{\text{Yes}}$ or
$\boxed{\text{No}}$.
\end{tcolorbox}

\subsection{Cultural Uniqueness Classifier} \label{app: Geographical_constraint_verifier}

This stage is a filtering step applied exclusively in the Geographical Constraint Removed ablation (RQ3). After country names are stripped from base prompts, not all remaining prompts are equally informative for probing implicit cultural knowledge. Some de-anchored prompts describe activities so universal — such as "a family having dinner" or "people watching television" — that no model, regardless of cultural competence, could reasonably be expected to generate a culturally specific output from them. Including such prompts in the ablation would dilute the signal and misattribute generic outputs as cultural failures.

\begin{tcolorbox}[colback=gray!5!white, colframe=gray!75!black,
                  title={System Prompt: Cultural Uniqueness Classifier},
                  arc=3mm, breakable]
You are a cultural analyst specializing in identifying whether a scene
description is distinctly tied to a specific country or culture.

\textbf{Your task:} Given a scene description (with the country name removed)
and the target country, decide whether the described scene is
\textbf{culturally unique} to that country.

\textbf{Definition of ``culturally unique'':}
A prompt is culturally unique if it contains at least one element---such as a
specific holiday, ritual, tradition, food, object, language term, landmark, or
custom---that is distinctly and strongly associated with the given country. A
casual observer familiar with world cultures would recognize it as belonging to
that country.

\textbf{A prompt is NOT culturally unique if:}
\begin{itemize}
    \item The scene could plausibly occur in any country (e.g., ``a family having dinner'', ``friends meeting at a cafe'')
    \item The activity is common across many cultures (e.g., ``people watching TV'', ``a wedding ceremony'')
    \item Only the time or season makes it distinct, not the cultural elements (e.g., ``family watching fireworks on New Year's Eve'')
    \item The elements are broadly Western/universal rather than country-specific
    \item The tradition or practice is shared across several neighboring countries without a country-specific distinguishing detail
\end{itemize}

\textbf{Examples:}

\textit{Country: Germany} \textbar{} \textit{Prompt: ``Couple meeting at a sports club gathering''} \\
$\rightarrow$ \texttt{is\_culturally\_unique: false} \\
$\rightarrow$ Sports club gatherings happen in virtually every country.

\textit{Country: Japan} \textbar{} \textit{Prompt: ``Tea ceremony with guests appreciating matcha''} \\
$\rightarrow$ \texttt{is\_culturally\_unique: true} \\
$\rightarrow$ The Japanese tea ceremony (\textit{chado/chanoyu}) is a codified cultural ritual unique to Japan.

\textit{Country: Poland} \textbar{} \textit{Prompt: ``Family sharing the op\l{}atek wafer before Christmas Eve dinner''} \\
$\rightarrow$ \texttt{is\_culturally\_unique: true} \\
$\rightarrow$ Sharing \textit{op\l{}atek} before the \textit{Wigilia} supper is a distinctly Polish Catholic tradition.

Apply the same careful reasoning to the input you receive.
\end{tcolorbox}

\section{Experiments} \label{app:experiments}

\subsection{Model Specifications}

\paragraph{Wan2.2 \cite{wan2025wanopenadvancedlargescale}.}
We use the \texttt{Wan2.2-T2V-A14B} checkpoint for the text-to-video task.
Generations are configured at a resolution of $1280 \times 720$ with a frame
count of 49 to produce 5-second clips. The sampling process uses 20 steps with
\texttt{convert\_model\_dtype} enabled and the T5 encoder running on CPU
(\texttt{--t5\_cpu}) to manage memory overhead. We generate a total of 2{,}943
videos ($981 \times 3$ variations) for this model.

\paragraph{LTX-2 \cite{hacohen2024ltxvideorealtimevideolatent}.}
We employ a multi-component setup to ensure maximum output quality, comprising
the \texttt{ltx-2-19b-\allowbreak dev-fp8.safetensors} base checkpoint, the
\texttt{gemma-3-12b-\allowbreak it-qat-q4\_0-\allowbreak unquantized} text
encoder, and \texttt{ltx-2-19b-\allowbreak distilled-lora-384.safetensors} for
enhanced distillation. To achieve the final resolution, we apply the
\texttt{ltx-2-\allowbreak spatial-upscaler-\allowbreak x2-1.0.safetensors}
spatial upsampler. As with Wan2.2, this model is evaluated across all 2{,}943
prompt instances.

\paragraph{Veo 3.1 Fast \cite{GoogleDeepMind2025Veo3}.}
We use the \texttt{veo-3.1-fast-generate-001} model via the Google DeepMind
API. Due to the high operational costs associated with the Veo API, we do not
run the full suite of 981 prompts. Instead, we take a stratified sample across
cultural categories to generate 288 videos, which supports a meaningful
comparison with the other models while remaining within computational and
budgetary constraints. In total, our generation pipeline produced 6{,}174
videos for evaluation.

\subsection{VideoScore across Models} \label{app:videoscore}
VideoScore \cite{he2024videoscore} is an automated evaluation metric designed to simulate fine-grained human feedback for generative video models. It calculates scores across: Visual Quality (spatial clarity, resolution, and aesthetic appeal), Temporal Consistency (the absence of flickering, warping, or sudden object morphing across frames), Dynamic Degree (the presence of fluid and significant motion), Text-to-Video Alignment (how accurately the visual content matches the semantic intent of the text prompt), and Factual Consistency (adherence to physical laws and common-sense logic).
In the following Tables \ref{tab:ltx_avg_base}, \ref{tab:ltx_avg_extended}, and \ref{tab:ltx_avg_no_country}, we have reported average videoscore across all five dimensions for LTX-2 \cite{hacohen2024ltxvideorealtimevideolatent}.
For Wan2.2 \cite{wan2025wanopenadvancedlargescale}, we have reported scores in Table \ref{tab:wan_avg_base_scores}. For Veo 3.1 \cite{GoogleDeepMind2025Veo3}, we have reported scores in Table \ref{tab:veo_avg_base_scores}, \ref{tab:veo_avg_extended_scores} and \ref{tab:veo_avg_no_country_scores}.

\begin{table*}[ht]
\centering
\vspace{4pt}
\resizebox{\textwidth}{!}{%
\begin{tabular}{ll CCCCC}
\toprule
\textbf{Country} & \textbf{Category} & \textbf{Visual Quality} & \textbf{Temporal Consistency} & \textbf{Dynamic Degree} & \textbf{Text-to-Video Alignment} & \textbf{Factual Consistency} \\
\midrule
Brazil & dates-of-significance & {3.485} & {3.445} & {3.431} & {3.243} & {3.496} \\
Brazil & etiquette & {3.685} & {3.679} & {3.627} & {3.555} & {3.674} \\
Brazil & family & {3.547} & {3.471} & {3.555} & {3.394} & {3.496} \\
Brazil & greetings & {3.621} & {3.582} & {3.65} & {3.488} & {3.619} \\
Brazil & religion & {3.528} & {3.525} & {3.456} & {3.236} & {3.55} \\
\midrule
Canada & dates-of-significance & {3.624} & {3.55} & {3.533} & {3.274} & {3.604} \\
Canada & etiquette & {3.58} & {3.549} & {3.504} & {3.27} & {3.56} \\
Canada & family & {3.401} & {3.332} & {3.396} & {3.187} & {3.376} \\
Canada & greetings & {3.673} & {3.622} & {3.695} & {3.489} & {3.655} \\
Canada & religion & {3.363} & {3.329} & {3.311} & {3.045} & {3.391} \\
\midrule
Chile & dates-of-significance & {3.376} & {3.304} & {3.356} & {3.129} & {3.389} \\
Chile & etiquette & {3.578} & {3.533} & {3.527} & {3.407} & {3.554} \\
Chile & family & {3.293} & {3.168} & {3.299} & {3.025} & {3.246} \\
Chile & greetings & {3.608} & {3.57} & {3.609} & {3.46} & {3.596} \\
Chile & religion & {3.437} & {3.367} & {3.317} & {3.043} & {3.432} \\
\midrule
China & dates-of-significance & {3.615} & {3.579} & {3.57} & {3.373} & {3.605} \\
China & etiquette & {3.775} & {3.713} & {3.704} & {3.567} & {3.746} \\
China & family & {3.791} & {3.728} & {3.753} & {3.546} & {3.76} \\
China & greetings & {3.65} & {3.56} & {3.609} & {3.416} & {3.638} \\
China & religion & {3.715} & {3.697} & {3.507} & {3.416} & {3.74} \\
\midrule
Germany & dates-of-significance & {3.281} & {3.238} & {3.152} & {3.093} & {3.285} \\
Germany & etiquette & {3.387} & {3.327} & {3.399} & {3.299} & {3.373} \\
Germany & family & {3.287} & {3.264} & {3.208} & {3.156} & {3.293} \\
Germany & greetings & {3.559} & {3.52} & {3.513} & {3.418} & {3.557} \\
Germany & religion & {3.007} & {3.081} & {2.859} & {2.906} & {3.094} \\
\midrule
India & dates-of-significance & {3.57} & {3.518} & {3.527} & {3.307} & {3.561} \\
India & etiquette & {3.668} & {3.614} & {3.605} & {3.488} & {3.646} \\
India & family & {3.562} & {3.493} & {3.517} & {3.373} & {3.552} \\
India & greetings & {3.668} & {3.62} & {3.596} & {3.472} & {3.671} \\
India & religion & {3.726} & {3.678} & {3.658} & {3.467} & {3.715} \\
\midrule
Iran & dates-of-significance & {3.623} & {3.582} & {3.557} & {3.335} & {3.63} \\
Iran & etiquette & {3.796} & {3.762} & {3.763} & {3.605} & {3.775} \\
Iran & family & {3.673} & {3.64} & {3.637} & {3.472} & {3.67} \\
Iran & greetings & {3.877} & {3.844} & {3.866} & {3.662} & {3.863} \\
Iran & religion & {3.624} & {3.57} & {3.605} & {3.415} & {3.634} \\
\midrule
Japan & dates-of-significance & {3.283} & {3.238} & {3.164} & {3.021} & {3.29} \\
Japan & etiquette & {3.595} & {3.514} & {3.521} & {3.394} & {3.545} \\
Japan & family & {3.171} & {3.169} & {3.042} & {3.008} & {3.2} \\
Japan & greetings & {3.467} & {3.447} & {3.345} & {3.271} & {3.488} \\
Japan & religion & {3.054} & {3.064} & {2.796} & {2.771} & {3.071} \\
\midrule
Poland & dates-of-significance & {3.377} & {3.33} & {3.217} & {3.044} & {3.364} \\
Poland & etiquette & {3.357} & {3.317} & {3.241} & {3.141} & {3.338} \\
Poland & family & {3.29} & {3.174} & {3.236} & {3.078} & {3.254} \\
Poland & greetings & {3.138} & {3.126} & {3.011} & {2.988} & {3.162} \\
Poland & religion & {3.075} & {3.081} & {2.876} & {2.874} & {3.116} \\
\midrule
South\_Africa & dates-of-significance & {3.645} & {3.607} & {3.54} & {3.357} & {3.641} \\
South\_Africa & etiquette & {3.624} & {3.591} & {3.549} & {3.366} & {3.603} \\
South\_Africa & greetings & {3.655} & {3.624} & {3.651} & {3.471} & {3.654} \\
South\_Africa & religion & {3.54} & {3.547} & {3.472} & {3.223} & {3.587} \\
\bottomrule
\end{tabular}%
}
\caption{Average VideoScore (LTX-2 \cite{hacohen2024ltxvideorealtimevideolatent}) for base prompts, grouped by country and category.}
\label{tab:ltx_avg_base}
\end{table*}

\begin{table*}[ht]
\centering
\vspace{4pt}
\resizebox{\textwidth}{!}{%
\begin{tabular}{ll CCCCC}
\toprule
\textbf{Country} & \textbf{Category} & \textbf{Visual Quality} & \textbf{Temporal Consistency} & \textbf{Dynamic Degree} & \textbf{Text-to-Video Alignment} & \textbf{Factual Consistency} \\
\midrule
Brazil & dates-of-significance & {3.579} & {3.547} & {3.586} & {3.434} & {3.583} \\
Brazil & etiquette & {3.386} & {3.322} & {3.416} & {3.291} & {3.367} \\
Brazil & family & {3.263} & {3.259} & {3.271} & {3.216} & {3.297} \\
Brazil & greetings & {3.465} & {3.431} & {3.493} & {3.385} & {3.462} \\
Brazil & religion & {3.646} & {3.628} & {3.625} & {3.451} & {3.657} \\
\midrule
Canada & dates-of-significance & {3.576} & {3.516} & {3.544} & {3.338} & {3.563} \\
Canada & etiquette & {3.302} & {3.274} & {3.21} & {3.193} & {3.299} \\
Canada & family & {3.268} & {3.207} & {3.269} & {3.164} & {3.231} \\
Canada & greetings & {3.205} & {3.203} & {3.25} & {3.194} & {3.222} \\
Canada & religion & {3.541} & {3.519} & {3.523} & {3.404} & {3.555} \\
\midrule
Chile & dates-of-significance & {3.365} & {3.304} & {3.379} & {3.186} & {3.363} \\
Chile & etiquette & {3.252} & {3.228} & {3.18} & {3.15} & {3.209} \\
Chile & family & {3.287} & {3.225} & {3.329} & {3.223} & {3.277} \\
Chile & greetings & {3.442} & {3.387} & {3.454} & {3.306} & {3.433} \\
Chile & religion & {3.516} & {3.487} & {3.514} & {3.366} & {3.531} \\
\midrule
China & dates-of-significance & {3.543} & {3.524} & {3.459} & {3.328} & {3.53} \\
China & etiquette & {3.773} & {3.727} & {3.741} & {3.57} & {3.743} \\
China & family & {3.703} & {3.688} & {3.654} & {3.463} & {3.699} \\
China & greetings & {3.504} & {3.47} & {3.413} & {3.346} & {3.525} \\
China & religion & {3.482} & {3.49} & {3.281} & {3.272} & {3.503} \\
\midrule
Germany & dates-of-significance & {3.281} & {3.21} & {3.277} & {3.088} & {3.259} \\
Germany & etiquette & {3.035} & {2.994} & {3.003} & {2.989} & {3.001} \\
Germany & family & {2.995} & {2.974} & {3.008} & {3.005} & {2.97} \\
Germany & greetings & {3.03} & {3.018} & {3.04} & {3.039} & {3.052} \\
Germany & religion & {3.311} & {3.341} & {3.142} & {3.117} & {3.341} \\
\midrule
India & dates-of-significance & {3.448} & {3.421} & {3.34} & {3.252} & {3.451} \\
India & etiquette & {3.56} & {3.508} & {3.483} & {3.377} & {3.553} \\
India & family & {3.586} & {3.51} & {3.52} & {3.352} & {3.583} \\
India & greetings & {3.411} & {3.406} & {3.271} & {3.264} & {3.442} \\
India & religion & {3.387} & {3.405} & {3.233} & {3.234} & {3.437} \\
\midrule
Iran & dates-of-significance & {3.453} & {3.432} & {3.414} & {3.279} & {3.456} \\
Iran & etiquette & {3.669} & {3.641} & {3.623} & {3.528} & {3.656} \\
Iran & family & {3.277} & {3.239} & {3.259} & {3.138} & {3.243} \\
Iran & greetings & {3.518} & {3.501} & {3.518} & {3.405} & {3.541} \\
Iran & religion & {3.459} & {3.479} & {3.366} & {3.341} & {3.483} \\
\midrule
Japan & dates-of-significance & {3.352} & {3.328} & {3.229} & {3.176} & {3.353} \\
Japan & etiquette & {3.432} & {3.383} & {3.381} & {3.305} & {3.413} \\
Japan & family & {3.332} & {3.304} & {3.31} & {3.22} & {3.34} \\
Japan & greetings & {3.126} & {3.147} & {2.875} & {3.011} & {3.169} \\
Japan & religion & {3.203} & {3.217} & {2.992} & {3.039} & {3.223} \\
\midrule
Poland & dates-of-significance & {3.35} & {3.298} & {3.312} & {3.164} & {3.341} \\
Poland & etiquette & {3.21} & {3.184} & {3.189} & {3.142} & {3.189} \\
Poland & family & {3.299} & {3.214} & {3.338} & {3.238} & {3.284} \\
Poland & greetings & {3.241} & {3.166} & {3.283} & {3.156} & {3.247} \\
Poland & religion & {3.328} & {3.318} & {3.204} & {3.157} & {3.325} \\
\midrule
South\_Africa & dates-of-significance & {3.546} & {3.521} & {3.506} & {3.33} & {3.566} \\
South\_Africa & etiquette & {3.137} & {3.138} & {3.097} & {3.032} & {3.165} \\
South\_Africa & greetings & {3.454} & {3.461} & {3.429} & {3.34} & {3.491} \\
South\_Africa & religion & {3.508} & {3.511} & {3.434} & {3.195} & {3.559} \\
\bottomrule
\end{tabular}%
}
\caption{Average VideoScore (LTX-2\cite{hacohen2024ltxvideorealtimevideolatent}) for extended prompts, grouped by country and category.}
\label{tab:ltx_avg_extended}
\end{table*}

\begin{table*}[ht]
\centering
\vspace{4pt}
\resizebox{\textwidth}{!}{%
\begin{tabular}{ll CCCCC}
\toprule
\textbf{Country} & \textbf{Category} & \textbf{Visual Quality} & \textbf{Temporal Consistency} & \textbf{Dynamic Degree} & \textbf{Text-to-Video Alignment} & \textbf{Factual Consistency} \\
\midrule
Brazil & dates-of-significance & {3.438} & {3.381} & {3.375} & {3.18} & {3.429} \\
Brazil & etiquette & {3.611} & {3.605} & {3.55} & {3.489} & {3.61} \\
Brazil & family & {3.488} & {3.388} & {3.497} & {3.367} & {3.447} \\
Brazil & greetings & {3.54} & {3.46} & {3.607} & {3.376} & {3.525} \\
Brazil & religion & {3.349} & {3.334} & {3.244} & {2.961} & {3.361} \\
\midrule
Canada & dates-of-significance & {3.539} & {3.505} & {3.412} & {3.273} & {3.547} \\
Canada & etiquette & {3.587} & {3.519} & {3.557} & {3.266} & {3.545} \\
Canada & family & {3.223} & {3.147} & {3.233} & {3.052} & {3.175} \\
Canada & greetings & {3.496} & {3.442} & {3.498} & {3.402} & {3.497} \\
Canada & religion & {2.975} & {2.975} & {2.84} & {2.787} & {3.062} \\
\midrule
Chile & dates-of-significance & {3.268} & {3.163} & {3.276} & {3.05} & {3.254} \\
Chile & etiquette & {3.336} & {3.312} & {3.196} & {3.12} & {3.319} \\
Chile & family & {3.148} & {3.059} & {3.112} & {2.923} & {3.124} \\
Chile & greetings & {3.54} & {3.474} & {3.513} & {3.397} & {3.516} \\
Chile & religion & {2.989} & {2.942} & {2.81} & {2.734} & {3.038} \\
\midrule
China & dates-of-significance & {3.333} & {3.273} & {3.253} & {3.082} & {3.32} \\
China & etiquette & {3.355} & {3.297} & {3.223} & {3.192} & {3.326} \\
China & family & {3.433} & {3.311} & {3.458} & {3.184} & {3.392} \\
China & greetings & {3.103} & {3.065} & {3.039} & {2.989} & {3.117} \\
China & religion & {3.296} & {3.32} & {2.976} & {2.982} & {3.344} \\
\midrule
Germany & dates-of-significance & {3.261} & {3.244} & {3.15} & {3.094} & {3.267} \\
Germany & etiquette & {3.557} & {3.516} & {3.57} & {3.457} & {3.524} \\
Germany & family & {3.576} & {3.493} & {3.534} & {3.256} & {3.522} \\
Germany & greetings & {3.617} & {3.539} & {3.578} & {3.43} & {3.576} \\
Germany & religion & {3.268} & {3.331} & {3.064} & {3.084} & {3.335} \\
\midrule
India & dates-of-significance & {3.497} & {3.418} & {3.45} & {3.229} & {3.479} \\
India & etiquette & {3.534} & {3.5} & {3.484} & {3.364} & {3.538} \\
India & family & {3.279} & {3.252} & {3.234} & {3.08} & {3.315} \\
India & greetings & {3.293} & {3.227} & {3.193} & {3.126} & {3.279} \\
India & religion & {3.432} & {3.431} & {3.358} & {3.233} & {3.466} \\
\midrule
Iran & dates-of-significance & {3.349} & {3.266} & {3.196} & {3.022} & {3.319} \\
Iran & etiquette & {3.334} & {3.309} & {3.251} & {3.189} & {3.318} \\
Iran & family & {3.257} & {3.211} & {3.22} & {3.131} & {3.24} \\
Iran & greetings & {3.27} & {3.232} & {3.26} & {3.184} & {3.233} \\
Iran & religion & {3.21} & {3.192} & {3.114} & {3.067} & {3.239} \\
\midrule
Japan & dates-of-significance & {3.306} & {3.269} & {3.195} & {3.081} & {3.323} \\
Japan & etiquette & {3.608} & {3.534} & {3.581} & {3.445} & {3.552} \\
Japan & family & {3.508} & {3.443} & {3.478} & {3.266} & {3.483} \\
Japan & greetings & {3.314} & {3.319} & {3.193} & {3.137} & {3.335} \\
Japan & religion & {3.118} & {3.097} & {2.946} & {2.886} & {3.119} \\
\midrule
Poland & dates-of-significance & {3.44} & {3.346} & {3.353} & {3.12} & {3.403} \\
Poland & etiquette & {3.643} & {3.591} & {3.608} & {3.46} & {3.581} \\
Poland & family & {3.372} & {3.226} & {3.353} & {3.122} & {3.315} \\
Poland & greetings & {3.268} & {3.243} & {3.205} & {3.144} & {3.295} \\
Poland & religion & {3.036} & {3.009} & {2.795} & {2.827} & {3.03} \\
\midrule
South\_Africa & dates-of-significance & {3.438} & {3.392} & {3.377} & {3.145} & {3.436} \\
South\_Africa & etiquette & {3.439} & {3.393} & {3.429} & {3.333} & {3.417} \\
South\_Africa & greetings & {3.516} & {3.479} & {3.52} & {3.362} & {3.531} \\
South\_Africa & religion & {3.203} & {3.169} & {3.097} & {2.922} & {3.235} \\
\bottomrule
\end{tabular}%
}
\caption{Average VideoScore (LTX-2 \cite{hacohen2024ltxvideorealtimevideolatent}) for geographical constraint removed prompts, grouped by country and category.}
\label{tab:ltx_avg_no_country}
\end{table*}

\begin{table*}[ht]
\centering
\vspace{4pt}
\resizebox{\textwidth}{!}{%
\begin{tabular}{ll CCCCC}
\toprule
\textbf{Country} & \textbf{Category} & \textbf{Visual Quality} & \textbf{Temporal Consistency} & \textbf{Dynamic Degree} & \textbf{Text-to-Video Alignment} & \textbf{Factual Consistency} \\
\midrule
Brazil & dates-of-significance & {2.76} & {2.79} & {2.64} & {2.73} & {2.74} \\
Brazil & etiquette & {2.74} & {2.82} & {2.54} & {2.78} & {2.69} \\
Brazil & family & {2.79} & {2.86} & {2.6} & {2.81} & {2.79} \\
Brazil & greetings & {2.73} & {2.82} & {2.54} & {2.7} & {2.73} \\
Brazil & religion & {2.76} & {2.84} & {2.56} & {2.69} & {2.79} \\
\midrule
Canada & dates-of-significance & {2.8} & {2.82} & {2.61} & {2.7} & {2.77} \\
Canada & etiquette & {2.75} & {2.76} & {2.6} & {2.73} & {2.67} \\
Canada & family & {2.74} & {2.79} & {2.61} & {2.83} & {2.71} \\
Canada & greetings & {2.73} & {2.83} & {2.55} & {2.76} & {2.7} \\
Canada & religion & {2.75} & {2.81} & {2.51} & {2.66} & {2.76} \\
\midrule
Chile & dates-of-significance & {2.8} & {2.84} & {2.62} & {2.74} & {2.77} \\
Chile & etiquette & {2.75} & {2.82} & {2.52} & {2.73} & {2.71} \\
Chile & family & {2.76} & {2.8} & {2.61} & {2.78} & {2.73} \\
Chile & greetings & {2.75} & {2.85} & {2.51} & {2.75} & {2.76} \\
Chile & religion & {2.75} & {2.83} & {2.53} & {2.65} & {2.78} \\
\midrule
China & dates-of-significance & {2.76} & {2.8} & {2.53} & {2.71} & {2.75} \\
China & etiquette & {2.74} & {2.77} & {2.5} & {2.71} & {2.66} \\
China & family & {2.75} & {2.83} & {2.47} & {2.7} & {2.71} \\
China & greetings & {2.74} & {2.83} & {2.46} & {2.72} & {2.78} \\
China & religion & {2.78} & {2.91} & {2.13} & {2.72} & {2.83} \\
\midrule
Germany & dates-of-significance & {2.73} & {2.79} & {2.55} & {2.66} & {2.72} \\
Germany & etiquette & {2.73} & {2.78} & {2.51} & {2.73} & {2.67} \\
Germany & family & {2.75} & {2.81} & {2.57} & {2.78} & {2.71} \\
Germany & greetings & {2.76} & {2.87} & {2.47} & {2.78} & {2.76} \\
Germany & religion & {2.76} & {2.88} & {2.34} & {2.73} & {2.82} \\
\midrule
India & dates-of-significance & {2.79} & {2.86} & {2.43} & {2.72} & {2.79} \\
India & etiquette & {2.75} & {2.8} & {2.43} & {2.71} & {2.71} \\
India & family & {2.77} & {2.85} & {2.45} & {2.76} & {2.76} \\
India & greetings & {2.78} & {2.87} & {2.49} & {2.74} & {2.81} \\
India & religion & {2.78} & {2.83} & {2.47} & {2.74} & {2.8} \\
\midrule
Iran & dates-of-significance & {2.74} & {2.84} & {2.46} & {2.63} & {2.74} \\
Iran & etiquette & {2.76} & {2.84} & {2.45} & {2.71} & {2.7} \\
Iran & family & {2.76} & {2.85} & {2.54} & {2.79} & {2.73} \\
Iran & greetings & {2.77} & {2.88} & {2.44} & {2.73} & {2.78} \\
Iran & religion & {2.74} & {2.87} & {2.39} & {2.78} & {2.75} \\
\midrule
Japan & dates-of-significance & {2.71} & {2.78} & {2.37} & {2.64} & {2.69} \\
Japan & etiquette & {2.74} & {2.8} & {2.39} & {2.71} & {2.67} \\
Japan & family & {2.75} & {2.83} & {2.52} & {2.79} & {2.74} \\
Japan & greetings & {2.71} & {2.76} & {2.42} & {2.67} & {2.64} \\
Japan & religion & {2.73} & {2.82} & {2.28} & {2.63} & {2.69} \\
\midrule
Poland & dates-of-significance & {2.72} & {2.78} & {2.49} & {2.64} & {2.68} \\
Poland & etiquette & {2.72} & {2.8} & {2.46} & {2.66} & {2.64} \\
Poland & family & {2.71} & {2.77} & {2.53} & {2.76} & {2.68} \\
Poland & greetings & {2.76} & {2.89} & {2.46} & {2.72} & {2.77} \\
Poland & religion & {2.77} & {2.88} & {2.44} & {2.68} & {2.79} \\
\midrule
South Africa & dates-of-significance & {2.77} & {2.83} & {2.56} & {2.7} & {2.77} \\
South Africa & etiquette & {2.75} & {2.84} & {2.51} & {2.77} & {2.72} \\
South Africa & greetings & {2.76} & {2.87} & {2.48} & {2.71} & {2.76} \\
South Africa & religion & {2.75} & {2.86} & {2.44} & {2.66} & {2.8} \\
\bottomrule
\end{tabular}
}
\caption{Average Videoscore (Wan 2.2 \cite{wan2025wanopenadvancedlargescale}) for base prompt grouped by Country and Category.}
\label{tab:wan_avg_base_scores}
\end{table*}

\begin{table*}[ht]
\centering
\vspace{4pt}
\resizebox{\textwidth}{!}{%
\begin{tabular}{ll CCCCC}
\toprule
\textbf{Country} & \textbf{Category} & \textbf{Visual Quality} & \textbf{Temporal Consistency} & \textbf{Dynamic Degree} & \textbf{Text-to-Video Alignment} & \textbf{Factual Consistency} \\
\midrule
Brazil & dates-of-significance & {2.90} & {3.06} & {2.70} & {2.84} & {3.01} \\
Brazil & etiquette & {2.66} & {2.62} & {2.70} & {2.78} & {2.53} \\
Brazil & family & {2.84} & {2.84} & {2.73} & {2.86} & {2.80} \\
Brazil & greetings & {2.82} & {2.87} & {2.76} & {2.88} & {2.77} \\
Brazil & religion & {2.72} & {2.70} & {2.68} & {2.80} & {2.67} \\
\midrule
Canada & dates-of-significance & {3.30} & {3.23} & {3.31} & {3.16} & {3.27} \\
Canada & etiquette & {2.80} & {2.79} & {2.82} & {2.80} & {2.80} \\
Canada & family & {2.80} & {2.91} & {2.73} & {2.86} & {2.83} \\
Canada & greetings & {2.68} & {2.80} & {2.64} & {2.77} & {2.64} \\
Canada & religion & {3.66} & {3.60} & {3.52} & {3.50} & {3.62} \\
\midrule
Chile & dates-of-significance & {3.02} & {2.96} & {2.95} & {2.92} & {2.91} \\
Chile & etiquette & {2.69} & {2.75} & {2.58} & {2.78} & {2.64} \\
Chile & family & {2.58} & {2.60} & {2.67} & {2.66} & {2.52} \\
Chile & greetings & {2.63} & {2.70} & {2.61} & {2.69} & {2.65} \\
Chile & religion & {2.83} & {2.92} & {2.57} & {2.83} & {2.86} \\
\midrule
China & dates-of-significance & {2.76} & {2.68} & {2.49} & {2.48} & {2.52} \\
China & etiquette & {2.76} & {2.84} & {2.58} & {2.83} & {2.66} \\
China & family & {2.75} & {2.66} & {2.66} & {2.59} & {2.56} \\
China & greetings & {2.66} & {2.74} & {2.73} & {2.79} & {2.73} \\
China & religion & {2.88} & {2.84} & {2.47} & {2.84} & {2.78} \\
\midrule
Germany & dates-of-significance & {2.66} & {2.65} & {2.48} & {2.64} & {2.62} \\
Germany & etiquette & {2.82} & {2.69} & {2.43} & {2.66} & {2.70} \\
Germany & family & {2.77} & {2.78} & {2.66} & {2.84} & {2.66} \\
Germany & greetings & {2.66} & {2.70} & {2.53} & {2.71} & {2.60} \\
Germany & religion & {2.86} & {2.90} & {2.61} & {2.85} & {2.88} \\
\midrule
India & dates-of-significance & {2.91} & {2.94} & {2.70} & {2.78} & {2.88} \\
India & etiquette & {2.71} & {2.75} & {2.50} & {2.57} & {2.51} \\
India & family & {2.76} & {2.82} & {2.39} & {2.60} & {2.62} \\
India & greetings & {2.75} & {2.84} & {2.48} & {2.77} & {2.75} \\
India & religion & {2.83} & {2.77} & {2.49} & {2.84} & {2.77} \\
\midrule
Iran & dates-of-significance & {2.80} & {2.80} & {2.61} & {2.73} & {2.73} \\
Iran & etiquette & {2.78} & {2.77} & {2.65} & {2.78} & {2.68} \\
Iran & family & {2.82} & {2.91} & {2.49} & {2.85} & {2.75} \\
Iran & greetings & {2.84} & {2.95} & {2.43} & {2.75} & {2.76} \\
Iran & religion & {2.80} & {2.83} & {2.47} & {2.88} & {2.77} \\
\midrule
Japan & dates-of-significance & {2.73} & {2.84} & {2.21} & {2.77} & {2.64} \\
Japan & etiquette & {2.91} & {2.83} & {2.83} & {2.78} & {2.75} \\
Japan & family & {2.73} & {2.84} & {2.31} & {2.73} & {2.66} \\
Japan & greetings & {2.68} & {2.62} & {2.62} & {2.62} & {2.61} \\
Japan & religion & {2.77} & {2.98} & {2.27} & {2.72} & {2.84} \\
\midrule
Poland & dates-of-significance & {2.74} & {2.76} & {2.63} & {2.80} & {2.72} \\
Poland & etiquette & {2.90} & {2.91} & {2.88} & {2.92} & {2.89} \\
Poland & family & {2.80} & {2.80} & {2.60} & {2.77} & {2.69} \\
Poland & greetings & {2.85} & {2.82} & {2.91} & {2.90} & {2.82} \\
Poland & religion & {3.19} & {3.22} & {2.93} & {3.07} & {3.16} \\
\midrule
South Africa & dates-of-significance & {3.20} & {3.20} & {3.24} & {3.13} & {3.23} \\
South Africa & etiquette & {2.63} & {2.70} & {2.63} & {2.63} & {2.56} \\
South Africa & greetings & {2.78} & {2.94} & {2.52} & {2.80} & {2.82} \\
South Africa & religion & {3.23} & {3.14} & {3.16} & {3.11} & {3.23} \\
\bottomrule
\end{tabular}
}
\caption{Average Videoscore (Veo 3.1 Fast \cite{GoogleDeepMind2025Veo3}) for base prompt grouped by country and category.}
\label{tab:veo_avg_base_scores}
\end{table*}

\begin{table*}[ht]
\centering
\vspace{4pt}
\resizebox{\textwidth}{!}{%
\begin{tabular}{ll CCCCC}
\toprule
\textbf{Country} & \textbf{Category} & \textbf{Visual Quality} & \textbf{Temporal Consistency} & \textbf{Dynamic Degree} & \textbf{Text-to-Video Alignment} & \textbf{Factual Consistency} \\
\midrule
Brazil & dates-of-significance & {2.60} & {2.67} & {2.58} & {2.62} & {2.66} \\
Brazil & etiquette & {2.64} & {2.75} & {2.66} & {2.84} & {2.66} \\
Brazil & family & {2.72} & {2.73} & {2.65} & {2.89} & {2.62} \\
Brazil & greetings & {2.73} & {2.91} & {2.48} & {2.80} & {2.75} \\
Brazil & religion & {2.76} & {2.88} & {2.52} & {2.82} & {2.80} \\
\midrule
Canada & dates-of-significance & {3.30} & {3.30} & {3.37} & {3.22} & {3.34} \\
Canada & etiquette & {2.75} & {2.69} & {2.72} & {2.73} & {2.72} \\
Canada & family & {3.03} & {3.08} & {2.91} & {3.03} & {3.00} \\
Canada & greetings & {2.82} & {2.86} & {2.79} & {2.93} & {2.80} \\
Canada & religion & {3.16} & {3.12} & {2.97} & {3.10} & {3.19} \\
\midrule
Chile & dates-of-significance & {3.67} & {3.59} & {3.66} & {3.44} & {3.64} \\
Chile & etiquette & {2.70} & {2.73} & {2.55} & {2.73} & {2.59} \\
Chile & family & {2.71} & {2.62} & {2.70} & {2.73} & {2.62} \\
Chile & greetings & {2.84} & {2.89} & {2.70} & {2.84} & {2.88} \\
Chile & religion & {2.80} & {2.87} & {2.54} & {2.73} & {2.79} \\
\midrule
China & dates-of-significance & {2.80} & {2.79} & {2.48} & {2.65} & {2.59} \\
China & etiquette & {2.73} & {2.84} & {2.36} & {2.70} & {2.59} \\
China & family & {2.80} & {2.88} & {2.61} & {2.80} & {2.66} \\
China & greetings & {2.73} & {2.79} & {2.67} & {2.80} & {2.75} \\
China & religion & {2.74} & {2.94} & {2.12} & {2.77} & {2.80} \\
\midrule
Germany & dates-of-significance & {2.69} & {2.69} & {2.46} & {2.59} & {2.57} \\
Germany & etiquette & {2.71} & {2.71} & {2.57} & {2.66} & {2.63} \\
Germany & family & {2.80} & {2.84} & {2.52} & {2.84} & {2.63} \\
Germany & greetings & {2.65} & {2.65} & {2.59} & {2.77} & {2.55} \\
Germany & religion & {2.80} & {2.91} & {2.56} & {2.85} & {2.82} \\
\midrule
India & dates-of-significance & {2.73} & {2.83} & {2.54} & {2.66} & {2.72} \\
India & etiquette & {2.60} & {2.63} & {2.48} & {2.64} & {2.48} \\
India & family & {2.88} & {2.84} & {2.62} & {2.88} & {2.78} \\
India & greetings & {2.89} & {2.98} & {2.41} & {2.81} & {2.81} \\
India & religion & {2.74} & {2.82} & {2.36} & {2.84} & {2.76} \\
\midrule
Iran & dates-of-significance & {3.24} & {3.27} & {3.07} & {3.09} & {3.22} \\
Iran & etiquette & {3.28} & {3.30} & {3.12} & {3.12} & {3.25} \\
Iran & family & {2.70} & {2.77} & {2.52} & {2.78} & {2.59} \\
Iran & greetings & {2.86} & {2.94} & {2.48} & {2.75} & {2.76} \\
Iran & religion & {2.72} & {2.67} & {2.66} & {2.77} & {2.67} \\
\midrule
Japan & dates-of-significance & {2.75} & {2.88} & {2.45} & {2.77} & {2.68} \\
Japan & etiquette & {2.69} & {2.51} & {2.73} & {2.64} & {2.50} \\
Japan & family & {2.82} & {2.91} & {2.45} & {2.88} & {2.78} \\
Japan & greetings & {2.77} & {2.94} & {2.00} & {2.50} & {2.69} \\
Japan & religion & {2.80} & {2.96} & {2.26} & {2.80} & {2.84} \\
\midrule
Poland & dates-of-significance & {2.80} & {2.90} & {2.46} & {2.83} & {2.79} \\
Poland & etiquette & {2.73} & {2.73} & {2.64} & {2.66} & {2.62} \\
Poland & family & {2.68} & {2.66} & {2.62} & {2.77} & {2.63} \\
Poland & greetings & {2.90} & {2.94} & {2.76} & {2.90} & {2.91} \\
Poland & religion & {2.78} & {2.82} & {2.64} & {2.77} & {2.77} \\
\midrule
South Africa & dates-of-significance & {2.86} & {2.91} & {2.75} & {2.85} & {2.90} \\
South Africa & etiquette & {2.67} & {2.72} & {2.56} & {2.56} & {2.66} \\
South Africa & greetings & {2.72} & {2.69} & {2.73} & {2.80} & {2.70} \\
South Africa & religion & {2.67} & {2.81} & {2.31} & {2.72} & {2.70} \\
\bottomrule
\end{tabular}
}
\caption{Average VideoScore evaluation of Veo 3.1 Fast \cite{GoogleDeepMind2025Veo3} for extended prompt generations across countries and categories.}
\label{tab:veo_avg_extended_scores}
\end{table*}

\begin{table*}[ht]
\centering
\vspace{4pt}
\resizebox{\textwidth}{!}{%
\begin{tabular}{ll CCCCC}
\toprule
\textbf{Country} & \textbf{Category} & \textbf{Visual Quality} & \textbf{Temporal Consistency} & \textbf{Dynamic Degree} & \textbf{Text-to-Video Alignment} & \textbf{Factual Consistency} \\
\midrule
Brazil & dates-of-significance & {2.69} & {2.86} & {2.33} & {2.72} & {2.75} \\
Brazil & etiquette & {2.56} & {2.59} & {2.73} & {2.81} & {2.53} \\
Brazil & family & {2.68} & {2.70} & {2.61} & {2.84} & {2.55} \\
Brazil & greetings & {2.74} & {2.86} & {2.59} & {2.79} & {2.80} \\
Brazil & religion & {2.69} & {2.70} & {2.55} & {2.75} & {2.67} \\
\midrule
Canada & dates-of-significance & {2.86} & {2.80} & {2.72} & {2.77} & {2.79} \\
Canada & etiquette & {2.87} & {2.90} & {2.80} & {2.91} & {2.88} \\
Canada & family & {2.69} & {2.79} & {2.73} & {2.89} & {2.65} \\
Canada & greetings & {2.72} & {2.77} & {2.62} & {2.66} & {2.62} \\
Canada & religion & {3.02} & {2.92} & {3.03} & {3.05} & {3.01} \\
\midrule
Chile & dates-of-significance & {3.63} & {3.49} & {3.73} & {3.48} & {3.48} \\
Chile & etiquette & {2.75} & {2.73} & {2.70} & {2.82} & {2.66} \\
Chile & family & {2.59} & {2.55} & {2.72} & {2.78} & {2.42} \\
Chile & greetings & {2.77} & {2.92} & {2.45} & {2.77} & {2.81} \\
Chile & religion & {3.00} & {2.97} & {2.88} & {2.89} & {2.95} \\
\midrule
China & dates-of-significance & {2.80} & {2.69} & {2.55} & {2.80} & {2.53} \\
China & etiquette & {2.70} & {2.70} & {2.55} & {2.64} & {2.54} \\
China & family & {3.24} & {3.23} & {3.09} & {3.09} & {3.12} \\
China & greetings & {3.12} & {3.12} & {3.20} & {3.15} & {3.14} \\
China & religion & {2.85} & {2.91} & {2.41} & {2.86} & {2.84} \\
\midrule
Germany & dates-of-significance & {2.71} & {2.71} & {2.46} & {2.62} & {2.62} \\
Germany & etiquette & {2.78} & {2.75} & {2.23} & {2.73} & {2.71} \\
Germany & family & {2.86} & {2.90} & {2.64} & {2.96} & {2.72} \\
Germany & greetings & {2.70} & {2.89} & {2.41} & {2.62} & {2.75} \\
Germany & religion & {2.79} & {2.88} & {2.42} & {2.74} & {2.84} \\
\midrule
India & dates-of-significance & {2.67} & {2.84} & {2.30} & {2.48} & {2.63} \\
India & etiquette & {3.14} & {3.07} & {2.99} & {2.91} & {2.98} \\
India & family & {2.64} & {2.61} & {2.62} & {2.70} & {2.53} \\
India & greetings & {2.89} & {3.02} & {2.38} & {2.80} & {2.84} \\
India & religion & {2.93} & {2.94} & {2.59} & {2.94} & {2.96} \\
\midrule
Iran & dates-of-significance & {2.75} & {2.80} & {2.57} & {2.75} & {2.73} \\
Iran & etiquette & {2.72} & {2.70} & {2.60} & {2.58} & {2.62} \\
Iran & family & {2.77} & {2.90} & {2.50} & {2.87} & {2.72} \\
Iran & greetings & {2.84} & {2.96} & {2.31} & {2.66} & {2.73} \\
Iran & religion & {2.75} & {2.62} & {2.75} & {2.83} & {2.62} \\
\midrule
Japan & dates-of-significance & {2.79} & {2.88} & {2.25} & {2.73} & {2.61} \\
Japan & etiquette & {2.71} & {2.70} & {2.67} & {2.73} & {2.59} \\
Japan & family & {2.76} & {2.82} & {2.54} & {2.81} & {2.73} \\
Japan & greetings & {2.89} & {2.69} & {3.06} & {2.89} & {2.84} \\
Japan & religion & {2.80} & {2.97} & {2.20} & {2.78} & {2.85} \\
\midrule
Poland & dates-of-significance & {2.76} & {2.84} & {2.50} & {2.86} & {2.70} \\
Poland & etiquette & {3.27} & {3.27} & {3.19} & {3.08} & {3.06} \\
Poland & family & {2.71} & {2.73} & {2.52} & {2.66} & {2.63} \\
Poland & greetings & {2.62} & {2.67} & {2.61} & {2.69} & {2.61} \\
Poland & religion & {2.67} & {2.78} & {2.46} & {2.56} & {2.62} \\
\midrule
South Africa & dates-of-significance & {3.18} & {3.16} & {3.20} & {3.12} & {3.21} \\
South Africa & etiquette & {2.66} & {2.76} & {2.65} & {2.71} & {2.71} \\
South Africa & greetings & {2.62} & {2.56} & {2.78} & {2.78} & {2.55} \\
South Africa & religion & {2.70} & {2.70} & {2.55} & {2.77} & {2.67} \\
\bottomrule
\end{tabular}
}
\caption{Average VideoScore evaluation of Veo 3.1 Fast \cite{GoogleDeepMind2025Veo3} for geographical Constrainst Removed Prompting.}
\label{tab:veo_avg_no_country_scores}
\end{table*}

\subsection{CultureScore across Models } \label{app:culturescore_aso}

In the following, we categorize CultureScore across countries and categories for LTX-2\cite{hacohen2024ltxvideorealtimevideolatent}, Wan 2.2 \cite{wan2025wanopenadvancedlargescale}, and Veo 3.1 Fast \cite{GoogleDeepMind2025Veo3}. Wan 2.2 \cite{wan2025wanopenadvancedlargescale} consistently outperformed for Identity and Context, and Veo 3.1 Fast \cite{GoogleDeepMind2025Veo3} for behavior.

Table  \ref{tab:accuracy_variation_ltx}, \ref{tab:accuracy_variation_wan}  for overall comparison across prompting strategies.
Table  \ref{tab:aso_base_CultureScore_ltx}, \ref{tab:aso_extended_CultureScore_ltx} for lTX-2 \cite{hacohen2024ltxvideorealtimevideolatent},
Table \ref{tab:CultureScore_wan_base}, \ref{tab:wan_extended_by_category}, \ref{tab:accuracy_veo_base}, and \ref{tab:CultureScore_veo_extended} for Wan 2.2 \cite{wan2025wanopenadvancedlargescale} and Veo 3.1 Fast \cite{GoogleDeepMind2025Veo3}
Further provide all scores across countries and categories.

\subsection{Comparison of UnifiedReward and CLIPScore against Human Judgment}
\label{app:comparison}

\textbf{CLIPScore} \cite{hessel2021clipscore} measures generic image--text alignment and requires a still image, so each generated video was reduced to a single frame extracted at a fixed 2-second offset with \texttt{ffmpeg} (\texttt{-ss 00:00:02 -frames:v 1}, JPEG quality \texttt{-q:v 2}); the offset was chosen to skip the blank or under-formed opening frames common in generated video while remaining inside the duration of every clip. Frames were scored with \texttt{openai/clip-vit-large-patch14}. For each (frame, prompt) pair, we take the CLIP image and text embeddings, apply $L_2$ normalization, and compute
their cosine similarity. The final score follows the standard formulation,
$\mathrm{CLIPScore} = w \cdot \max(0, \cos(\mathbf{v}, \mathbf{t}))$, where the rectification discards negative similarities and $w = 2.5$ rescales the
typically narrow cosine range into a more readable interval.

\noindent
\textbf{UnifiedReward} \cite{wang2025unified} is a VLM-as-judge reward model. We use UnifiedReward-qwen-7b, which is built on a Qwen2.5-VL backbone \cite{bai2025qwen25vltechnicalreport}, scoring the same frame and the same prompt text used for CLIPScore. The model receives the frame together with a rubric asking it to evaluate two criteria: caption alignment and overall image quality. Scores are reported on the native 1--5 scale and additionally min-max normalized to $[0, 1]$ as $(s - 1)/4$.
We report Spearman's $\rho$ against human judgments collected from 45 culturally knowledgeable native annotators across 9 countries.

\begin{table*}[ht]
\centering
\begin{tabular}{llrr}
\toprule
Metric & Dimension & $\rho$ & $p$  \\
\midrule
CultureScore  & Behavior & $\mathbf{+0.25}$ & $\mathbf{0.045}$  \\
CultureScore  & Identity & $+0.21$ & $0.08$   \\
CultureScore  & Context  & $+0.04$ & $0.71$   \\
UnifiedReward & overall  & $+0.22$ & $0.08$  \\
CLIPScore     & overall  & $-0.22$ & $0.06$   \\
VideoScore    & overall  & $-0.14$ & $0.28$  \\
\bottomrule
\end{tabular}
\caption{Correlation between automatic metrics and human cultural judgement.}
\label{tab:metric-correlations}
\end{table*}

As specified in Table \ref{tab:metric-correlations}, only the Behavior dimension, which requires temporal grounding unavailable to static metrics, reaches significance; no other dimension or baseline does. This
is consistent with temporal information being an important driver of CultureScore's advantage over image-level metrics, although because of limited ratings for human preference, we read this as suggestive rather than conclusive. CLIPScore trends negative ($\rho = -0.22$, $p = 0.06$), consistent with prior findings that
generic image--text alignment does not track cultural faithfulness. UnifiedReward, in contrast, shows a positive trend ($\rho = +0.22$, $p = 0.08$) comparable to CultureScore's Identity dimension. Neither reaches significance, but the contrast suggests that reward-tuned image metrics may partially capture cultural cues that CLIPScore misses, while still lacking CultureScore's ability to localize failures across Identity, Behavior, and Context. Notably, VideoScore
does not correlate with human cultural judgment ($\rho = -0.14$, $p = 0.28$) despite being video-specific.

\begin{table*}[ht]
\centering
\vspace{4pt}
\resizebox{\textwidth}{!}{%
\begin{tabular}{l rr r rr r rr r rr r}
\toprule
& \multicolumn{3}{c}{\textbf{Behavior}} & \multicolumn{3}{c}{\textbf{Context}} & \multicolumn{3}{c}{\textbf{Identity}} & \multicolumn{3}{c}{\textbf{Overall}} \\
\cmidrule(lr){2-4} \cmidrule(lr){5-7} \cmidrule(lr){8-10} \cmidrule(lr){11-13}
\textbf{Country} & \textbf{Base} & \textbf{Ext} & \textbf{Gain} & \textbf{Base} & \textbf{Ext} & \textbf{Gain} & \textbf{Base} & \textbf{Ext} & \textbf{Gain} & \textbf{Base} & \textbf{Ext} & \textbf{Gain} \\
\midrule
Brazil & 33.3 & 46.8 & \textbf{+13.5} & 44.3 & 63.8 & \textbf{+19.5} & 21.4 & 39.5 & \textbf{+18.1} & 32.7 & 49.7 & \textbf{+17.0} \\
Canada & 29.2 & 54.1 & \textbf{+24.9} & 27.3 & 54.2 & \textbf{+26.9} & 19.0 & 40.5 & \textbf{+21.5} & 24.9 & 49.3 & \textbf{+24.4} \\
Chile & 27.5 & 44.5 & \textbf{+17.0} & 26.4 & 50.6 & \textbf{+24.2} & 20.9 & 35.3 & \textbf{+14.4} & 24.8 & 43.2 & \textbf{+18.4} \\
China & 29.5 & 46.8 & \textbf{+17.3} & 42.1 & 71.2 & \textbf{+29.1} & 31.1 & 55.3 & \textbf{+24.2} & 34.0 & 57.4 & \textbf{+23.4} \\
Germany & 32.4 & 53.8 & \textbf{+21.4} & 39.8 & 64.8 & \textbf{+25.0} & 33.0 & 47.8 & \textbf{+14.9} & 35.0 & 55.3 & \textbf{+20.3} \\
India & 29.3 & 46.2 & \textbf{+16.9} & 41.4 & 66.2 & \textbf{+24.8} & 34.2 & 55.8 & \textbf{+21.6} & 34.6 & 55.6 & \textbf{+20.9} \\
Iran & 25.8 & 38.5 & \textbf{+12.7} & 36.0 & 61.2 & \textbf{+25.2} & 27.8 & 46.4 & \textbf{+18.5} & 29.7 & 48.4 & \textbf{+18.7} \\
Japan & 39.8 & 55.1 & \textbf{+15.3} & 41.0 & 74.5 & \textbf{+33.5} & 27.6 & 46.9 & \textbf{+19.3} & 36.1 & 58.6 & \textbf{+22.5} \\
Poland & 31.8 & 48.6 & \textbf{+16.9} & 33.8 & 59.8 & \textbf{+25.9} & 27.9 & 43.9 & \textbf{+16.0} & 31.1 & 50.5 & \textbf{+19.4} \\
South Africa & 21.9 & 34.6 & \textbf{+12.6} & 30.3 & 49.4 & \textbf{+19.2} & 23.2 & 34.9 & \textbf{+11.6} & 25.1 & 39.4 & \textbf{+14.4} \\

\bottomrule
\end{tabular}%
}
\caption{Overall CultureScore (\%) across countries and prompting strategies for LTX-2 \cite{hacohen2024ltxvideorealtimevideolatent}. Ext implies extended prompt and Gain implies (\%) gain from Base to Extended Prompt.}
\label{tab:accuracy_variation_ltx}
\end{table*}

\begin{table*}[ht]
\centering
\vspace{4pt}
\resizebox{\textwidth}{!}{%
\begin{tabular}{l rr r rr r rr r rr r}
\toprule
& \multicolumn{3}{c}{\textbf{Behavior}} & \multicolumn{3}{c}{\textbf{Context}} & \multicolumn{3}{c}{\textbf{Identity}} & \multicolumn{3}{c}{\textbf{Overall}} \\
\cmidrule(lr){2-4} \cmidrule(lr){5-7} \cmidrule(lr){8-10} \cmidrule(lr){11-13}
\textbf{Country} & \textbf{Base} & \textbf{Ext} & \textbf{Gain} & \textbf{Base} & \textbf{Ext} & \textbf{Gain} & \textbf{Base} & \textbf{Ext} & \textbf{Gain} & \textbf{Base} & \textbf{Ext} & \textbf{Gain} \\
\midrule
Brazil & 32.7 & 46.8 & \textbf{+14.1} & 52.8 & 74.1 & \textbf{+21.3} & 29.6 & 45.7 & \textbf{+16.1} & 38.1 & 55.2 & \textbf{+17.1} \\
Canada & 38.6 & 58.8 & \textbf{+20.2} & 43.8 & 62.3 & \textbf{+18.5} & 34.3 & 51.8 & \textbf{+17.5} & 38.9 & 57.5 & \textbf{+18.6} \\
Chile & 27.9 & 44.9 & \textbf{+17.0} & 40.4 & 63.8 & \textbf{+23.4} & 26.7 & 41.1 & \textbf{+14.4} & 31.5 & 49.6 & \textbf{+18.1} \\
China & 41.6 & 54.3 & \textbf{+12.7} & 57.0 & 77.8 & \textbf{+20.9} & 48.3 & 67.7 & \textbf{+19.3} & 48.7 & 66.3 & \textbf{+17.5} \\
Germany & 35.1 & 51.1 & \textbf{+16.0} & 51.6 & 73.0 & \textbf{+21.5} & 36.6 & 55.4 & \textbf{+18.8} & 40.9 & 59.7 & \textbf{+18.8} \\
India & 32.1 & 49.6 & \textbf{+17.5} & 46.7 & 69.2 & \textbf{+22.5} & 41.2 & 62.6 & \textbf{+21.3} & 39.6 & 60.0 & \textbf{+20.3} \\
Iran & 27.1 & 42.5 & \textbf{+15.4} & 36.7 & 65.8 & \textbf{+29.1} & 28.8 & 45.0 & \textbf{+16.2} & 30.7 & 50.7 & \textbf{+20.0} \\
Japan & 46.3 & 62.5 & \textbf{+16.2} & 59.3 & 76.4 & \textbf{+17.1} & 36.5 & 57.0 & \textbf{+20.5} & 47.2 & 65.1 & \textbf{+17.9} \\
Poland & 30.7 & 40.2 & \textbf{+9.5} & 42.9 & 64.3 & \textbf{+21.4} & 28.2 & 44.3 & \textbf{+16.0} & 33.7 & 49.1 & \textbf{+15.4} \\
South Africa & 23.4 & 45.0 & \textbf{+21.6} & 43.3 & 63.2 & \textbf{+19.9} & 21.1 & 38.4 & \textbf{+17.3} & 29.0 & 48.5 & \textbf{+19.5} \\

\bottomrule
\end{tabular}%
}
\caption{Overall CultureScore (\%) across countries and prompting strategies for Wan 2.2 \cite{wan2025wanopenadvancedlargescale}. Ext implies extended prompt and Gain implies (\%) gain from Base to Extended Prompt.}
\label{tab:accuracy_variation_wan}
\end{table*}



\begin{table*}[ht]
\centering
\begin{tabular}{ll CCC}
\toprule
 & & \multicolumn{3}{c}{\textbf{CultureScore (\%)}} \\
\cmidrule(lr){3-5}
\textbf{Country} & \textbf{Category} & \textbf{Identity} & \textbf{Behavior} & \textbf{Context} \\
\midrule
Brazil & dates-of-significance & 24.5 & 35.6 & 38.6 \\
Brazil & etiquette & 21.2 & 27.6 & 51.7 \\
Brazil & family & 18.4 & 36.2 & 51.1 \\
Brazil & greetings & 19.5 & 44.7 & 53.7 \\
Brazil & religion & 20.4 & 23.6 & 32.0 \\
\midrule
Canada & dates-of-significance & 18.8 & 25.0 & 16.9 \\
Canada & etiquette & 14.5 & 33.3 & 27.3 \\
Canada & family & 16.7 & 28.9 & 40.0 \\
Canada & greetings & 36.7 & 50.0 & 33.3 \\
Canada & religion & 15.4 & 17.6 & 30.6 \\
\midrule
Chile & dates-of-significance & 29.6 & 28.7 & 19.3 \\
Chile & etiquette & 19.7 & 28.0 & 35.0 \\
Chile & family & 9.3 & 12.2 & 23.1 \\
Chile & greetings & 30.3 & 50.0 & 43.8 \\
Chile & religion & 8.3 & 20.6 & 18.2 \\
\midrule
China & dates-of-significance & 24.5 & 23.9 & 43.4 \\
China & etiquette & 44.3 & 35.4 & 42.9 \\
China & family & 30.8 & 30.9 & 39.6 \\
China & greetings & 31.6 & 39.4 & 51.4 \\
China & religion & 20.9 & 21.7 & 32.6 \\
\midrule
Germany & dates-of-significance & 24.0 & 23.8 & 30.6 \\
Germany & etiquette & 40.0 & 45.0 & 53.4 \\
Germany & family & 33.3 & 32.3 & 43.2 \\
Germany & greetings & 51.4 & 44.7 & 53.1 \\
Germany & religion & 27.3 & 21.9 & 25.8 \\
\midrule
India & dates-of-significance & 29.7 & 30.9 & 36.3 \\
India & etiquette & 34.2 & 40.5 & 50.0 \\
India & family & 34.0 & 13.3 & 38.5 \\
India & greetings & 55.0 & 36.8 & 54.5 \\
India & religion & 29.6 & 19.4 & 35.8 \\
\midrule
Iran & dates-of-significance & 26.2 & 24.5 & 32.6 \\
Iran & etiquette & 26.0 & 28.6 & 38.0 \\
Iran & family & 20.0 & 18.8 & 36.4 \\
Iran & greetings & 36.8 & 35.1 & 41.7 \\
Iran & religion & 37.1 & 22.9 & 34.4 \\
\midrule
Japan & dates-of-significance & 19.5 & 30.9 & 35.6 \\
Japan & etiquette & 41.7 & 44.9 & 46.3 \\
Japan & family & 24.4 & 36.4 & 36.4 \\
Japan & greetings & 26.7 & 66.7 & 48.8 \\
Japan & religion & 26.0 & 25.5 & 40.8 \\
\midrule
Poland & dates-of-significance & 28.4 & 31.5 & 38.1 \\
Poland & etiquette & 37.5 & 36.0 & 43.8 \\
Poland & family & 14.3 & 19.6 & 33.3 \\
Poland & greetings & 35.7 & 58.8 & 25.0 \\
Poland & religion & 24.4 & 22.9 & 18.6 \\
\midrule
South Africa & dates-of-significance & 17.3 & 19.8 & 29.0 \\
South Africa & etiquette & 24.7 & 25.9 & 30.7 \\
South Africa & greetings & 34.2 & 40.6 & 35.9 \\
South Africa & religion & 25.5 & 8.0 & 27.7 \\
\midrule
\textbf{Overall} & & \textbf{26.8} & \textbf{30.3} & \textbf{36.6} \\
\bottomrule
\end{tabular}
\caption{CultureScore for LTX-2 \cite{hacohen2024ltxvideorealtimevideolatent} base prompt by country and category.}
\label{tab:aso_base_CultureScore_ltx}
\end{table*}

\begin{table*}[ht]
\centering
\begin{tabular}{ll CCC}
\toprule
 & & \multicolumn{3}{c}{\textbf{CultureScore (\%)}} \\
\cmidrule(lr){3-5}
\textbf{Country} & \textbf{Category} & \textbf{Identity} & \textbf{Behavior} & \textbf{Context} \\
\midrule
Brazil & dates-of-significance & 42.6 & 46.7 & 54.5 \\
Brazil & etiquette & 43.9 & 41.4 & 79.3 \\
Brazil & family & 34.7 & 55.3 & 64.4 \\
Brazil & greetings & 34.1 & 57.4 & 63.4 \\
Brazil & religion & 37.0 & 36.4 & 62.0 \\
\midrule
Canada & dates-of-significance & 39.6 & 51.1 & 52.8 \\
Canada & etiquette & 45.5 & 50.0 & 67.3 \\
Canada & family & 31.5 & 51.1 & 42.0 \\
Canada & greetings & 56.7 & 93.3 & 73.3 \\
Canada & religion & 35.9 & 35.3 & 38.9 \\
\midrule
Chile & dates-of-significance & 29.6 & 52.1 & 42.0 \\
Chile & etiquette & 49.3 & 42.0 & 61.7 \\
Chile & family & 14.8 & 26.5 & 44.2 \\
Chile & greetings & 54.5 & 68.4 & 65.6 \\
Chile & religion & 36.1 & 26.5 & 48.5 \\
\midrule
China & dates-of-significance & 50.9 & 36.3 & 68.9 \\
China & etiquette & 65.9 & 56.6 & 79.8 \\
China & family & 55.8 & 45.5 & 62.5 \\
China & greetings & 39.5 & 60.6 & 62.9 \\
China & religion & 58.1 & 43.5 & 76.7 \\
\midrule
Germany & dates-of-significance & 41.3 & 53.5 & 68.4 \\
Germany & etiquette & 55.4 & 65.0 & 72.4 \\
Germany & family & 48.7 & 41.9 & 54.1 \\
Germany & greetings & 51.4 & 60.5 & 59.4 \\
Germany & religion & 48.5 & 37.5 & 58.1 \\
\midrule
India & dates-of-significance & 56.2 & 44.9 & 63.7 \\
India & etiquette & 52.1 & 50.0 & 81.2 \\
India & family & 51.1 & 44.4 & 59.0 \\
India & greetings & 72.5 & 50.0 & 75.8 \\
India & religion & 51.9 & 43.5 & 52.8 \\
\midrule
Iran & dates-of-significance & 36.4 & 40.2 & 57.9 \\
Iran & etiquette & 57.1 & 37.7 & 73.2 \\
Iran & family & 42.2 & 35.4 & 54.5 \\
Iran & greetings & 47.4 & 54.1 & 47.2 \\
Iran & religion & 57.1 & 22.9 & 68.8 \\
\midrule
Japan & dates-of-significance & 43.4 & 50.0 & 60.6 \\
Japan & etiquette & 53.6 & 60.7 & 87.8 \\
Japan & family & 48.9 & 52.3 & 68.2 \\
Japan & greetings & 46.7 & 79.6 & 81.4 \\
Japan & religion & 42.0 & 34.5 & 81.6 \\
\midrule
Poland & dates-of-significance & 50.5 & 51.9 & 59.0 \\
Poland & etiquette & 39.3 & 40.0 & 56.2 \\
Poland & family & 34.7 & 48.2 & 73.8 \\
Poland & greetings & 50.0 & 67.6 & 64.3 \\
Poland & religion & 40.0 & 37.5 & 48.8 \\
\midrule
South Africa & dates-of-significance & 22.7 & 31.1 & 42.0 \\
South Africa & etiquette & 44.7 & 44.4 & 50.7 \\
South Africa & greetings & 47.4 & 37.5 & 59.0 \\
South Africa & religion & 35.3 & 24.0 & 55.3 \\
\midrule
\textbf{Overall} & & \textbf{45.0} & \textbf{47.0} & \textbf{62.1} \\
\bottomrule
\end{tabular}
\caption{CultureScore (\%) by country and category for LTX-2 \cite{hacohen2024ltxvideorealtimevideolatent} with the extended prompt.}
\label{tab:aso_extended_CultureScore_ltx}
\end{table*}

\begin{table*}[ht]
\centering
\begin{tabular}{ll CCC}
\toprule
 & & \multicolumn{3}{c}{\textbf{CultureScore (\%)}} \\
\cmidrule(lr){3-5}
\textbf{Country} & \textbf{Category} & \textbf{Identity} & \textbf{Behavior} & \textbf{Context} \\
\midrule
Brazil & dates-of-significance & 31.9 & 38.9 & 50.0 \\
Brazil & etiquette & 25.8 & 25.9 & 63.8 \\
Brazil & family & 24.5 & 38.3 & 53.3 \\
Brazil & greetings & 39.0 & 40.4 & 53.7 \\
Brazil & religion & 27.8 & 18.2 & 44.0 \\
\midrule
Canada & dates-of-significance & 36.5 & 40.9 & 34.8 \\
Canada & etiquette & 21.8 & 27.8 & 47.3 \\
Canada & family & 40.7 & 46.7 & 58.0 \\
Canada & greetings & 60.0 & 53.3 & 60.0 \\
Canada & religion & 17.9 & 20.6 & 27.8 \\
\midrule
Chile & dates-of-significance & 31.6 & 29.8 & 47.7 \\
Chile & etiquette & 23.9 & 30.0 & 36.7 \\
Chile & family & 11.1 & 14.3 & 28.8 \\
Chile & greetings & 42.4 & 47.4 & 53.1 \\
Chile & religion & 27.8 & 17.6 & 33.3 \\
\midrule
China & dates-of-significance & 47.3 & 42.5 & 56.6 \\
China & etiquette & 52.3 & 48.5 & 64.3 \\
China & family & 32.7 & 14.5 & 33.3 \\
China & greetings & 55.3 & 54.5 & 71.4 \\
China & religion & 55.8 & 47.8 & 58.1 \\
\midrule
Germany & dates-of-significance & 32.7 & 24.8 & 51.0 \\
Germany & etiquette & 50.8 & 51.7 & 53.4 \\
Germany & family & 17.9 & 35.5 & 51.4 \\
Germany & greetings & 45.7 & 39.5 & 62.5 \\
Germany & religion & 33.3 & 31.2 & 38.7 \\
\midrule
India & dates-of-significance & 29.7 & 29.4 & 43.4 \\
India & etiquette & 45.2 & 40.5 & 48.4 \\
India & family & 40.4 & 11.1 & 48.7 \\
India & greetings & 70.0 & 44.7 & 60.6 \\
India & religion & 42.6 & 35.5 & 41.5 \\
\midrule
Iran & dates-of-significance & 24.3 & 25.5 & 35.8 \\
Iran & etiquette & 31.2 & 24.7 & 46.5 \\
Iran & family & 24.4 & 27.1 & 20.5 \\
Iran & greetings & 31.6 & 40.5 & 38.9 \\
Iran & religion & 40.0 & 22.9 & 37.5 \\
\midrule
Japan & dates-of-significance & 25.7 & 37.3 & 46.2 \\
Japan & etiquette & 53.6 & 57.3 & 68.3 \\
Japan & family & 31.1 & 40.9 & 65.9 \\
Japan & greetings & 42.2 & 59.3 & 60.5 \\
Japan & religion & 32.0 & 38.2 & 65.3 \\
\midrule
Poland & dates-of-significance & 32.1 & 33.3 & 44.8 \\
Poland & etiquette & 26.8 & 32.0 & 41.7 \\
Poland & family & 22.4 & 21.4 & 40.5 \\
Poland & greetings & 35.7 & 52.9 & 57.1 \\
Poland & religion & 22.2 & 18.8 & 32.6 \\
\midrule
South Africa & dates-of-significance & 16.4 & 23.6 & 34.0 \\
South Africa & etiquette & 23.5 & 24.7 & 52.0 \\
South Africa & greetings & 28.9 & 25.0 & 46.2 \\
South Africa & religion & 21.6 & 20.0 & 46.8 \\
\midrule
\textbf{Overall} & & \textbf{33.4} & \textbf{33.9} & \textbf{47.8} \\
\bottomrule
\end{tabular}
\caption{CultureScore by country and category for Wan 2.2 \cite{wan2025wanopenadvancedlargescale} with the base prompt.}
\label{tab:CultureScore_wan_base}
\end{table*}

\begin{table*}[ht]
\centering
\begin{tabular}{ll CCC}
\toprule
 & & \multicolumn{3}{c}{\textbf{CultureScore (\%)}} \\
\cmidrule(lr){3-5}
\textbf{Country} & \textbf{Category} & \textbf{Identity} & \textbf{Behavior} & \textbf{Context} \\
\midrule
Brazil & dates-of-significance & 43.6 & 52.2 & 72.7 \\
Brazil & etiquette & 48.5 & 34.5 & 74.1 \\
Brazil & family & 44.9 & 36.2 & 84.4 \\
Brazil & greetings & 39.0 & 66.0 & 80.5 \\
Brazil & religion & 51.9 & 43.6 & 62.0 \\
\midrule
Canada & dates-of-significance & 52.1 & 63.6 & 57.3 \\
Canada & etiquette & 43.6 & 50.0 & 69.1 \\
Canada & family & 55.6 & 57.8 & 60.0 \\
Canada & greetings & 70.0 & 76.7 & 80.0 \\
Canada & religion & 43.6 & 41.2 & 52.8 \\
\midrule
Chile & dates-of-significance & 46.9 & 45.7 & 60.2 \\
Chile & etiquette & 43.7 & 54.0 & 71.7 \\
Chile & family & 13.0 & 34.7 & 51.9 \\
Chile & greetings & 63.6 & 55.3 & 81.2 \\
Chile & religion & 41.7 & 32.4 & 60.6 \\
\midrule
China & dates-of-significance & 68.2 & 47.8 & 80.2 \\
China & etiquette & 71.6 & 62.6 & 86.9 \\
China & family & 50.0 & 40.0 & 64.6 \\
China & greetings & 68.4 & 75.8 & 62.9 \\
China & religion & 79.1 & 54.3 & 81.4 \\
\midrule
Germany & dates-of-significance & 44.2 & 45.5 & 75.5 \\
Germany & etiquette & 73.8 & 68.3 & 74.1 \\
Germany & family & 48.7 & 45.2 & 73.0 \\
Germany & greetings & 62.9 & 55.3 & 75.0 \\
Germany & religion & 54.5 & 37.5 & 61.3 \\
\midrule
India & dates-of-significance & 62.5 & 48.5 & 65.5 \\
India & etiquette & 56.2 & 55.4 & 73.4 \\
India & family & 70.2 & 28.9 & 74.4 \\
India & greetings & 77.5 & 57.9 & 78.8 \\
India & religion & 53.7 & 54.8 & 62.3 \\
\midrule
Iran & dates-of-significance & 36.4 & 44.1 & 68.4 \\
Iran & etiquette & 46.8 & 37.7 & 83.1 \\
Iran & family & 48.9 & 39.6 & 45.5 \\
Iran & greetings & 52.6 & 51.4 & 55.6 \\
Iran & religion & 54.3 & 42.9 & 59.4 \\
\midrule
Japan & dates-of-significance & 46.9 & 54.5 & 70.2 \\
Japan & etiquette & 66.7 & 71.9 & 90.2 \\
Japan & family & 48.9 & 54.5 & 72.7 \\
Japan & greetings & 68.9 & 88.9 & 67.4 \\
Japan & religion & 60.0 & 43.6 & 77.6 \\
\midrule
Poland & dates-of-significance & 48.6 & 48.1 & 67.6 \\
Poland & etiquette & 35.7 & 32.0 & 60.4 \\
Poland & family & 42.9 & 33.9 & 66.7 \\
Poland & greetings & 42.9 & 44.1 & 64.3 \\
Poland & religion & 46.7 & 35.4 & 58.1 \\
\midrule
South Africa & dates-of-significance & 33.6 & 47.2 & 60.0 \\
South Africa & etiquette & 41.2 & 50.6 & 64.0 \\
South Africa & greetings & 44.7 & 43.8 & 74.4 \\
South Africa & religion & 39.2 & 32.0 & 59.6 \\
\midrule
\textbf{Overall} & & \textbf{51.3} & \textbf{49.8} & \textbf{69.3} \\
\bottomrule
\end{tabular}
\caption{CultureScore (\%) by country and category for Wan 2.2 \cite{wan2025wanopenadvancedlargescale} with the extended prompt.}
\label{tab:wan_extended_by_category}
\end{table*}

\begin{table*}[ht]
\centering
\begin{tabular}{ll CCC}
\toprule
 & & \multicolumn{3}{c}{\textbf{CultureScore (\%)}} \\
\cmidrule(lr){3-5}
\textbf{Country} & \textbf{Category} & \textbf{Identity} & \textbf{Behavior} & \textbf{Context} \\
\midrule
Brazil & dates-of-significance & 33.3 & 66.7 & 50.0 \\
Brazil & etiquette & 0.0 & 28.6 & 66.7 \\
Brazil & family & 50.0 & 80.0 & 66.7 \\
Brazil & greetings & 0.0 & 71.4 & 50.0 \\
Brazil & religion & 0.0 & 66.7 & 60.0 \\
\midrule
Canada & dates-of-significance & 0.0 & 0.0 & 0.0 \\
Canada & etiquette & 50.0 & 50.0 & 20.0 \\
Canada & family & 66.7 & 80.0 & 50.0 \\
Canada & greetings & 16.7 & 25.0 & 33.3 \\
Canada & religion & 50.0 & 100.0 & 16.7 \\
\midrule
Chile & dates-of-significance & 50.0 & 33.3 & 0.0 \\
Chile & etiquette & 83.3 & 33.3 & 80.0 \\
Chile & family & 33.3 & 0.0 & 20.0 \\
Chile & greetings & 40.0 & 66.7 & 60.0 \\
Chile & religion & 28.6 & 66.7 & 66.7 \\
\midrule
China & dates-of-significance & 33.3 & 16.7 & 33.3 \\
China & etiquette & 83.3 & 60.0 & 60.0 \\
China & family & 0.0 & 0.0 & 50.0 \\
China & greetings & 28.6 & 28.6 & 83.3 \\
China & religion & 28.6 & 66.7 & 42.9 \\
\midrule
Germany & dates-of-significance & 28.6 & 28.6 & 50.0 \\
Germany & etiquette & 50.0 & 33.3 & 100.0 \\
Germany & family & 33.3 & 40.0 & 50.0 \\
Germany & greetings & 66.7 & 42.9 & 16.7 \\
Germany & religion & 20.0 & 0.0 & 33.3 \\
\midrule
India & dates-of-significance & 16.7 & 0.0 & 66.7 \\
India & etiquette & 33.3 & 50.0 & 33.3 \\
India & family & 14.3 & 20.0 & 20.0 \\
India & greetings & 33.3 & 50.0 & 0.0 \\
India & religion & 50.0 & 42.9 & 16.7 \\
\midrule
Iran & dates-of-significance & 14.3 & 16.7 & 33.3 \\
Iran & etiquette & 66.7 & 14.3 & 66.7 \\
Iran & family & 16.7 & 42.9 & 66.7 \\
Iran & greetings & 50.0 & 80.0 & 50.0 \\
Iran & religion & 57.1 & 33.3 & 66.7 \\
\midrule
Japan & dates-of-significance & 33.3 & 16.7 & 50.0 \\
Japan & etiquette & 0.0 & 0.0 & 66.7 \\
Japan & family & 0.0 & 57.1 & 28.6 \\
Japan & greetings & 33.3 & 62.5 & 33.3 \\
Japan & religion & 33.3 & 16.7 & 33.3 \\
\midrule
Poland & dates-of-significance & 16.7 & 50.0 & 16.7 \\
Poland & etiquette & 0.0 & 0.0 & 50.0 \\
Poland & family & 0.0 & 14.3 & 0.0 \\
Poland & greetings & 0.0 & 71.4 & 66.7 \\
Poland & religion & 16.7 & 12.5 & 0.0 \\
\midrule
South Africa & dates-of-significance & 16.7 & 0.0 & 14.3 \\
South Africa & etiquette & 16.7 & 16.7 & 0.0 \\
South Africa & greetings & 16.7 & 60.0 & 40.0 \\
South Africa & religion & 75.0 & 33.3 & 33.3 \\
\midrule
\textbf{Overall} & & \textbf{29.7} & \textbf{36.0} & \textbf{41.0} \\
\bottomrule
\end{tabular}
\caption{CultureScore for Veo 3.1 Fast \cite{GoogleDeepMind2025Veo3} with the base prompt, by country and category.}

\label{tab:accuracy_veo_base}
\end{table*}

\begin{table*}[ht]
\centering
\begin{tabular}{ll CCC}
\toprule
 & & \multicolumn{3}{c}{\textbf{CultureScore (\%)}} \\
\cmidrule(lr){3-5}
\textbf{Country} & \textbf{Category} & \textbf{Identity} & \textbf{Behavior} & \textbf{Context} \\
\midrule
Brazil & dates-of-significance & 50.0 & 100.0 & 83.3 \\
Brazil & etiquette & 14.3 & 14.3 & 50.0 \\
Brazil & family & 50.0 & 80.0 & 66.7 \\
Brazil & greetings & 50.0 & 71.4 & 83.3 \\
Brazil & religion & 0.0 & 33.3 & 40.0 \\
\midrule
Canada & dates-of-significance & 0.0 & 57.1 & 57.1 \\
Canada & etiquette & 50.0 & 100.0 & 40.0 \\
Canada & family & 50.0 & 80.0 & 33.3 \\
Canada & greetings & 66.7 & 75.0 & 83.3 \\
Canada & religion & 50.0 & 100.0 & 33.3 \\
\midrule
Chile & dates-of-significance & 50.0 & 66.7 & 20.0 \\
Chile & etiquette & 66.7 & 33.3 & 60.0 \\
Chile & family & 33.3 & 20.0 & 20.0 \\
Chile & greetings & 60.0 & 66.7 & 40.0 \\
Chile & religion & 42.9 & 50.0 & 50.0 \\
\midrule
China & dates-of-significance & 33.3 & 16.7 & 83.3 \\
China & etiquette & 83.3 & 40.0 & 100.0 \\
China & family & 50.0 & 42.9 & 50.0 \\
China & greetings & 28.6 & 71.4 & 83.3 \\
China & religion & 57.1 & 66.7 & 100.0 \\
\midrule
Germany & dates-of-significance & 28.6 & 28.6 & 50.0 \\
Germany & etiquette & 50.0 & 16.7 & 60.0 \\
Germany & family & 50.0 & 60.0 & 33.3 \\
Germany & greetings & 83.3 & 85.7 & 66.7 \\
Germany & religion & 0.0 & 0.0 & 50.0 \\
\midrule
India & dates-of-significance & 33.3 & 42.9 & 50.0 \\
India & etiquette & 66.7 & 25.0 & 83.3 \\
India & family & 71.4 & 80.0 & 60.0 \\
India & greetings & 33.3 & 50.0 & 100.0 \\
India & religion & 50.0 & 57.1 & 16.7 \\
\midrule
Iran & dates-of-significance & 28.6 & 16.7 & 66.7 \\
Iran & etiquette & 16.7 & 14.3 & 83.3 \\
Iran & family & 33.3 & 57.1 & 50.0 \\
Iran & greetings & 66.7 & 60.0 & 50.0 \\
Iran & religion & 71.4 & 66.7 & 83.3 \\
\midrule
Japan & dates-of-significance & 16.7 & 16.7 & 66.7 \\
Japan & etiquette & 33.3 & 33.3 & 100.0 \\
Japan & family & 33.3 & 85.7 & 57.1 \\
Japan & greetings & 33.3 & 87.5 & 83.3 \\
Japan & religion & 33.3 & 50.0 & 50.0 \\
\midrule
Poland & dates-of-significance & 50.0 & 50.0 & 50.0 \\
Poland & etiquette & 50.0 & 33.3 & 50.0 \\
Poland & family & 28.6 & 71.4 & 57.1 \\
Poland & greetings & 20.0 & 71.4 & 66.7 \\
Poland & religion & 50.0 & 37.5 & 57.1 \\
\midrule
South Africa & dates-of-significance & 33.3 & 42.9 & 42.9 \\
South Africa & etiquette & 33.3 & 66.7 & 20.0 \\
South Africa & greetings & 16.7 & 80.0 & 40.0 \\
South Africa & religion & 50.0 & 33.3 & 66.7 \\
\midrule
\textbf{Overall} & & \textbf{41.9} & \textbf{52.1} & \textbf{59.0} \\
\bottomrule
\end{tabular}
\caption{CultureScore by country for Veo 3.1 Fast \cite{GoogleDeepMind2025Veo3} with the extended prompt.}
\label{tab:CultureScore_veo_extended}
\end{table*}

\end{document}